%% file: body.tex
\newcommand\SEKIusersusepackages
{\usepackage{latexsym,amssymb,graphicx}
\input header

\input headerforformulas
\input headersugarterms

\input headerframe
\input headerepsi
\input quotation
}
\newcommand\SEKIREVIEWSTATUS{1}
\newcommand\SEKIREVIEWER
{\horacekname
\\\addressuniSBcomplete
\\E-mail: {\tt horacek@ags.uni-sb.de}
\\WWW: \url{http://www.ags.uni-sb.de/~horacek}}
\newcommand\SEKIDOCUMENTCLASS{0}
\newcommand\SEKIYEAR{2006}
\newcommand\SEKINUMBEROFYEAR{02}
\newcommand\SEKITYPE{0}
\newcommand\SEKIPUBLISHEDTYPE{2}
\newcommand\SEKIPUBLISHEDAS
{{\em A New Indefinite Semantics for {H}ilbert's epsilon},
 \eleventhTABLEAUtwo{298}{314}, 2002,
 \url{www.ags.uni-sb.de/~cp/p/epsi}}
\newcommand\thepeinsexample
{\noindent\mbox{}\getittotheright
{\em (continuing \examref{example choice-conditions and predecessor})}\notop
\noitem\begin{example}[Instantiation with Higher-Order \CC]\label
{example cyclic lemma application}\\
Suppose that\par\noindent 
\inpit{\psymbol1}~~~\mbox{\math{
\forall v\stopq\inparentheses{\exists y\stopq\inpit
{v\tightequal y\tight+\onepp}\implies\inpit{v\tightequal\ppp v\tight+\onepp}}}}
\par\noindent
is one of our lemmas for the predecessor function \nlbmath\psymbol\
in the arithmetic of natural numbers,
and that we want to use this lemma as justification for replacing 
\sforallvari y{\inpit{\psymbol1}} under \math R-\cc\
\bigmaths{\app{C}{\sforallvari y{\inpit{\psymbol1}}}=
\lambda v\stopq\inpit{v\tightequal\sforallvari y{\inpit{\psymbol1}}(v)\tight+\onepp}}{}
globally with \nlbmath\psymbol\@. \ 
Note that this was required in \examref
{example choice-conditions and predecessor}. \ 
By \theoref{theorem strong reduces to}(6), 
for \math{\sigma:=\{\sforallvari y{\inpit{\psymbol1}}\tight\mapsto\psymbol\}},
we have to show
\bigmaths{\inpit{\app{Q_C}{\sforallvari y{\inpit{\psymbol1}}}}\sigma},
which does not seem to be any problem because 
\bigmath{\inpit{\app{Q_C}{\sforallvari y{\inpit{\psymbol1}}}}\sigma}
is just the above lemma \nlbmath{\inpit{\psymbol1}}.\end{example}}
\title{\hilbert's epsilon as\\an Operator of Indefinite Committed Choice}
\author{\wirthname
\\{\small\Institute}
\\{\small\emailcp}}
\date
{\small Submitted \Dec\,16, 2004
\\First Print Edition: August\,25, 2006
\\Improved \Jan\,28, 2007
\\Minorly improved (\noteref{note bell} added) \Apr\,5, 2009 
\\Minorly improved 
(\sectref{section linguistic literature}, \noteref{note liberalized}), 
\May\,21, 2010 
\SEKIedition
\\Minorly improved 
(\sectref{section quantifiers}), 
\Jan\,16, 2012 
\SEKIedition\vspace*{-.5cm}}
\input seki-deckblatt-4
\usepackage{named}
\input namedhelper

\catcode`\@=11
\input ENDNOTES.sty
\catcode`\@=12
\def\enotesize{\normalsize}
\def\endnoteendsep{\vspace{5.0\topsep}}
\let\footnote=\endnote
\newcommand\daspaper{paper}

\newcommand\HIII{\ident{Heinrich\,III}}
\newcommand\HIV{\ident{Heinrich\,IV}}
\newcommand\HG{\ident{Holy\,Ghost}}
\newcommand\strongexpansionrule[6]{\LINEmath{\begin
{array}[t]{@{}c@{}l@{\mbox{~~~~~~}}l@{}}#1&&#3\\\cline
{1-1}\mediumheadroom#2&&#4\\\end{array}}}
\def\vec#1{\mathchoice{\mbox{\boldmath$\displaystyle#1$}}
{\mbox{\boldmath$\textstyle#1$}}
{\mbox{\boldmath$\scriptstyle#1$}}
{\mbox{\boldmath$\scriptscriptstyle#1$}}}
\newcommand\sentenceonsetofvariables
{We use `\math{\VARsingleindex k(\Gamma)}' \nolinebreak 
to denote the set of variables from
\nolinebreak\Vsingleindex k occurring in \nlbmath\Gamma. \ 
}

\newcommand\whatitisgoodfor
{serve as a paradigm useful in the specification and computation 
 of semantics of discourses in natural language}
\catcode`\@=11
\renewcommand\tableofcontents{%
    \section*{\contentsname
        \@mkboth{%
           \MakeUppercase\contentsname}{\MakeUppercase\contentsname}}%
    \vskip .1ex
    \@starttoc{toc}%
    }
\def\l@section#1#2{%
  \ifnum \c@tocdepth >\z@
    \addpenalty\@secpenalty
    \addvspace{1.5em \@plus\p@}%
    \setlength\@tempdima{2.5em}%
    \begingroup
      \parskip -8pt
      \parindent \z@ \rightskip \@pnumwidth
      \parfillskip -\@pnumwidth
      \leavevmode
      \advance\leftskip\@tempdima
      \hskip -\leftskip
      \bf#1\nobreak\hfil \nobreak\hb@xt@\@pnumwidth{\hss #2}\par
    \endgroup
  \fi}
\catcode`\@=12
\newcommand\litsectrefstwoandthreepartone{\sectrefs
{section general introduction}{section in the literature}}
\newcommand\sectrefsectionrightuniquesemantics
{\sectref{section right-unique semantics}}
\begin{document}
\makecover
% Let your text start here, possibly changing the following a lot.
\maketitle
\begin{abstract}\sloppy%
% After reviewing the literature on semantics of \hilbert's epsilon operator,
% we present a new one that is similar to some cases of 
% referential interpretation of indefinite articles in natural language.
\bernaysname\ and \hilbertname\ carefully avoided overspecification of 
\hilbert's \mbox{\math\varepsilon-operator} and axiomatized only what was
relevant for their proof-theoretic investigations.
Semantically, this left the \math\varepsilon-operator underspecified.
In the meanwhile, 
there have been several suggestions for semantics of the \nlbmath\varepsilon\
as a choice operator. 
After reviewing the literature on semantics of \hilbert's epsilon operator,
we propose a new semantics with the following features:
We avoid overspecification (such as right-uniqueness),
but admit indefinite choice, committed choice, and classical logics.
Moreover,
our semantics for the \nlbmath\varepsilon\ 
supports proof search optimally and
is natural in the sense that it does not only mirror some cases of 
referential interpretation of indefinite articles in natural language,
but may also contribute to philosophy of language.
Finally, we ask the question whether our \nlbmath\varepsilon\ within our 
free-variable framework can \whatitisgoodfor.\vspace*{-1ex}
\end{abstract}
{\small\tableofcontents}\vfill\pagebreak

\section{Motivation, Requirements Specification, and Overview}\label
{section requirement specification}

In \cite{wirthcardinal} we have analyzed the combination of 
mathematical induction in the liberal style of \fermat's\emph\descenteinfinie\
with state-of-the-art logical deduction into a formal system in which
a working mathematician can straightforwardly develop his proofs 
supported by powerful automation.
We \nolinebreak have found only a single semantical justification
meeting the requirements resulting from this analysis.
The means for this semantical justification include
a novel semantics for \hilbert's \nlbmath\varepsilon-symbol,
namely an indefinite choice mirroring some cases of 
referential interpretation of indefinite articles in natural languages.

\hilbert's \mbox{\math\varepsilon-symbol}
is a binder that forms terms; \ 
just like \peano's \math\iota-symbol, which is  
some\-times\footnote
{\label{note history of iota}{\bf (History of the Symbols used to denote 
 the \math\iota-Binder)}\par\noindent
It may be necessary to say something on the symbols
used for the \nlbmath\iota\ in the \nth{19} and \nth{20} century. \ 
In \nolinebreak\cite{peanoiotabar}, \peanoname\ \peanolifetime\
wrote \nlbmath{\bar\iota} instead of the \nlbmath\iota\ of
\examref{example iota},
and
\ \mbox{\maths{\bar\iota\setwith x A}{}} \ 
\mbox{instead of \ \math{\iota x.\,A}}. \ \ 
(Note that we have changed the class notation to modern standard here.  
 We \nolinebreak will do so in the following without mentioning it. 
 \peano\ actually wrote \maths{\overline{x\tight\in}A}{} 
 instead of \setwith x A in \cite{peanoiotabar}.) \ 

More than in \frege's logic calculus,
\peano\ was interested in logic as a written language (ideography)
with a clear description of its semantics in natural language.
He also created an artificial substitute for natural language 
(\tightemph{Latino sine flexione}, \cfnlb\ \eg\ \cite{kennedypeanolife}).
Therefore,
it \nolinebreak does not come as a surprise that 
it was \peano\ who 
\mbox{invented the \math\iota-\em binder}.
\Cf, however, \noteref{note frege} on \frege's \math\iota-operator of 1893.
In \cite{peanoiotabargerman}, we find an alternative notation besides
\math{\bar\iota}, namely a \math\iota-symbol upside-down,
\ie\nolinebreak\ inverted,
\ie\nolinebreak\ rotated by \math\pi\ around its center.
I \nolinebreak do not know whether this is the first occurrence of the 
\mbox{inverted \math\iota-symbol}. \ 
It \nolinebreak was later used also in \nolinebreak\cite{PM}, the 
infamous\emph\PM\ first published in 1910\ff. \ 
Thus, we should speak of\emph{\peano's \nlbmath\iota-symbol} 
and not of\emph{\russell's \math\iota-symbol}.

We call the famous\emph\PM\ infamous, 
because it is still rare and unaffordable, 
 and---as standard notions and notation have changed quite a bit 
 in the meanwhile---has 
 become also quite 
 incomprehensible for the occasional reader. \ 
 It is a shame that there is no public interactive \WWW\
 version of the\emph{Principia}, 
 which facilitates look-up by translation into modern notation
 and online help with obsolete names.

Let us come back to \peano's \nlbmath\iota. \ 
The bar above as well as the inversion of the \nlbmath\iota\
were to indicated that \math{\bar\iota} was implicitly
defined as the inverse operator of the 
operator \math\iota\ defined by \bigmaths{\iota y:=\{y\}}, 
which occurred already in \cite{peanoiota} and still in \cite{ML}.
\par\halftop\noindent
The definition of \nlbmath{\bar\iota}
reads literally \cite[\litdefiref{22}]{peanoiotabar}:
\par\noindent\phantom{}\LINEmaths{
a\tightin K\ .\ \exists a:x,y\in a\ .\supset_{x,y}.\ x\tightequal y:\ \supset\ :
x=\bar\iota a\ .=.\ a\tightequal\iota x
}{}\par\noindent
This straightforwardly translates into more modern notation as follows:
\par\noindent\mbox{}\hfill\mbox{\math{
  \mbox{For any class \nlbmath a:\ \ \ \ }
  a\tightnotequal\emptyset\und
  \forall x,y\stopq\inpit{x,y\in a\nottight{\nottight\implies}x\tightequal y}
  \nottight{\nottight{\nottight\implies}}\forall x\stopq\inpit{
  x=\bar\iota a\nottight{\nottight\equivalent}a\tightequal\iota x}
}}\par\noindent
Giving up the flavor of an explicit definition of 
\ \ ``~\math{x=\bar\iota a}~\closequotecomma\mbox{}~~this can be 
simplified to the following logically equivalent form:
\par\noindent\phantom{}\LINEmaths{
\mbox{For any class \nlbmath a:}\ \ \ \  
\exists! x\stopq x\tightin a
\nottight{\nottight{\nottight\implies}}\bar\iota a\tightin a
}{}(\math{\bar\iota_0})\par\noindent
Besides notational difference, this is (\math{\iota_0})
of our \sectref{section Semantics of the iota-Operator}.

It has become standard to write a simple non-inverted \nlbmath\iota\ for 
the upside-down \math\iota\ because
\peano's original notation \ ``\,\math{\iota\tight y}\,'' \  
has long ago been replaced with \ ``\,\nlbmath{\{y\}}\,'' \ 
and because the upside-down \math\iota\ 
is not easily available in today's
typesetting. \ 
For instance, there does not seem to exist a \TeX\ macro for it
 and---to enable font-independent archiving and republishing---some 
 publishers do not permit the usage of nonstandard symbols.\vfill\pagebreak} 
attributed to \russell\ and written as
 \nlbmath{\bar\iota} or as an inverted \nlbmath\iota. \ 
Roughly speaking, the \mbox{term \ \math{\varepsilon x.\, A}} \  
formed from a variable \nlbmath x
and a formula \math A 
denotes\emph{an} object 
that is\emph{chosen} such that \mbox{---if} \nolinebreak possible---\bigmath
A \nolinebreak(seen as a predicate on \nlbmath x) \mbox{holds for it}.

For the usefulness of\emph{descriptive terms} such as 
\ \mbox{\math{\varepsilon x.\, A}} \ and \ \mbox{\math{\iota x.\, A}}, \  
we consider the requirements listed below to be the most important ones. \ 
Our new indefinite \mbox{\math\varepsilon-operator} satisfies these requirements
and---as it is defined by novel semantical techniques---may 
serve as the paradigm for the design of similar operators
satisfying these requirements.
As such descriptive terms are
of universal interest and applicability,
we \nolinebreak suppose that our novel treatment will turn out to be useful in 
many additional areas
where logic is designed or applied as a tool for description and reasoning.
\newcommand\requirementeins{Requirement\,I\ (Syntax)}
\newcommand\requirementzwei{Requirement\,II\ (Reasoning)}
\newcommand\requirementdrei{Requirement\,III\ (Semantics)}
\begin{description}
\item[\requirementeins: ]
The syntax must clearly express
where exactly 
a\emph{commitment} to a choice of a special object is required,
%corresponding with the description
%satisfying \nlbmath A (s a predicate on \nlbmath x) is required, 
and where---to the contrary---different
objects corresponding with the description
%satisfying \nlbmath A 
may be chosen for different occurrences of the
same descriptive term.
%\math\varepsilon-term \nolinebreak\ \mbox{\math{\varepsilon x.\, A}}. \ \ 
\item[\requirementzwei: ] \sloppy
In a reductive proof step,
it must be possible 
to replace a descriptive term
%an \mbox{\math\varepsilon-term \ \ \math{\varepsilon x.\,A}} \ 
with a term %\nlbmath t 
that corresponds with its description.
% roughly speaking in the sense that 
% %satisfies the predicate \nlbmath A or---more precisely---the 
% the formula \nlbmath{A\{x\tight\mapsto t\}} is valid.
The soundness of such a replacement must be 
expressible and should be verifiable in the original calculus.
\item[\requirementdrei: ]
The semantics should be simple, straightforward, natural, formal, 
and model-based.
Overspecification should be avoided carefully.
Furthermore, the semantics should be modular and abstract in the sense that
it adds the operator to a variety of logics,
independently of the details of a concrete logic.
\end{description}
This \daspaper\ organizes as follows:
%In \sectref{section requirement specification} the basic requirements on the 
%goals of this \daspaper\ are briefly specified.
After a general introduction to the \nlbmath\varepsilon\ 
in \sectref{section general introduction} 
and a review of 
the literature on the \nlbmath\varepsilon's
semantics \wrt\ adequacy and \hilbert's intentions
in \sectref{section in the literature},
we explain and formalize our novel approach to 
the \nlbmath\varepsilon's semantics,
first informally in \sectref{section new indefinite}
and then formally in \sectref{section formal discussion}.
Finally, in \sectref{section philosophy of language},
we discuss some possible implications on
philosophy of language,
put our \math\varepsilon\ to test with a list of linguistic examples,
and ask the question whether our \nlbmath\varepsilon\ within our 
free-variable framework can \whatitisgoodfor.
\vfill\pagebreak

\section{General Introduction to \hilbert's \math\varepsilon}
\label{section general introduction}
To make this \daspaper\ accessible to a broader readership,
in this \sectref{section general introduction}, 
we motivate the \math\varepsilon\ by 
introducing first the \math\iota\ (\sectref{section from iota to epsilon}),
then the \nlbmath\varepsilon\ itself (\sectref{section epsilon}),
its proof-theoretic origin (\sectref{section proof-theoretic origin}),
and our contrasting semantical objective in this \daspaper\ 
(\sectref{section our objective})
with its emphasis on\emph{definite choice} 
(\sectref{section indefinite choice})
and\emph{committed choice} (\sectref{section committed choice}).
Although the well-informed expert is likely to be amused,
he may well skip this \sectref{section general introduction} and continue with 
\sectref{section in the literature}.

\subsection
{From the \math\iota\ to the \math\varepsilon}\label
{section from iota to epsilon}
\subsubsection{Intuition behind the \math\iota-Operator}
It has turned out not to be completely superfluous to remark that
I do not want to hurt any religious feelings with the following example. 
The delicate subject is chosen for its mnemonic value.

\begin{example}[\math\iota-binder]\hfill{\em(Buggy!)}\label
{example iota}\par\noindent
For an informal introduction to the \nlbmath\iota-binder, 
consider \math\Fathersymbol\ to be a predicate
for which\par\noindent\LINEmath{\Fatherpp\HIII\HIV
}\par\noindent holds,
\ie\ %\\\linenomath
{``Heinrich\,III is father of Heinrich\,IV\closequotefullstop} 
\\Now, %\\\linenomath
{``\tightemph{the} father of Heinrich\,IV''} can be denoted by 
\bigmaths{\iota x.\,\Fatherpp x\HIV},
and because this is nobody but Heinrich\,III, \ie\ 
\par\noindent\LINEmaths{
\HIII\nottight{\nottight =}\iota x.\,\Fatherpp x\HIV
},\par\noindent
we know that \par\noindent\LINEmath{
\Fatherpp{\iota x.\,\Fatherpp x\HIV}\HIV}\par\noindent
Similarly, 
\par\noindent\phantom{\ref{example iota}.1,}\LINEmaths
{\Fatherpp{\iota x.\,\Fatherpp x{\ident{Adam}}}{\ident{Adam}}},
(\ref{example iota}.1)\par\noindent
and thus \bigmaths{\exists y.\,\Fatherpp y{\ident{Adam}}},
but, oops! Adam and Eve do not have any fathers.\par\noindent
If you do not agree, you would probably appreciate the following problem
that occurs when somebody has God as an additional father.
\par\noindent\phantom.\LINEmaths{
  \Fatherpp\HG{\ident{Jesus}}
  \nottight{\nottight\und}
  \Fatherpp{\ident{Joseph}}{\ident{Jesus}}
}.(\ref{example iota}.2)\par\noindent
Then the Holy Ghost is\emph{the} father of Jesus 
and Joseph is\emph{the} father of Jesus,
\ie\par\noindent\LINEmath{
  \HG
  =
  \iota x.\,\Fatherpp x{\ident{Jesus}}
  {\nottight\und}
  \ident{Joseph}
 =
 \iota x.\,\Fatherpp x{\ident{Jesus}}
}(\ref{example iota}.3)\par\noindent
which implies something\emph{the} Pope may not accept, namely that
\par\noindent\LINEmaths{
  \HG
  \nottight{\nottight =}
  \ident{Joseph}
},\par\noindent
and he anathematized Heinrich\,IV in the year 1076:
\par\noindent\LINEmaths{
  \ident{Anathematized}\beginargs\iota x.\,\ident{Pope}{\beginargs x\allargs}
  \separgs\HIV\separgs{1076}\allargs
}.(\ref{example iota}.4)\end{example}\vfill\pagebreak

\subsubsection{Semantics of the \math\iota-Operator}\label
{section Semantics of the iota-Operator}
There are basically\footnote
{\label{note frege}{\bf
(Other \math\iota-Operators Besides those of \russell, \hilbert, and \peano)}
\par\noindent
In \cite[\Vol\,I, \litsectref{11}]{frege-grundgesetze}, we find another
\math\iota-operator. As this \Vol\,I was published by 
\fregename\ \fregelifetime\ 
in\,1893, this seems
to be the first occurrence of a \mbox{\math\iota-operator} in the literature.
The symbol he uses for the \math\iota\ is a boldface backslash.
As a\emph{boldface} version of the 
backslash does not seem to be available in standard \TeX, 
we use a simple 
backslash \nolinebreak(\ttbackslash) here. \ 
\frege\ defines \bigmaths{\mbox\ttbackslash\xi:=x}{} 
\udiff\ there is some \math x
such that 
\ \mbox{\math{\forall y\stopq\inpit{\app\xi y=\inpit{x\tightequal y}}}}. \ 
Writing the binder as a modern \nlbmath\lambda\ 
instead of \frege's\emph{spiritus lenis},
\frege\ actually requires extensional equality of \nlbmath\xi\
and \bigmaths{\lambda y.\,\inpit{x\tightequal y}}.
Now this would be basically \peano's \math\iota-operator
(\cfnlb\ our \sectref{section Semantics of the iota-Operator} and
 \noteref{note history of iota})
unless \frege\ overspecified it by defining
\bigmaths{\mbox\ttbackslash\xi:=\xi}{} for all other cases.
\par
Similarly, in set theories without urelements, the \math\iota-operator
is often defined by something like
\bigmaths{\iota y.\,A:=
\setwith{z}{\exists x\stopq\inpit{z\tightin x\und\forall y\stopq\inpit
{A\equivalent\inpit{x=y}}}}}{} for new \math x and \math z,
\cf\ \eg\ \cite{ML}.
This is again an overspecification resulting in 
\bigmaths{\iota y.\,A=\emptyset}{} in case of 
\bigmaths{\neg\exists! y.\,A}.}
three ways of giving semantics to the \math\iota-terms:
\halftop\begin{description}\item[\russell's \math{\iota}-operator: ]
In \cite{PM}, \ 
the \nlbmath\iota-terms do not refer to an object
but make sense only in the context of a sentence.
This was nicely described already in \cite{denoting}, without
using any symbol for the \nlbmath\iota, however; \ 
\cfnlb\ our \sectref{section russell meinong}. \ 
\item[\hilbert's \math\iota-operator: ]\sloppy
To overcome the complex difficulties of that non-referential definition,
in \cite[\Vol\,I, \p\,392\ff]{grundlagen}, \ 
a completed proof of \bigmath{\exists!x.\,A} was required
to precede any formation of the term \bigmaths{\iota x.\,A},
which otherwise was not to be considered a well-formed term at 
\nolinebreak all.
\item[\peano's \math\iota-operator: ]
Since the inflexible treatment of \hilbert's \nlbmath\iota-operator
makes the \nlbmath\iota\ quite impractical and 
the formal syntax of logic undecidable in general, \ 
in \Vol\,II of the same book, \ 
the \math\varepsilon, \ however, \ is already given a more flexible treatment. \
There, the simple idea is to leave the \math\varepsilon-terms uninterpreted, \ 
as will be described below. \ 
In this \daspaper, \  
we present this more flexible view also for the \nlbmath\iota, \ 
just as required by an anonymous referee of a previous version of this 
\daspaper. \ 
Moreover, \
this view is already \peano's original one, \
\cfnlb\ (\math{\bar\iota_0}) of \noteref{note history of iota}.
\end{description}
At least in non-modal classical logics, it is a well justified standard 
that\emph{any term denotes}.
More precisely---in \nolinebreak
each model or structure
\nlbmath\salgebra\ under consideration---any occurrence of
a proper term must denote an 
object in the universe of \nlbmath\salgebra.
\ (This does not mean that this object has to satisfy properties
   of ontological existence or definedness in \nlbmath\salgebra, 
\ \cf\ \sectref{section quadrangle is quadrangular new short}.) \ 
Following that standard, 
to be able to write down \ \mbox{\math{\iota x.\, A}} \ 
without further consideration,
we have to treat \nolinebreak\ \mbox{\math{\iota x.\, A}} \ 
as an uninterpreted term about which we only know
\par\yestop\noindent\LINEmaths{
  \exists!x.\, A\nottight{\nottight\implies} A\{x\mapsto\iota x.\, A\}
}{}(\math{\iota_0})\par\yestop\noindent
or in different notation 
\par\noindent\LINEmaths{
  \inpit{\exists!x.\,\inpit{\app A x}}
  \nottight{\nottight\implies}\app A{\iota x.\,\inpit{\app A x}}
}{}\par\noindent
or in set notation 
\par\noindent\LINEmaths{
  \exists!x\stopq\inpit{x\tightin A}
  \nottight{\nottight\implies}\iota x\stopq\inpit{x\tightin A}\in A
}{}\par\halftop\noindent
where, for some new \nlbmath y, we can define
\par\halftop\noindent
\LINEmaths{
\exists!x.\,A\nottight{\nottight{\nottight{\nottight{:=}}}}
\exists y.\,\forall x.\,\inparentheses{x\boldequal y\nottight\equivalent A}}
{}\par\yestop\noindent
With (\math{\iota_0}) as the only axiom 
for the \nlbmath\iota, 
the term \bigmaths{\iota x.\,A}{} has to 
satisfy \nlbmath A (seen as a predicate on \nlbmath x) only if there
exists a unique object such that \math A \nolinebreak holds for it. \ 
Moreover, the 
problems presented in \examref{example iota} do not appear 
because
(\ref{example iota}.1) and (\ref{example iota}.3) are not valid. \ 
Indeed, the description of (\ref{example iota}.1)
lacks existence and the descriptions of 
(\ref{example iota}.3) and (\ref{example iota}.4) 
lack uniqueness. \ 
The price we have to pay here is that---roughly speaking---\ \mbox
{\math{\iota x.\, A}} \ is of no use unless the unique existence 
\bigmaths{\exists!x.\, A}{} can be derived.
\vfill\pagebreak

\subsubsection{Why \math\varepsilon\ is more useful than \math\iota}\label
{section Why}
Compared to the \nlbmath\iota, \ 
the \nlbmath\varepsilon\ is more useful
because---instead of (\math{\iota_0})---it comes with the stronger axiom
\par\halftop\noindent\phantom{(\math{\varepsilon_0})}\LINEmath
{\exists x.\, A\nottight\implies A\{x\mapsto\varepsilon x.\, A\}
}(\math{\varepsilon_0})\par\yestop\noindent
More precisely, as the formula 
\bigmaths{\exists x.\, A}{}
(which has to be true to guarantee a meaningful interpretation
of the \math\varepsilon-term \bigmaths{\varepsilon x.\, A}{}) \ 
is weaker than the corresponding formula 
\bigmaths{\exists!x.\, A}{}
(for the \resp\ \nlbmath\iota-term),
the \nolinebreak area of useful application is wider for the \math\varepsilon-
than for the \math\iota-operator. \ 
Moreover, in case 
of \bigmaths{\exists!x.\, A}, \ the \nlbmath\varepsilon-operator picks 
the same element as the \math\iota-operator, \ \ie
\par\noindent\LINEmath
{\exists!x.\, A
 \nottight{\nottight\implies}
 \inparentheses{\varepsilon x.\, A\ =\ \iota x.\, A}}\par\noindent
Although the \nlbmath\iota\ is thus somewhat outdated 
since the appearance of the superior \nlbmath\varepsilon, \ 
it is still alive: 
\ \cfnlb\ \eg\ \cite{andrews}, \cite{isabellehol}. \ 
For example, in \cite{isabellehol}, \p\,85,
for the\emph{definiendum} \nolinebreak\bigmath{\mu x.\,A} of 
the \math\mu-operator (which picks the least natural number satisfying
a formula) \ we \nolinebreak find the\emph{definiens}
\par\noindent\LINEmaths{
\iota x.\,\inparentheses{A\nottight\und
\forall y.\,\inpit{A\{x\tight\mapsto y\}\implies x\tight\preceq y}}}
{}\par\noindent
for some new variable \nlbmath y,
although the logic of \ISABELLEHOL\ (as given in \cite{isabellehol}) \ 
\mbox{contains an \nlbmath\varepsilon},
and a\emph{definiens} of 
\par\noindent\LINEmaths{
\varepsilon x.\,\inparentheses{A\nottight\und
\forall y.\,\inpit{A\{x\tight\mapsto y\}\implies x\tight\preceq y}}}
{}\par\noindent
% or\par\yestop\noindent\LINEmaths{
% \varepsilon x.\,\inparentheses{A\nottight\und
% \neg\exists y.\,\inpit{A\{x\tight\mapsto y\}\und y\tight\nprec x}}}
% {}\par\yestop\noindent
would imply the relevant axiom
\par\yestop\noindent\phantom{(\math{\mu_0})}\LINEmaths{
\exists x.\, A\nottight{\nottight{\nottight{\nottight{\nottight\implies}}}}
A\{x\mapsto\mu x.\, A\}
\nottight{\nottight{\nottight\und}}
\forall y.\,\inpit{A\{x\tight\mapsto y\}\nottight{\nottight\implies}\mu x.\, A
\preceq y}}
{}(\math{\mu_0})\par\yestop\noindent
for any \wellfounded\ total quasi-ordering, 
while the definition via \nlbmath\iota\
requires antisymmetry in addition.
Moreover, a special additional 
uniqueness proof is required for each unfolding of the definition
via \nlbmath\iota, hopefully realized
automatically, however, with a closely integrated Linear Arithmetic,
\cf\ \eg\ \cite{samoacalculemus}. \ 
Indeed, the superior definition via \nlbmath\varepsilon\ is
found in \cite{isabelles-logics-hol}.

\yestop\yestop\subsection{What is \hilbert's \math\varepsilon?}\label
{section epsilon}

\yestop\noindent
As the basic methodology of \hilbertname's formal program is to treat
all symbols as meaningless, he does not give us 
any semantics but only the axiom (\nlbmath{\varepsilon_0}). \ 

\yestop\noindent
Although no meaning is required, 
it \nolinebreak furthers the understanding.
And therefore, in \cite{grundlagen}, 
the fundamental work on the contributions of \hilbertname\ and his group
to the logical foundations of mathematics,
\bernaysname\ writes:
\vfill\pagebreak 

\begin{quote}{\fraknomath
\math{\varepsilon x.\, A} \ 
\ldots\ ``\germantextone''}\getittotheright
{\cite[\Vol\,II, \p 12, modernized orthography]{grundlagen}}\end
{quote}\begin{quote}
\math{\varepsilon x.\, A} \ \ldots\ 
``\englishtextone''
\getittotheright{(our translation)}
\end{quote}

\yestop\yestop\begin{example}
[\math\varepsilon\ instead of \math\iota, part\,I]
\hfill{\em (continuing \examref{example iota})}%
\label{example epsilon instead of iota 1} 
\par\noindent
Just as for the \nlbmath\iota, for the \nlbmath\varepsilon\ we again 
have\par\LINEmath{
  \HIII\nottight{\nottight =}\varepsilon x.\,\Fatherpp x\HIV
}\\and\\\LINEmaths{
\Fatherpp{\varepsilon x.\,\Fatherpp x\HIV}\HIV
}.\par\noindent
But, from the contrapositive of (\nlbmath{\varepsilon_0}) and
\par\noindent\LINEmaths{
\neg\Fatherpp{\varepsilon x.\,\Fatherpp x{\ident{Adam}}}{\ident{Adam}}
},\\
we now can conclude that\\\LINEmaths{
  \neg\exists y.\,\Fatherpp y{\ident{Adam}}
}.\end{example}

\yestop\subsection{On the \math\varepsilon's Proof-Theoretic Origin}
\label{section proof-theoretic origin}

\yestop\noindent
\hilbertname\ did not need any semantics or precise intention for the
\math\varepsilon-symbol because it was introduced merely as a formal
syntactical device to facilitate proof-theoretic investigations,
motivated by the possibility to get rid of the existential and
universal quantifiers via
\par\halftop\noindent\LINEmath
{\mbox{}~~\exists x.\, A\nottight{\nottight\equivalent}
 A\{x\mapsto\varepsilon x.\, A\}}(\math{\varepsilon_1})\par\halftop\noindent
and
\par\halftop\noindent\LINEmath
{\mbox{}~~~~\forall x.\, A\nottight{\nottight\equivalent}
 A\{x\mapsto\varepsilon x.\,\neg A\}}(\math{\varepsilon_2})
\par\yestop\noindent
Note that \math{\varepsilon_0}, \math{\varepsilon_1}, 
 and \math{\varepsilon_2}
 are no ordinal numbers but simply the original labels from 
 \nolinebreak\cite{grundlagen}. \ 
 \math{\varepsilon_5} \nolinebreak is \nolinebreak from \cite{hermesepsilon}.
 The other labels we use are mostly from \nolinebreak\cite{leisenring},
 such as (E2) and (Q2). 
 We recommend \cite{leisenring} as 
 an excellent treatment of the subject of the \firstorder\ 
 \math\varepsilon-calculus, using a language more modern than the one of 
 \cite{grundlagen}.
\vfill\pagebreak

When we remove all quantifiers in a derivation of the \hilbert-style 
predicate calculus of \nolinebreak\cite{grundlagen}
along (\math{\varepsilon_1}) and (\math{\varepsilon_2}), the following 
transformations occur:
Tautologies 
% (``identical formulas'' in \nolinebreak\cite{grundlagen}) 
are turned into tautologies, the axiom schemes\footnote{%
% The bound variable \nlbmath x must be replaced with the free 
% variable \nlbmath{\wforallvari x{}} 
% (actually: \wfuv, \cf\ \sectref{section free})
% because  in \cite{grundlagen,wirthcardinal} and also in this \daspaper,
% free and bound variables come from disjoint sets.
\ \ To be precise, \ 
 in the standard predicate calculus of \cite{grundlagen} 
 there are no axiom schemes but only axioms with predicate variables. \ \ 
 The axiom schemes we use here simplify the presentation and refer
 to the\emph{modified form of the predicate calculus} of
 \cite[Vol.\,II, \p\,403]{grundlagen}, \ 
 which is closer to today's standard syntax of \firstorder\ logic.}
\ \ \bigmath{A\{x\tight\mapsto t\}\nottight{\nottight\implies}\exists x.\,A
} \ \ 
and
\ \ \bigmath{\forall x.\,A\nottight{\nottight\implies}A\{x\tight\mapsto t\}
} \ \ 
are turned into 
\par\yestop\noindent\phantom{(\tightemph{\math\varepsilon-formula})}\LINEmath
{A\{x\tight\mapsto t\}\nottight{\nottight\implies}A\{x\mapsto\varepsilon x.\,A\}
}(\tightemph{\math\varepsilon-formula})\par\yestop\noindent
and---roughly speaking \wrt\ two-valued logics---its contrapositive, 
respectively. 
The inference steps are turned into
inference steps:\emph{modus ponens} into\emph{modus ponens};
instantiation of free variables as well as quantifier introduction into
instantiation including \math\varepsilon-terms.
Finally, the \math\varepsilon-formula is taken as a new axiom scheme instead
of \nolinebreak(\math{\varepsilon_0}) because it has the advantage of
being free of quantifiers.

This argumentation is actually the start of the proof transformation of 
the\emph{\nth 1 \mbox{\math\varepsilon-theorem}}, 
in which the elimination of the
\math\varepsilon-formulas did not come easy to
\ackermannname\ and \hilbertname.

\begin{theorem}{\bf(\Extd\ \nth 1 \math\varepsilon-Theorem, 
 \cite[\Vol\,II, \p 79\f]{grundlagen})}\\
If we can derive \bigmath
{\exists x_1.\,\ldots
\exists x_r.\, A}
(containing no bound variables besides 
the ones bound by the prenex \math{\exists x_1.\,\ldots
\exists x_r.}) \ from the formulas
\math{P_1,\ldots,P_k} (containing no bound variables) 
in the predicate calculus 
(\incl, as axiom schemes, \mbox{\math\varepsilon-formula} and,
 for equality, reflexivity and substitutability),
then, from \math{P_1,\ldots,P_k}, 
in the elementary calculus (\ie\ tautologies plus\emph
{modus ponens} and instantiation of free variables),
we can derive a (finite) disjunction of the form
\bigmaths{
%  \displaystyle
  \bigvee_{i=0}^s\ A\{x_1,\ldots,x_r\mapsto t_{i,1},\ldots,t_{i,r}\}
}{} in a derivation where bound variables do not occur at all.\end{theorem}
Note that \math{r,s} range over natural numbers including \nlbmath 0, and that
\math A, \math{t_{i,j}}, and \math{P_i} are \math\varepsilon-free because
otherwise they would have to include (additional) bound variables.

\yestop\noindent
Moreover, the\emph{\nth 2 \math\varepsilon-Theorem} 
in \cite[\Vol\,II]{grundlagen},
states that the \nlbmath\varepsilon\ 
(just \nolinebreak as 
 the \nlbmath\iota, \cfnlb\ \cite[\Vol\,I]{grundlagen})
is a conservative extension of the predicate calculus
in the sense that any formal proof of an \math\varepsilon-free formula
can be transformed into a formal 
proof that does not use the \nlbmath\varepsilon\ at all.
Generally, however, it is not a conservative extension to
add the \math\varepsilon\ either with (\math{\varepsilon_0}), 
with \nolinebreak (\math{\varepsilon_1}), or 
with the \math\varepsilon-formula 
to other \firstorder\ logics---may they be weaker 
such as intuitionistic logic,\footnote{\label{footnote tau}{\bf
 (Consequences of the \math\varepsilon-Formula in Intuitionistic Logic)}\par
 \noindent Adding the \math\varepsilon\ 
 either with (\math{\varepsilon_0}), 
 with \nolinebreak (\math{\varepsilon_1}), or 
 with the \math\varepsilon-formula 
 (\cfnlb\ \sectrefs{section Why}{section proof-theoretic origin})
 to intuitionistic \firstorder\ logic
 is equivalent on the \math\varepsilon-free theory to adding\emph
 {\plato's Principle}, \ie\ \ \mbox{\maths
 {\exists x\stopq\inpit{\exists y.\,A\implies A\{y\tight\mapsto x\}}}{}} \ 
 with \math x not occurring in \nlbmath A, 
 \cf\ \cite[\litsectref{3.3}]{meyervioldiss}. \\ 
 Moreover, the non-trivial direction of (\math{\varepsilon_2}) is
 \getittotheright{\maths
 {\forall x.\, A\nottight{\nottight\antiimplies}
  A\{x\mapsto\varepsilon x.\,\neg A\}}.}\\
 Even intuitionistically, this entails its contrapositive
 \getittotheright{\math
 {\neg\forall x.\, A\nottight{\nottight\implies}
  \neg A\{x\mapsto\varepsilon x.\,\neg A\}}}\\
 and then, \eg\ by the trivial direction of (\math{\varepsilon_1})
 (when \math A is replaced with \math{\neg A})
 \\\noindent\majorheadroom\majorfootroom\LINEmath
 {\neg\forall x.\, A\nottight{\nottight\implies}
  \exists x.\,\neg A~~~~~~~}(Q2)\\\noindent
 which is not valid in intuitionistic logic in general.
 Thus, the universal quantifier in \hilbert's intended object 
 logic---if it includes \nolinebreak
 (\math{\varepsilon_2}) or anything similar for the
 universal quantifier (such as \hilbert's \math\tau-operator, \cf\
 \cite{logischegrundlagen})---is 
 strictly weaker than in intuitionistic logic.
 More precisely,
 adding 
 \\\noindent\majorfootroom\LINEmath
 {\forall x.\, A
  \nottight{\nottight\antiimplies }
  A\{x\mapsto\tau x.\, A\}
  }(\math{\tau_0})\\\noindent
 is equivalent on the \math\tau-free theory to adding \bigmaths
 {\exists x\stopq\inpit{
  \forall y.\,A
  \antiimplies 
  A\{y\tight\mapsto x\}
 }}{}  with \math x not occurring in \nlbmath A, 
 which again implies (Q2), \cfnlb\ \cite[\litsectref{3.4.2}]{meyervioldiss}.
 \par
 From a semantical view, \cf\ \cite{intuitionisticsemantics},
 the intuitionistic \nlbmath\forall\ may be eliminated, however,
 by first applying the \goedel\ translation 
 into the modal logic S4 with classical \nlbmath\forall\
 and \nlbmath\neg,
 \cf\ \eg\ \cite{fittingquantifiedmodal},
 and then adding the \nlbmath\varepsilon\ conservatively, \eg\ 
 by avoiding substitutions via \math\lambda-abstraction as in 
 \nolinebreak\cite{fittingepsilon}.\vfill\pagebreak}
or stronger such as set theories with axiom schemes over arbitrary terms
including the \nlbmath\varepsilon,
\cf\ \sectref{section choice functions}.
%logics where only finite structures are considered.\footnote
%{\Cf\ \cite{logicofchoice} for references}
Moreover, even in standard \firstorder\ logic there is no 
translation from the formulas containing the \math\varepsilon\
to formulas not containing it.

\subsection{Our Objective}\label{section our objective}
While the historical and technical research
on the \math\varepsilon-theorems is still going on
and the method of \mbox{\math\varepsilon-elimination}
and \math\varepsilon-substitution
did not die with \hilbertsprogram, 
this is not our subject here.
We \nolinebreak are \nolinebreak less interested in \hilbertsprogram\
and the consistency of mathematics than
in the powerful use of logic in creative processes.
And, instead of the tedious syntactical proof transformations,
which easily lose their usefulness and elegance 
within their technical complexity and
which---more importantly---can only refer to an already existing logic,
we look for\emph{semantical} 
means for finding new logics and new applications.
And the question that still has to be answered in this field
%especially for classical logic 
is: {\em
What would be a proper semantics for \hilbert's \nlbmath\varepsilon?}

\subsection{Indefinite Choice}
\label{section indefinite choice}
Just as the \math\iota-symbol is usually taken to be the 
referential interpretation of the\emph{definite} articles in natural languages, 
it is our opinion that
the \math\varepsilon-symbol should be that of the\emph{indefinite} 
determiners (articles and pronouns) such as ``a(n)'' or 
``some\closequotefullstop

\begin{example}
[\math\varepsilon\ instead of \math\iota, 
 part\,II]\hfill{\em (continuing \examref{example iota})}\label
{example Pope better}
\par\noindent
It may well be the case that
\par\noindent\LINEmath{
  \HG
  \nottight{\nottight =}
  \varepsilon x.\,\Fatherpp x{\ident{Jesus}}
  \nottight{\nottight\und}
  \ident{Joseph}
  \nottight{\nottight =}
  \varepsilon x.\,\Fatherpp x{\ident{Jesus}}
}\par\noindent
\ie\ that
{``The Holy Ghost is \emph{\underline a} \ father of Jesus 
       and Joseph is \emph{\underline a} \ father of Jesus.''}
But this does not bring us into trouble with the Pope because we do not know
whether all fathers of Jesus are equal.
This will become clearer when we reconsider this example in 
\examref{example Canossa}.\end{example}

\yestop\noindent
Philosophy of language will be further discussed in 
\sectref{section philosophy of language}.

\yestop\yestop
\subsection{Committed Choice}\label{section committed choice}
Closely connected to indefinite choice 
(also called ``indeterminism'' or ``don't care nondeterminism'') 
is \nolinebreak the notion of ``\tightemph{committed choice}\closequotefullstop
For example, when we have a new telephone, we typically\emph{don't care} 
which number we get, but once the provider has chosen a number for our
telephone, we want them to\emph{commit to this choice},
\ie\ not to change our phone number between two incoming calls.

\begin{example}[Committed Choice]\hfill{\em(Buggy!)}\label
{example committed choice}\\\noindent\math{\begin
{array}{@{}l r@{\,\,}c@{\,\,}l@{}}
  \mbox{Suppose we want to prove}
 &\exists x.\,(x
 &\not= 
 &x)
\\\mbox{According to (\math{\varepsilon_1}) from \sectref
{section proof-theoretic origin} this reduces to}
 &\varepsilon x.\,\inpit{x\tightnotequal x}
 &\not= 
 &\varepsilon x.\,\inpit{x\tightnotequal x}
\\\mbox
  {Since there is no solution to \math{x\tightnotequal x} we can replace}
\\\varepsilon x.\,\inpit{x\tightnotequal x}\mbox
  { with anything. \ Thus, the above reduces to}
 &\zeropp
 &\not= 
 &\varepsilon x.\,\inpit{x\tightnotequal x}
\\\mbox{and then, by exactly the same argumentation, to}
 &\zeropp
 &\not= 
 &\onepp
\\\mbox{which is valid.}
\\\end{array}}\end{example}

\noindent
Thus we have proved our original formula 
\bigmaths{\exists x.\,\inpit{x\not= x}},
which, however, happens to be invalid. What went wrong?
Of course, we have to commit to our choice for all occurrences of 
the \math\varepsilon-term
introduced when eliminating the existential quantifier:
If we choose \nlbmath\zeropp\ on the left-hand side, 
we have to commit to the choice of
\nlbmath\zeropp\ on the right-hand side, too.
\vfill\pagebreak

\section
{Semantics for \hilbert's \math\varepsilon\ in the Literature}\label
{section in the literature}
In this \sectref{section in the literature}, 
we review the literature on the \nlbmath\varepsilon's
semantics with a an emphasis on practical adequacy and \hilbert's intentions.

\subsection{Right-Unique Semantics}\label{section right-unique semantics}
In contrast to the indefiniteness we suggested in 
\sectref{section indefinite choice},
in the literature 
nearly all semantics for \hilbert's \nlbmath\varepsilon-operator
are functional, \ie\emph{\opt{right-} unique}; \ 
\cfnlb\ \cite{leisenring} and the references there.

\subsubsection{\ackermann's (II,4) \ = \ \bourbaki's (S7) \ = \  
\leisenring's (E2)}\label{section E2}
In \cite{ackermann-mengentheoretische-Begruendung} under the label (II,4),
in \cite{bourbaki} 
under the label \nolinebreak(S7)
(where a \nlbmath\tau\ is written for the \nlbmath\varepsilon,
 which must not be confused with \hilbert's \math\tau-operator, \cf\
 \noteref{footnote tau}), 
 and in \cite{leisenring} under the label
 \nolinebreak(E2), we find the following axiom scheme:
 \par\noindent\phantom{(E2)}\LINEmaths{
  \headroom\footroom
  \forall x\stopq\inparenthesesinlinetight{A_0\tightequivalent A_1}
  \nottight{\nottight\implies}
  \varepsilon x.\, A_0
  =
  \varepsilon x.\, A_1
}{}(E2)\par\noindent
Contrary to our version (E2\math') in \lemmref{lemma where E2' is}
of \sectref{section where E2' is},
in the standard framework the axiom \nolinebreak (E2) 
imposes a right-unique behavior for the \math\varepsilon-operator, 
which is based on the extension of the predicate. 

Axiom systems including (E2) are called\emph{extensional}
because---from a semantical point of view---the value of \nolinebreak
\bigmath{\varepsilon x.\, A} in each semantical structure \nlbmath\salgebra\
is functionally dependent on the extension of the formula \nlbmath A, 
\ \ie\nolinebreak\ on 
\bigmaths{
  \setwith
    o
    {\app
       {\EVAL{\salgebra\uplus\{x\tight\mapsto o\}}}
       A}},
where `\EVALSYM' is the standard evaluation function that
maps a structure \mbox{(or algebra}, interpretation) 
(including a valuation of the free variables) 
to \nolinebreak a \nolinebreak function mapping terms 
and formulas to values.

To get more freedom for the definition of a semantics of 
the \nlbmath\varepsilon, \ in \cite{meyervioldiss} and in
\cite{ahrendtgiese} the value of 
\bigmaths{\varepsilon x.\, A}{} 
may additionally depend on the syntax besides the semantics.\footnote
 {Besides the already mentioned extensional treatment of \math\varepsilon, 
 in \cite{ahrendtgiese} we also find an\emph{intentional} treatment 
 (which, roughly speaking, 
  results from requiring the axiom \nolinebreak (\nlbmath{\varepsilon_0}))
 and a\emph{substitutive} treatment where also
 the validity of the Substitution \opt{Value} Lemma
 for \math\varepsilon-terms is required: 
 \\\noindent\LINEmath{\headroom\footroom
   \displaystyle
   \displayapp
     {\displayapp\EVALSYM\salgebra}
     {\inpit{\varepsilon x\stopq A}\{\freevari y{}\tight\mapsto t\}}
  =\displayapp
     {\displayapp
        \EVALSYM
        {\salgebra\uplus\{\freevari y{}\mapsto\app{\EVAL\salgebra}t\}}}
     {\varepsilon x\stopq A}
 }\\\noindent
 Here \math x is a bound and \freevari y{} is a free variable.
 Since logics where the Substitution Lemma for \math\varepsilon-free 
 formulas does not hold are not considered
 (such as the \firstorder\ modal logic of 
  \nolinebreak\cite{fittingquantifiedmodal}),
 in \cite{ahrendtgiese} we find a theorem basically
 saying that every extensional structure is substitutive.} \ 
It \nolinebreak is \nolinebreak 
then given as a function depending on a semantical
structure and on 
the syntactical details of the term \ \nlbmath{\varepsilon x.\, A}.\hskip1.2em%
We read:
\begin{quote}
``This definition contains no restriction whatsoever on the valuation of
  \math\varepsilon-terms.''\getittotheright{\cite[\p 177]{ahrendtgiese}}
\end{quote}
This is, however, not true because it imposes the restriction of a 
right-unique behavior, which denies the possibility of an indefinite
behavior, as we will see below.

Note that (E2)
has a disastrous effect in intuitionistic logic. \ 
This is already the case for its proper consequence 
\newcommand\theevilformulapartone
{\varepsilon x.\, A_0
 \neq
 \varepsilon x.\, A_1}
\newcommand\theevilformulaparttwo
{\neg\inpit{
 \forall x.\,A_0\und
 \forall x.\,A_1}}
\newcommand\theevilformula
{\theevilformulapartone\nottight{\implies}\theevilformulaparttwo}
\ \ \bigmaths{\theevilformula}{} \ \ 
which---together with \nolinebreak(\math{\varepsilon_0}) and say 
``\math{0\tightnotequal 1}''---turns every classical 
validity into an intuitionistic one.\footnote
{\label{note bell}\\{\bf(\bigmath{0\tightnotequal 1\comma
 \theevilformula
 \nottight{\nottight\yields}
 B\oder\neg B} in intuitionistic logic)}
 \par\noindent
 For the proof of the weaker 
 \bigmaths{0\tightnotequal 1\comma
 \mbox{\rm(E2)}\nottight{\nottight\yields}B\oder\neg B}{}
 for any formula \math B,
 \cf\ already \cite[Proof of \littheoref{6.4}]{logicaloptions}, \ 
 which already occurs in more detail in 
 \cite[\litsectref{3}]{bellclassicalepsilon}, \ 
 and is sketched in
 \cite[\litsectref{7}]{bellintuitionisticepsilon}.
 \par
 Let \math B be an arbitrary formula. \ 
 We are going to show 
 that \bigmaths{\yields B\oder\neg B}{} holds 
 in intuitionistic logic under the assumptions of
 reflexivity, symmetry, 
 and transitivity of ``\nlbmath=\closequotecommaextraspace
 the \math\varepsilon-formula (or \nolinebreak(\math{\varepsilon_0})), \ 
 and of the formulas \bigmaths{0\tightnotequal 1}{} and 
 \bigmaths\theevilformula.
 \par
 Let \math x be a variable not occurring in \math B. \ \ 
 Set \bigmaths{A_i:=\inpit{B\oder x\tightequal i}}. \
 \par
 Now what we have to show is a trivial consequence of
 the following Claims~1~and~2, \bigmaths{\yields\theevilformula},
 and Claim\,3.
 \initial{\underline{Claim\,1:}} 
 \bigmaths
 {0\tightequal 0\comma
  1\tightequal 1\comma
  (\mbox{\math\varepsilon-formula})\{A\tight\mapsto A_0,t\tight\mapsto 0\}\comma
  (\mbox{\math\varepsilon-formula})\{A\tight\mapsto A_1,t\tight\mapsto 1\}}{}\\
  \bigmaths{\yields\ \ 
  B\oder\inpit{\varepsilon x.\,A_0\tightequal 0\und
 \varepsilon x.\,A_1\tightequal 1}}.
 \initial{\underline{Claim\,2:}}
 \bigmaths{{\varepsilon x.\,A_0\tightequal 0\und
 \varepsilon x.\,A_1\tightequal 1}\comma
 0\tightnotequal 1\comma
 \forall x,y,z\stopq\inpit
 {y\tightequal x\und y\tightequal z\implies x\tightequal z}}{}\\
 \bigmaths{\yields\ \ \theevilformulapartone}.
 \par\noindent\underline{Claim\,3:} \
 \bigmaths{\theevilformulaparttwo\nottight{\nottight\yields}\neg B}.
 \par\noindent\underline{Proof of Claim\,1:} \ 
 From the \math\varepsilon-formula and 
 reflexivity of ``\nlbmath=\closequotecommaextraspace 
 we get \bigmaths{\yields A_i\{x\tight\mapsto\varepsilon x.\,A_i\}}.
 Thus, \bigmaths{\yields A_0\{x\tight\mapsto\varepsilon x.\,A_0\}
 \und A_1\{x\tight\mapsto\varepsilon x.\,A_1\}}. \
 From this, Claim\,1 follows by distributivity.
 \QED{Claim\,1}
 \par\noindent\underline{Proof of Claim\,2:} \ Trivial.
 \QED{Claim\,2}
 \par\noindent\underline{Proof of Claim\,3:} \ 
 As \math x does not occur in \math B,
 we get \bigmaths{B\yields\forall x.\,A_i}.
 The rest is trivial.
 \QED{Claim\,3}\vfill\pagebreak} \
For the strong consequences  
of the \math\varepsilon-formula in intuitionistic logic,
\cf\ our \noteref{footnote tau}.
\vfill\pagebreak

\subsubsection{Roots of the Right-Uniqueness Requirement}\label
{section root citations}
The omnipresence of the right-uniqueness requirement
may have its historical justification in the fact that 
if \nolinebreak we expand the dots 
\nolinebreak``\ldots'' in the
quotation preceding \examref{example epsilon instead of iota 1}
in \sectref{section epsilon}, the full quotation reads:
\begin{quote}{\fraknomath``\germantextfour''}\getittotheright
{\cite[\Vol\,II, \p 12, modernized orthography]{grundlagen}}\end
{quote}\begin{quote}
``\englishtextfour\ \englishtextone'' 
\getittotheright{(our translation)}\\
\getittotheright{(``Syntactically''
  may be replaced with ``Structurally'')}
\end{quote}
Here the word ``function'' could be understood in its mathematical sense 
to denote a (right-) unique relation.
And, what kind of function could it be but a choice function,
choosing an element from the set of objects that satisfy \nlbmath A\,? \ 
Accordingly, at a different place, we read:
\begin{quote}{\fraknomath``\germantextthree''}\getittotheright
{\cite[\p\,68]{grundlagenvortrag}}\end{quote}\begin{quote}
``Beyond that, the \math\varepsilon\ has the \role\ of the choice function,
\ie\ in the case where \bigmaths{A\,a}{} may hold for several objects,
\bigmath{\varepsilon\,A} is\emph{an arbitrary one} of the 
objects \nlbmath a for which \bigmath{A\,a} holds.''
\getittotheright{(our translation)}\\
\getittotheright{(in more modern notation, we would possibly write}\\
\getittotheright{``\app A a''
 for ``\maths{A\,a}{}'' and ``\math{\varepsilon x.\inpit{\app A x}}''
 for ``\maths{\varepsilon\,A}{}'')}
\end{quote}
\vfill\pagebreak

\subsubsection{Universal and Generalized Choice Functions}\label
{section choice functions}
Since---in \cite[one but last paragraph]{logischegrundlagen}---\hilbertname\ 
himself seems to have confused the consequences
of the \math\varepsilon\ on the Axiom of Choice 
(\cfnlb\ \cite{axiomofchoice}, \cite{weakaxiomofchoice}),
we point out: Although the \nlbmath\varepsilon\
supplies us with a syntactical means for expressing a\emph
{universal choice function},
the axioms 
(E2), (\math{\varepsilon_0}), (\math{\varepsilon_1}), and (\math{\varepsilon_2})
do not imply the Axiom of Choice in set theories,
%neither in set theory or in higher-order logic,
unless the axiom schemes of Replacement (Collection) and Comprehension 
(Separation, Subset)
also range over expressions containing the \nlbmath\varepsilon; \ 
\cfnlb\ \cite{leisenring}, \nolinebreak\litsectref{IV\,4.4}.

Moreover, to be precise, 
the notion of a ``choice function'' must be generalized here 
because we need a\emph{total} function on the power set of any (non-empty)
universe. Thus, a \nolinebreak
value must be supplied even at the empty set: \ 
\math f is defined to be a\emph{generalized choice function} \udiff\ 
\bigmath{\FUNDEF f{\DOM f}{\bigcup\inpit{\DOM f}}}
and \bigmaths{
  \forall x\tightin\DOM f\stopq
  \inpit{x\tightequal\emptyset\nottight\oder\app f x\tightin x}}.

\yestop
\subsubsection{\hermesname' (\math{\varepsilon_5}) and \devidiname's (vext)}
In \cite[\p 18]{hermesepsilon}, the \math\varepsilon\
suffers from some overspecification in addition to\nolinebreak\ (E2): 
\par\noindent\phantom{(\math{\varepsilon_5})}\LINEmaths{
  \headroom\footroom
  \varepsilon x.\,\falsepp
  \ =\ 
  \varepsilon x.\,\truepp
}{}(\math{\varepsilon_5})\par\noindent
This sets
the value of the generalized choice function \nlbmath f at the empty set to 
the value of \nlbmath f at the whole universe.
For classical logic, we can combine (E2) and (\math{\varepsilon_5})
into the following axiom of \cite{vext} for ``\underline very 
\underline{ext}ensional'' semantics:
\par\noindent\phantom{(vext)}\LINEmaths{
  \forall x\stopq\inparenthesesoplist{
  \inpit{\exists y.\,A_0\{x\tight\mapsto y\}\nottight\implies A_0}
  \oplistequivalent
  \inpit{\exists y.\,A_1\{x\tight\mapsto y\}\nottight\implies A_1}}
  \nottight{\nottight\implies}
  \varepsilon x.\, A_0
  =
  \varepsilon x.\, A_1
}{}(vext)\par\noindent
Indeed, (vext) implies (E2) and (\math{\varepsilon_5}). \ 
The other direction, however,
does not hold for intuitionistic logic, where, roughly speaking, (vext)
additionally implies 
that if the same elements make \nlbmath{A_0} and \nlbmath{A_1}
as true as possible, then the \math\varepsilon-operator picks the same 
element of this set, even if the suprema \bigmath
{\exists y.\,A_0\{x\tight\mapsto y\}} and \bigmath
{\exists y.\,A_1\{x\tight\mapsto y\}} 
(in the complete \heyting\ algebra) are not equally true.

\yestop
\subsubsection{Completeness Aspirations of \leisenring\ and \asser}
Different possible choices for the value of the generalized choice function
\nlbmath f at the empty set
are discussed in \cite{leisenring},
but as the consequences of any special choice are quite queer,
the only solution that is found to be sufficiently adequate in \cite{leisenring}
is to consider validity in\emph{any} 
model given by%\emph{each} choice from the empty set and
\emph{each} generalized choice function on the power set of the universe.
Notice, however, that even in this case, in each single model, 
the value of \ \nlbmath{\varepsilon x.\,A} \ is still\emph{functionally}
dependent on the extension of \nlbmath A. \ 
Roughly speaking,  in \cite{leisenring} the axioms
(\math{\varepsilon_1}), and (\math{\varepsilon_2})
from \sectref{section proof-theoretic origin} 
and (E2) from \sectref{section E2}
are shown to be complete \wrt\ this semantics of 
the \nlbmath\varepsilon\ in \firstorder\ logic.

This completeness makes it unlikely that this semantics
exactly matches \hilbert's intentions:
Indeed, if \hilbert's intended semantics for the \nlbmath\varepsilon\
could be completely captured by adding the single and straightforward
axiom \nolinebreak (E2), this axiom would not have been omitted in 
\cite{grundlagen}. \ 
It \nolinebreak is my opinion that the reason for this omission is that 
\hilbert's intentions for the \nlbmath\varepsilon\
were not right-unique but indefinite:
If \hilbert\ had intended a right-unique behavior, 
it would not be impossible to derive (E2) from his axiomatization!

\begin{sloppypar}
Completeness---detached from practical usefulness, but 
the theoreticians' favorite puzzle---has misled others, too:
In \cite{asserepsilon} the objective is to find a semantics
such that \hilbert's \math\varepsilon-calculus of \cite{grundlagen} 
is sound and\emph{complete} for it.
This semantics, however, 
has to depend on the details of the syntactical form 
of the \math\varepsilon-terms and, moreover, 
turns out to be necessarily so 
artificial that in \cite{asserepsilon} the author himself 
does not recommend it and admits 
not to believe that \hilbert\ could have intended \nolinebreak it:
\end{sloppypar}

\yestop\halftop\begin{quote}
{\fraknomath``\germantexteight''}\getittotheright
{\cite[\p\,59, modernized orthography]{asserepsilon}}
\end{quote}\begin{quote}
``This notion of a choice function, however,'' 
  (\ie\ the type-3 choice function, 
   providing a semantics for the \math\varepsilon-operator) 
``is so intricate that its application in informal mathematics is hardly to be
  recommended.'' 
\getittotheright{(our translation)}\\\getittotheright
{(``informal mathematics'' may be replaced with 
 ``intuitive mathematics\closequotecomma}
\\\getittotheright{
 ``\naive\ mathematics\closequotecomma
 or ``mathematics with semantical contents'')}
\yestop\yestop\end{quote}\begin{quote}
{\fraknomath``\germantextnine''}\getittotheright
{\cite[\p\,65, modernized orthography]{asserepsilon}}
\end{quote}\begin{quote}
``The intricacy of the notion of the type-3 choice function puts up the question
  whether the intention in \cite{grundlagen} ('' \ldots\ ``) really was
  to describe this notion axiomatically. 
  I believe I can draw from the presentation in \cite{grundlagen} 
  that that is not the case,'' \getittotheright{(our translation)}
\yestop\yestop\end{quote}

\subsubsection{My Assumption on \hilbert's Intentions}\label
{section my assumption}
The statements of \bernays\ and \hilbert\ in German language cited 
in \sectref{section root citations}
are ambiguous with respect to the question of an 
intended (right-) unique behavior of the \nlbmath\varepsilon-operator. \ 
\hilbert\ probably
wanted to have what today we call 
``\tightemph{committed choice}\closequotecomma
but simply used the word ``function'' for the following three reasons:
\hilbert\ was not too much interested in semantics anyway. 
The technical term ``committed choice'' did not exist at \hilbert's time. 
Last but not least,
right-uniqueness conveniently serves as a global commitment to any choice
and thereby avoids the problem illustrated 
in \examref{example committed choice} of \sectref{section committed choice}. 

But the price we would have to pay for such an overspecification is high:
Right-Uniqueness restricts 
operationalization (\cf\ 
\sectref{section Instantiating Strong Free Universal Variables})
and 
applicability:
\Cfnlb\ \eg\ \cite{geurts-one} and our 
\sectref{section geurts}
for the price of right-uniqueness
in capturing the semantics of sentences in natural language.

{\em And what we are going to show in this \daspaper\ 
is that there is no reason to pay that price!}
\vfill\pagebreak

\subsection{Indefinite Semantics in the Literature}\label
{subsection indefinite}
The only occurrences of an indefinite semantics for 
\hilbert's \nlbmath\varepsilon\ 
in the literature seem to be \cite{logicofchoice}
and the references there.
% In \cite{logicofchoice}
% \nlbmath\delta\ is written instead of \nlbmath\varepsilon\
% because the right-unique behavior is considered to be 
% essential for the \nlbmath\varepsilon. \ 

Consider the 
formula
\ \ \math{
  \varepsilon x.\,\inpit{x\tightequal x}
  \nottight{\nottight=}
  \varepsilon x.\,\inpit{x\tightequal x}
} \ \ 
from \cite{logicofchoice} or the even simpler
\par\yestop\noindent\phantom{({\sc Reflex})}\LINEmath{
  \varepsilon x.\,\truepp
  \nottight{\nottight=}
  \varepsilon x.\,\truepp
}({\sc Reflex})\par\yestop\noindent
which may be valid or not, depending on the question whether
the same object is taken on both sides of the equation or not.
In natural language this like 
\par\halftop\noindent\LINEnomath{
``Something is equal to something.''
}\par\halftop\noindent
whose truth is indefinite. If you do not think so, consider
\ \ \math{
  \varepsilon x.\,\truepp
  \nottight{\nottight{\not=}}
  \varepsilon x.\,\truepp
} \ \ 
in addition, \ \ie\
{``Something is unequal to something.''},
and notice that the two sentences seem to be contradictory.

In \cite{logicofchoice}, \kleene's strong three-valued logic is 
taken as a mathematically elegant means to solve the problems with
indefiniteness. 
In spite of the theoretical significance of this solution, however, 
from a practical point of view, 
\kleene's strong three-valued logic severely restricts its applicability.
In applications, 
a \nolinebreak logic is not an object of investigation but a meta-logical tool, 
and logical arguments are never made explicit
because the presence of logic is either 
not realized at all or taken to be trivial,
even by academics (unless they are formalists),
\cf\ \eg\ \cite[\p 14\f]{dialogprojectantrag}, for Wizard of Oz studies 
with young students.
Thus, regarding applications,
we have to stick to 
our common meta-logic, 
which in the western world is a subset of (modal) classical logic.
A western court may accept that 
Lee Harvey Oswald killed John F. Kennedy as well as that he did not;
but cannot accept a third possibility, a\emph{tertium}, as required for 
\kleene's strong three-valued logic, and especially not the interpretation
given in \cite{logicofchoice} that he\emph{both} did and did not kill him,
which directly contradicts any common sense.
% Note, however, that the treatment of \cite{logicofchoice} is fine 
% from a mathematical point of
% view and also Buddha with his four-valued logic might be happy with it.
\vfill\pagebreak

\section
[Introduction to Our Novel Indefinite Free-Variable Semantics]
{Introduction to Our Novel Indefinite Free-Variable\\Semantics}
\label{section new indefinite}

\subsection{Free \math\gamma- and Free \math{\delta}-Variables}\label
{section free}

Before we can introduce to our treatment of the \math\varepsilon,
we have to 
provide some technical background. \ 
\Cfnlb\ \cite{wirthcardinal} for a technically more detailed introduction.

In this \sectref{section free}, 
we will introduce free \math\gamma-, \deltaminus-,
and \deltaplus-variables. 
Free variables frequently occur in mathematical practice. 
Their logical function varies locally.
It is typically determined implicitly by the
context and the obviously intended semantics. 

In this \daspaper, however, we make this function explicit by 
using disjoint sets of variable-symbols for different functions.
The classification of a free variable is indicated by adjoining the
respective \math\gamma, \deltaminus, or \deltaplus\ to the upper
right of the symbol for the variable. 

As already noted in 
\cite[\p 155]{mathematicalphilosophy}, in mathematical practice,
the free variables \freevari a{} and \nlbmath{\freevari b{}} 
in the (quasi-) formula
\newcommand\writedeltaexample[3]
{\inpit{#1+#2}^2\nottight{\nottight{\nottight=}}
 \inpit{#1}^2+2\,#1\,#2+\inpit{#2}^2}%
\par\noindent\LINEmaths{
\mbox{}\ \ \ \writedeltaexample{\freevari a{}}{\freevari b{}}{\inpit}
}{}\par\noindent
obviously have a universal intention and the quasi-formula itself
is not meant to denote a propositional function but actually stands for
the closed formula 
\par\noindent\LINEmaths{
  \mbox{}\ \ 
  \forall a,b\stopq\inparentheses{\writedeltaexample{a}{b}{}}
  \quad\quad\quad\quad
  \mbox{}
}{}\par\noindent
In this \daspaper, however, we indicate by 
\par\noindent\LINEmaths{
  \writedeltaexample{\wforallvari a{}}{\wforallvari b{}}{\inpit}
}{}\par\noindent
a proper formula with \wfuv s, which---independently of its context---is 
logically equivalent to the universally quantified formula.

\yestop\noindent
Changing from universal to existential intention,
it is somehow clear that the linear system%
\newcommand\writegammaexample[4]{\mbox{}\hfill\maths{#4
  {\left(\begin{array}[c]{@{}c c@{}}2&3\\5&7\\\end{array}\right)
   \left(\begin{array}[c]{@{}c@{}}#1\\#2\\\end{array}\right)
  =\left(\begin{array}[c]{@{}c@{}}11\\13\\\end{array}\right)}
}{}\hspace{#3}\hspace{15em}\mbox{}\par\noindent\ignorespaces}%
\par\noindent\writegammaexample{\freevari x{}}{\freevari y{}}{1.3em}{}
asks us to find solutions for 
\math{\freevari x{}} and \nlbmath{\freevari y{}}\@.
We make this intention syntactically explicit by writing 
\\\writegammaexample{\existsvari x{}}{\existsvari y{}}{1.3em}{} instead.
This formula with \fev s is not only logically equivalent to
\par\noindent\writegammaexample x y{0em}{\exists x,y\stopq\inparentheses}
but may additionally enable us to retrieve the solutions for
\existsvari x{} and \existsvari y{} as the substitutions for
\existsvari x{} and \existsvari y{} chosen in a formal proof.

Finally, the \sfuv s are to represent our \math\varepsilon-terms in the end.
The names \math\gamma, \deltaminus, and \deltaplus\ refer to
the classification of reductive inference rules into \math\alpha-, \math\beta-, 
\math\gamma-, and \math\delta-rules of\nolinebreak\ \cite{smullyan},
as used in the following \sectref{subsection Rules}.
\vfill\pagebreak

\subsection{\math\gamma- and \math\delta-Rules}\label
{subsection Rules}
Suppose we want to prove the existential property 
\bigmaths{\exists x.\, A}.
The \math\gamma-rules of old-fashioned inference systems 
(such as \cite{gentzen} or \cite{smullyan}, \eg)
require us to choose a\emph{fixed} witnessing term \nlbmath t
as a substitute for the bound variable\emph{immediately}
when eliminating the quantifier.

Let \math{A} be a formula.
 We do not permit binding of variables that already occur 
 bound in a term or formula;\label{section binding restriction} 
 that is: \bigmath{\forall x\stopq A} is only a formula 
 if no binder on \math x already occurs in \nlbmath A.
 % \mbox{``\math{\forall x\stopq}''}, 
 % \nolinebreak\mbox{``\math{\exists x\stopq}''}, 
 % \nolinebreak\mbox{``\math{\lambda x\stopq}''}, 
 % \nolinebreak\mbox{``\math{\varepsilon x\stopq}''}.
 The simple effect is that our formulas are easier to read
 and
 our \math\gamma- \nolinebreak 
 and \math\delta-rules can replace\emph{all} occurrences of \nlbmath x.
 Moreover, we assume that all binders have 
 minimal scope, \eg\ \bigmath{\forall x,y\stopq A\und B} reads \ \mbox
 {\math{\inpit{\forall x\stopq\forall y\stopq A}\und B}}. \ \ 
Let
\math{\Gamma} and \math{\Pi} 
be\emph{sequents}, \ie\ 
disjunctive lists of formulas.
\par\halftop\noindent{\bf\math\gamma-rules: }
Let \math{t} be any term:
\nopagebreak\par\noindent
\strongexpansionrule
{\Gamma~~~\exists x.A~~~\Pi}
{A\{x\tight\mapsto t\}~~~\Gamma~~~\exists x.A~~~\Pi}
{}{}{}{}
\strongexpansionrule
{\Gamma~~~\neg\forall x.A~~~\Pi}
{\majorheadroom\overline{\,A\{x\tight\mapsto t\}\,}~~~\Gamma
 ~~~\neg\forall x.A~~~\Pi}
{}{}{}{}
\par\noindent Note that \overline A is 
the\emph{conjugate} of the formula \nlbmath A, \ 
\ie\ \nlbmath B if \nlbmath A is of the form\ \nlbmath{\neg B}, \ 
and \nlbmath{\neg A} otherwise.
Moreover,
in the good old days when trees grew upwards, 
\gentzenname\ \gentzenlifetime\
would have inverted the inference rules 
such that passing the line means consequence.
In \nolinebreak our case, passing the line means reduction,
and trees grow downwards.

More modern inference systems, however, (such as the ones in \cite{fitting})
enable us to delay the crucial choice of the term \nlbmath t 
until the state of the proof attempt may provide more information 
to make a successful decision.
This delay is achieved by introducing a special kind of variable,
called 
``dummy'' in \cite{prawitzimproved}, % and \cite{kanger},  
``free'' in \cite{fitting} and 
  in \litfootref{11} of \cite{prawitzimproved},
and ``meta'' in the field of planning and constraint solving.
We call these variables\emph{\fev s}
and write them like \nlbmath{\existsvari x{}}. \ 
When these additional variables are available, we can reduce 
\ \mbox{\math{\exists x.\, A}} \ first to
\bigmath{A\{x\tight\mapsto\existsvari x{}\}} 
and then sometime 
later in the proof we may globally substitute \nlbmath{\existsvari x{}}
with an appropriate term.

The addition of the \fev s changes the notion of a term but not
the \math\gamma-rules, whereas it \nolinebreak becomes visible in the 
\math\delta-rules. \ 
\math\delta-rules introduce \fuv s. \ 
The \fuv s are also called ``parameters'' or ``eigenvariables''
and typically stand for arbitrary objects of which nothing is known. \ 
Now the occurrence of such a \fuv\ must be disallowed 
in the terms that may be substituted for those \fev s
which have already been in use when an application of a \mbox{\math\delta-rule}
introduced this \fuv. \ 
The reason for this restriction of substitution for \fev s is
that the dependence or scoping of the quantifiers must somehow
be reflected in a dependence of the free variables.
This dependence is to be captured in a binary relation on the free variables,
called\emph\vc.

Indeed, 
it is sometimes unsound to instantiate a \fev\ \existsvari x{} 
with a term containing a \fuv\ \forallvari y{}
that was introduced later than \existsvari x{}:
\vfill\pagebreak

\begin{example}
\newcommand\outdent{\hspace{15em}\mbox{}}
The formula \hfill\math{
  \exists x\stopq\forall y\stopq\inpit{x\tightequal y}
}\outdent\\\noindent
is not generally valid.
We can start a proof attempt as follows:
\\\indent\math\gamma-step:
\hfill\math{
  \forall y\stopq
  \inpit{\existsvari x{}\tightequal y}
  \comma~~
  \exists x\stopq\forall y\stopq\inpit{x\tightequal y}
}\outdent\\\indent\math\delta-step:
\hfill\math{
  \inpit{\existsvari x{}\tightequal\forallvari y{}}
  \comma~~
  \exists x\stopq\forall y\stopq\inpit{x\tightequal y}
}\outdent\\\noindent
Now, if the \fev\ \existsvari x{} could be substituted by
the \fuv\ \nolinebreak\forallvari y{}, we would get the tautology
\nolinebreak
\bigmath{
  \inpit{
    \forallvari y{}\tightequal\forallvari y{}
  }
,}
\ie\ we would have proved an invalid formula. \ 
To \nolinebreak prevent this, the \math\delta-step has
to record \nlbmath{(\existsvari x{},\forallvari y{})}
in a \vc, where \pair{\existsvari x{}}{\forallvari y{}}
means that \existsvari x{} is somehow ``necessarily older'' 
than \forallvari y{},
so \nolinebreak that we must not instantiate the
\fev\ \existsvari x{} with a term containing the 
\fuv\ \nolinebreak\forallvari y{}.\end{example}

\yestop\noindent
Starting with an empty \vc, we extend the \vc\ during a proof
by \math\delta-steps and by steps that globally instantiate 
\math\gamma- and \deltaplus-variables. 
This kind of instantiation of\emph{rigid}
variables is only sound if the resulting \vc\ is still acyclic
after adding, for each free variable \nlbmath{\freevari x{}} instantiated with a
term \nlbmath{t} and for each free variable \nlbmath{\freevari z{}} occurring
in \nlbmath t, the pair \nlbmath{\pair{\freevari z{}}{\freevari x{}}}
to the \vc.

\yestop\yestop\noindent\label{section where delta rules are}%
To make things more complicated, there are basically two different versions
of the \math\delta-rules: standard \mbox{\deltaminus-rules}
(also \nolinebreak simply called ``\math\delta-rules'')
and \deltaplus-rules
(also called ``\tightemph{liberalized} \math\delta-rules'').
They differ in the kind of \fuv\ they introduce and---crucially---in 
the way they enlarge the \vc, depicted to the lower right of the bar:
\yestop
\begin{description}\halftop\item[\deltaminus-rules: ]
Let
\nlbmath{\wforallvari x{}}
be a new \wfuv:
\par\halftop\noindent
\strongexpansionrule
{\Gamma~~~\forall x.A~~~\Pi}
{A\{x\tight\mapsto\wforallvari x{}\}~~~\Gamma~~~\Pi}
{}
{\VARsomesall{\Gamma\ \forall x.A\ \Pi}\times\{\wforallvari x{}\}}
{}{}
\par\halftop\noindent
\strongexpansionrule
{\Gamma~~~\neg\exists x.A~~~\Pi}
{\majorheadroom
 \overline{\,A\{x\tight\mapsto\wforallvari x{}\}\,}~~~\Gamma~~~\Pi}
{}
{\VARsomesall{\Gamma\ \neg\exists x.A\ \Pi}\times\{\wforallvari x{}\}}
{}{}

\yestop\halftop\item[\deltaplus-rules: ]
Let 
\nlbmath{\sforallvari x{}}
be a new \sfuv:\nopagebreak\par\halftop\noindent
\strongexpansionrule
{\Gamma~~~\forall x.A~~~\Pi}
{A\{x\tight\mapsto\sforallvari x{}\}~~~\Gamma~~~\Pi}
{\{\pair
     {\sforallvari x{}}
     {\headroom\overline{\,A\{x\tight\mapsto\sforallvari x{}\}\,}}
\}}
{\VARfree{\forall x.A}
 \times
 \{\sforallvari x{}\}
}
{\revrelapp{\tight\leq}{\VARall A}
 \times
 \{\sforallvari x{}\}
}{}
\nopagebreak\par\yestop\noindent
\strongexpansionrule
{\Gamma~~~\neg\exists x.A~~~\Pi}
{\majorheadroom\overline{\,A\{x\tight\mapsto\sforallvari x{}\}\,}~~~\Gamma
 ~~~\Pi}
{\{\pair
     {\sforallvari x{}}
     {A\{x\tight\mapsto\sforallvari x{}\}}
\}}
{\VARfree{\neg\exists x.A}
 \times
 \{\sforallvari x{}\}
}
{\revrelapp{\tight\leq}{\VARall A}
 \times
 \{\sforallvari x{}\}
}{}\halftop\end{description}

\noindent
Notice that \bigmaths{\VARsomesall{\Gamma\ \forall x.A\ \Pi}}{}
denotes the set of the free \math\gamma- and \deltaplus-variables
occurring in the whole upper sequent, whereas
\nolinebreak\bigmaths{\VARfree{\forall x.A}}{} denotes the set of all free 
(\math\gamma-, \deltaminus-, \deltaplus-) variables,
but only the ones occurring in the\emph{principal formula}
\bigmaths{\forall x.A}. \ % or \bigmaths{\neg\exists x.A}, respectively. \ 
The smaller \vc s generated by the \mbox{\deltaplus-rules} 
mean more proofs. \ 
Indeed, the \mbox{\deltaplus-rules} enable additional proofs
on the same level of\emph{multiplicity} (\ie\nolinebreak\ the number of repeated
\math\gamma-steps applied to the identical principal formula); \ 
\cf\ \eg\ \cite[\litexamref{2.8}, \p\,21]{wirthcardinal}\@. \ 
For certain classes of theorems,
some of these proofs are exponentially and even 
non-elementarily shorter than the shortest proofs which
apply only \deltaminus-rules; \ 
for a survey \cf\ \cite[\litsectref{2.1.5}]{wirthcardinal}.
Moreover, the 
\mbox{\deltaplus-rules} provide additional proofs that are
not only shorter but also more natural 
and easier to find both automatically and for human beings; \ 
\cf\ the discussion on design goals for inference systems in
\cite[\litsectref{1.2.1}]{wirthcardinal},
and the proof of the limit theorem for \nlbmath + 
in \cite{nonpermut}\@. \ 
All \nolinebreak in all, the name ``liberalized'' for the \deltaplus-rules
is indeed justified: They provide more freedom to the prover.\footnote
{\label{note liberalized}\ 
 Regarding the classification of one of the \math\delta-rules 
 as ``liberalized\closequotecomma
 we could try to object that  \bigmaths{\VARfree A}{}
 is not necessarily a subset of 
 \bigmaths{\VARsomesall{\Gamma\ \forall x.A\ \Pi}},
 because it may include some additional \wfuv s.

 But the additional \wfuv s blocked by the \deltaplus-rules
 (as compared to the \mbox{\deltaminus-rules}) \ 
 do not block proofs in practice. \ 
 This has following reason: \ 
 With a reasonably minimal \vc\ \nlbmath R, \ 
 the only additional cycles that
 could occur are of the form \bigmaths
% {\existsvari y{}\nottight R\wforallvari z{}\nottight R\sforallvari x{}
%  \nottight R\freevari u{}\nottight{\refltransclosureinline R}\existsvari y{}}{}
% with \bigmaths{\existsvari y{},\wforallvari z{}\in\VAR{\Gamma\ \forall x.A\ \Pi}}; \ 
 {\rigidvari y{}\nottight R\wforallvari z{}\nottight R\sforallvari x{}
  \nottight{\transclosureinline R}\rigidvari y{}}{}
 with \bigmaths{\rigidvari y{},\wforallvari z{}
 \in\VAR{\Gamma\ \forall x.A\ \Pi}}; \ 
 unless we substitute something for \nlbmath{\sforallvari x{}}. \ \
 And in this case the corresponding \deltaminus-rule would result in 
 the cycle \bigmaths
% {\existsvari y{}\nottight R\wforallvari x{}\nottight R\freevari u{}
%  \nottight{\refltransclosureinline R}\existsvari y{}}{} anyway. \ 
 {\rigidvari y{}\nottight R\wforallvari x{}
  \nottight{\transclosureinline R}\rigidvari y{}}{} anyway. \ 

 Moreover, \deltaminus-rules and \wfuv s do not occur 
 in inference systems with \deltaplus-rules before \cite{wirthcardinal}, \ 
 so that in the earlier systems 
 \bigmaths{\VARfree A}{}
 is indeed a subset of 
 \bigmaths{\VARsomesall{\Gamma\ \forall x.A\ \Pi}}.}  

Moreover, note that the singleton sets indicated to the upper right of the
bar of the above 
\deltaplus-rules are to augment another global binary relation 
besides the \vc, namely a function called the\emph\cc.
This will be explained in \sectref{section replacing epsilon}\f\ \ 

There is a popular alternative to \vc s, namely Skolemization,
where the \fuv s become functions (\ie\ their order is incremented) 
and the \deltaminus-
and \deltaplus-rules
give them the \fev s of \bigmaths{\VARsome{\Gamma\ \forall x.A\ \Pi}}{}
and \bigmaths{\VARsome{\forall x.A}}, \resp, as initial arguments.
Then, the occur-check of unification implements the 
restrictions on substitution of \fev s.
In some inference systems, however,
Skolemization is unsound (\eg\ for higher-order systems 
such as the one in \cite{kohlhasetableauupdated} 
or the system in \cite{wirthcardinal} for\emph\descenteinfinie)
or inappropriate (\eg\ in the matrix systems of \cite{wallen}).
We prefer inference systems with \vc s 
as this is a simpler, more general, and not less efficient approach 
compared to Skolemizing inference systems.
Notice that \vc s do not add unnecessary complexity:
% to applications where Skolemization is no problem:
 Firstly, if \vc s are superfluous we can work with an empty \vc\ 
 as if there would be no \vc\ at all.
 Secondly, we will need the \vc s anyway for our \cc s,
 which again are needed to formalize our novel approach to 
 \hilbert's \math\varepsilon-operator.
\vfill

\subsection{Quantifier Elimination and Subordinate \math\varepsilon-terms}\label
{section quantifier elimination and subordinate}

Before we can introduce to our treatment of the \math\varepsilon,
we also have to get more acquainted with the \nlbmath\varepsilon\ in general.

The elimination of \math\forall- and \math\exists-quantifiers 
with the help of \math\varepsilon-terms 
(\cf\ \sectref{section proof-theoretic origin})
may be more difficult
than expected when some
\math\varepsilon-terms become ``subordinate'' to others.

\begin{definition}[Subordinate\halftop]\label{definition subordinate}\sloppy
An \math\varepsilon-term \nolinebreak\ \mbox{\math{\varepsilon v.\, B}} \ 
(or, more generally, 
 a binder on \nlbmath v together with its scope \nlbmath B) \ 
is\emph{superordinate} 
to an (occurrence of an) 
\mbox{\math\varepsilon-term \ \math{\varepsilon x.\, A}} \ 
\udiff\ 
\begin{enumerate}\item\noitem
\math{\varepsilon x.\, A} \ is a subterm of \math B \ 
% (\ie\ in the scope of \nlbmath v)
and \noitem\item
an occurrence of the variable \nlbmath v in \bigmath{\varepsilon x.\, A} 
is free in \nlbmath B \ 
\\(\ie\ the binder on \nlbmath v binds an occurrence of 
\nlbmath v in \ \nlbmath{\varepsilon x.\, A} ). \ 
\noitem\end{enumerate}
An (occurrence of an) \math\varepsilon-term \nlbmath a
is\emph{subordinate} to an \math\varepsilon-term \mbox{\math{\varepsilon v.\, B}} \ 
(or, more generally, 
 a binder on \nlbmath v together with its scope \nlbmath B) \ 
\udiff\ \ \mbox{\math{\varepsilon v.\, B}} is superordinate to 
\nlbmath a.\end{definition}
{In \cite[\Vol\,II, \p\,24]{grundlagen},
 these subordinate \math\varepsilon-terms,
 which are responsible for the difficulty to prove
 the \math\varepsilon-theorems  constructively,
 are called ``{\fraknomath\germantextfive}\closequotefullstop
 Note that we do not use a special name 
 for \math\varepsilon-terms with free occurrences of variables---such as 
 ``\math\varepsilon-Au\esi dr\ue cke''
 (``quasi \math\varepsilon-terms'')
 instead of ``\math\varepsilon-Terme''
 (``\math\varepsilon-terms'')---but simply call them 
 ``\math\varepsilon-terms\closequotecomma too.}
\vfill\pagebreak

\begin{example}[Quantifier Elimination 
and Subordinate \math\varepsilon-Terms]\label
{example subordinate}\\\noindent
Consider the formula 
\bigmaths{\forall x\stopq\exists y\stopq\forall z\stopq\Ppppdrei x y z}.
Let us apply (\math{\varepsilon_1}) 
and (\math{\varepsilon_2}) from \sectref{section proof-theoretic origin}
to remove the three quantifiers completely. 
We introduce the following abbreviations:
\par\noindent\mbox{}\hfill\math{\begin{array}[t]{l l l}
  \app{\app{z_a}x}y
 &=
 &\varepsilon z\stopq\neg\Ppppdrei x y z
\\
\\\app{y_a}x
 &=
 &\varepsilon y\stopq\Ppppdrei x y{\app{\app{z_a}x}y}
\\\app{y_b}x
 &=
 &\varepsilon y\stopq\forall z\stopq\Ppppdrei x y z
\\\end{array}}
\hfill
\math{\begin{array}[t]{|l}
\\
\\
\\
\\
\\\end{array}}
\hfill
\math{\begin{array}[t]{l l l}
  x_a
 &=
 &\varepsilon x\stopq\neg\Ppppdrei
    {x}{\app{y_a}{x}}{\app{\app{z_a}x}{\app{y_a}{x}}}
\\x_b
 &=
 &\varepsilon x\stopq\neg\Ppppdrei 
  x{\app{y_a}x}{\app{\app{z_a}x}{\app{y_b}x}}
\\x_c
 &=
 &\varepsilon x\stopq\neg\Ppppdrei 
  x{\app{y_b}x}{\app{\app{z_a}x}{\app{y_b}x}}
\\x_d
 &=
 &\varepsilon x\stopq\neg\forall z\stopq\Ppppdrei x{\app{y_b}x}z
\\x_e
 &=
 &\varepsilon x\stopq\neg\exists y\stopq\forall z\stopq\Ppppdrei x y z
\\\end{array}}\hfill\mbox{}\par\noindent
When we eliminate inside--out 
(\ie\ start with the elimination of \nlbmath{\forall z.}) 
the transformation is
\\\noindent\math{
\forall x\stopq\exists y\stopq\Ppppdrei x y{\app{\app{z_a}x}y}
}, \hfill
\math{
\forall x\stopq\Ppppdrei x{\app{y_a}x}{\app{\app{z_a}x}{\app{y_a}x}}
}, \hfill
\math{
\Ppppdrei{x_a}{\app{y_a}{x_a}}{\app{\app{z_a}{x_a}}{\app{y_a}{x_a}}}
}\par\noindent
When we eliminate outside--in 
(\ie\ start with the elimination of \nlbmath{\forall x.}) 
the transformation is
\\\noindent\math{
\exists y\stopq\forall z\stopq\Ppppdrei{x_e}y z
}, \hfill
\math{
\forall z\stopq\Ppppdrei{x_e}{\app{y_b}{x_e}}z
}, \hfill
\math{
\Ppppdrei{x_e}{\app{y_b}{x_e}}{\app{\app{z_a}{x_e}}{\app{y_b}{x_e}}}
},\\
\mbox{} \hfill\hfill\ldots, \hfill
\math{
\Ppppdrei{x_a}{\app{y_a}{x_a}}{\app{\app{z_a}{x_a}}{\app{y_a}{x_a}}}
}\\\noindent
where the dots represent the rewritings of \math{x_e} over \math{x_d},
\math{x_c}, \math{x_b} to \math{x_a} (four times) and of \math{y_b} to
\math{y_a} (twice in \nolinebreak addition).

Note that the resulting formula is the same in both cases. Indeed,
it does not depend on the order in which 
we eliminate the quantifiers.
Moreover, notice that this formula is quite deep.
Indeed, in general 
\math n \nolinebreak nested quantifiers result in an \math\varepsilon-nesting
depth of \nlbmath{2^n\tight-1} and huge \math\varepsilon-terms 
(such as \math{x_a}) occur up to
\math n \nolinebreak times with commitment to their choice. 
Let us have a closer look to see this.
If we write the resulting formula as 
\par\noindent\LINEmaths
{\Ppppdrei{x_a}{y_c}{z_d}
}{}(\ref{example subordinate}.1)\par\noindent
by setting
\bigmaths{y_c={\app{y_a}{x_a}}}, 
and
\bigmaths{z_d={\app{\app{z_a}{x_a}}{\app{y_a}{x_a}}}}, 
then we have
\par\noindent\math{
\mbox{}\hfill\begin{array}{@{}l l l l@{}}\mbox{}\hfill~~~~~~~~~~~~~\mbox{}
 &z_d
 &=
 &\varepsilon z\stopq\neg\Ppppdrei{x_a}{y_c}{z}
  \hfill\mbox{(\ref{example subordinate}.2)}
\\&y_c
 &=
 &\varepsilon y\stopq\Ppppdrei{x_a}{y}{\app{z_c}{{y}}}
  \hfill\mbox{(\ref{example subordinate}.3)}
\\&&&\begin{array}{@{}l l l l}
  \mbox{with~~~~~}
 &\app{z_c}{y}
 &=
 &\varepsilon z\stopq\neg\Ppppdrei{x_a}{{y}}{z}
\\\end{array}
  \hfill\mbox{(\ref{example subordinate}.4)}
\\&x_a
 &=
 &\varepsilon x\stopq\neg\Ppppdrei{x}{\app{y_a}{x}}{\app{z_b}{x}}
  \hfill\mbox{(\ref{example subordinate}.5)}
\\&&&\begin{array}{@{}l l l l}
  \mbox{with~~~~~}
 &\app{z_b}{x}
 &=
 &\varepsilon z\stopq\neg\Ppppdrei{{x}}{\app{y_a}{{x}}}{z}
\\\mbox{and}
 &\app{y_a}{x}
 &=
 &\varepsilon y\stopq\Ppppdrei{x}{y}{\app{\app{z_a}x}y}
\\&&&\begin{array}{@{}l l l l}
  \mbox{with~~~~~}
 &\app{\app{z_a}x}y
 &=
 &\varepsilon z\stopq\neg\Ppppdrei{x}{y}{z}
\\\end{array}
\\\end{array}
  \hfill
\begin{array}{r@{}}
  \mbox{(\ref{example subordinate}.6)}
\\\mbox{(\ref{example subordinate}.7)}
\\\mbox{(\ref{example subordinate}.8)}
\\\end{array}
%\\\multicolumn{4}{@{}l@{}}{\makebox[\textwidth]{}}
\\\end{array}}\par\noindent
Firstly, note that the free variables \math x and \nlbmath y
in the \math\varepsilon-terms 
\app{z_c}y, \app{z_b}x, \app{y_a}x, \app{\app{z_a}x}y are  
actually bound by the next \nlbmath\varepsilon\ to the left,
to which the respective \math\varepsilon-terms thus become 
subordinate.
For example, the \math\varepsilon-term \nlbmath{\app{z_c}{y}} is subordinate
to the \math\varepsilon-term \nlbmath{y_c}.
Secondly, the top \math\varepsilon-binders on the right-hand sides
of the defining equations
are exactly those that require a commitment to their choice.
This means that each of \math{z_a}, \math{z_b}, \math{z_c}, \math{z_d} and
each of \math{y_a}, \math{y_c} may be chosen 
differently without affecting soundness of the equivalence transformation.
Note that the variables are strictly nested into each other.
Thus we must choose in the order of 
\math{z_a}, \math{y_a}, \math{z_b}, \math{x_a}, \math{z_c}, \nlbmath{y_c}, 
\nlbmath{z_d}. \ 
Moreover, for \math{z_c}, \math{z_b}, \math{y_a}, \math{z_a} we actually
have to choose a function instead of a simple value.
In \nolinebreak \hilbert's view, however, 
there are neither functions nor objects at all, but only terms,
where \math{x_a} reads\\\LINEmaths{
 \varepsilon x\stopq\ \neg
 \apptotuple\Ppsymbol
 {\superdisplaytrip 
    {x}
    {\varepsilon y_\alpha.\,\Ppppdrei
        {x}
        {y_\alpha}
        {\varepsilon z_\alpha.\,\neg\Ppppdrei{x}{y_\alpha}{z_\alpha}}}
    {\varepsilon z_\beta.\,\neg\apptotuple\Ppsymbol{\displaytrip
       {x}
       {\varepsilon y_\alpha.\,\Ppppdrei
          x
          {y_\alpha}
          {\varepsilon z_\alpha.\,\neg\Ppppdrei x{y_\alpha}{z_\alpha}}
        \smallheadroom\smallfootroom}
       {z_\beta}}\headroom\footroom}}
}{}\\\noindent
and \math{y_c} and \math{z_d} take several lines more to write them down.
\end{example}
\vfill\pagebreak

\yestop\noindent
For 
\bigmaths{\forall x\stopq\forall y\stopq\forall z\stopq\Ppppdrei x y z}{}
instead of 
\bigmaths{\forall x\stopq\exists y\stopq\forall z\stopq\Ppppdrei x y z},
we get the same exponential growth of nesting depth
as in \examref{example subordinate} above, when we completely eliminate the 
quantifiers using \nolinebreak (\math{\varepsilon_2}).
The only difference is that we get additional occurrences of 
\nolinebreak`\math\neg' in \math{y_a}, \math{y_b}, and \nlbmath{y_c}. \ 
But when we have quantifiers of the same kind like 
`\math\exists' or `\math\forall\closesinglequotecomma
we had better choose them in parallel, \eg, for  
\bigmaths{\forall x\stopq\forall y\stopq\forall z\stopq\Ppppdrei x y z}{}
we choose
\ \mbox{\maths
{v_a
 :=\varepsilon v\stopq\neg\Ppppdrei
    {\app{\nth 1}{v}}
    {\app{\nth 2}{v}}
    {\app{\nth 3}{v}}},} \ 
and then take \bigmaths{
  \Ppppdrei
     {\app{\nth 1}{v_a}}
     {\app{\nth 2}{v_a}}
     {\app{\nth 3}{v_a}}
}{}
as result of the elimination.

\begin{sloppypar}
Roughly speaking, in today's theorem proving, \cf\ \eg\ 
\cite{fitting}, \cite{wirthcardinal}, 
the exponential explosion of term depth of \examref{example subordinate} 
is avoided by an outside--in 
removal of \math\delta-quantifiers\emph{without
removing the quantifiers below \math\varepsilon-binders}
and by a replacement of \math\gamma-quantified variables
with \fev s.
For \nolinebreak the case of \examref{example subordinate}, 
this yields \bigmaths{\Ppppdrei{x_e}{\existsvari y{}}{z_e}}{}
with
\bigmaths{
  {z_e}
  =
  \varepsilon z\stopq\neg
  \Ppppdrei{x_e}{\existsvari y{}}{z}
}{} and
\ \mbox{\maths{
  {x_e}
  =
  \varepsilon x\stopq\neg\exists y\stopq\forall z\stopq\Ppppdrei{x}{y}{z}
}.} \ \ 
Thus, in general,
the nesting of binders for the complete elimination of a prenex of 
\math n quantifiers does not become deeper than
\nlbmath{\frac 1 4 \inpit{n\tight+1}^2}. \ 

Moreover, if we are only interested in reduction and not in equivalence
transformation of a formula, 
we can abstract Skolem terms 
from the \math\varepsilon-terms and 
just reduce to the formula 
\bigmaths{\Ppppdrei
  {\forallvari x{}}
  {\existsvari y{}}
  {\app{\forallvari z{}}{\existsvari y{}}}
}.
In a non-Skolemizing inference system with a \vc\ we get
\bigmaths{\Ppppdrei
  {\forallvari x{}}
  {\existsvari y{}}
  {\forallvari z{}}
}{}
instead, with \bigmaths{\{\pair{\existsvari y{}}{\forallvari z{}}\}}{}
as an extension to the \vc.
Note that with Skolemization or \vc s
we have no growth of nesting depth at all, and the same will be the case
for our approach to \math\varepsilon-terms.
\end{sloppypar} 

\subsection{Do not be afraid of Indefiniteness!}
\label{section do not be afraid}
From the discussion in \sectref{section indefinite choice} and 
\sectref{section in the literature},
one could get the impression that an 
indefinite logical treatment of the \nlbmath\varepsilon\ is not easy to find.
Indeed, on the first sight, 
there is the problem that some standard axiom schemes 
cannot be taken for granted,
such as substitutability 
\par\noindent\LINEmaths{
  s\tightequal t\nottight{\nottight\implies}\app f s\tightequal\app f t
}{}\par\noindent{(note that this is similar to (E2) 
 of \sectref{section E2}
 when we take logical equivalence as equality!)}\\
and such as reflexivity
\\\noindent\LINEmaths{
  t\tightequal t
}{}\par\noindent{(note that ({\sc Reflex}) of 
\sectref{subsection indefinite} 
is an instance of this!)}\par
This means that it is not definitely okay to
replace a subterm with an equal term
and that even syntactically equal terms may not be definitely equal.

It may be interesting to see that---in computer programs---we 
are quite used to committed choice and to an indefinite behavior of choosing,
and that the violation of 
substitutability and even reflexivity is no problem there:

\begin{example}
[Violation of Substitutability and Reflexivity in Programs\halftop]
\par\noindent
In the implementation of the specification of the web-based hypertext system
of \cite{productmodel}
we needed a function that chooses an element from a set implemented as
a list. Its \ml\ code is\footroom\notop\begin{verbatim}
fun choose s = case s of Set (i :: _) => i | _ => raise Empty;
\end{verbatim}\notop\headroom
And, of course, it simply returns the first element of the list.
For another set that is equal---but where the list may have another order---the
result may be different. 
Thus, the behavior of the function
{\tt choose} is indefinite for a given set, 
but any time it is called
for an implemented set, it chooses a special element and\emph
{commits to this choice}, \ie\ when called again, it returns the same 
value. In this case we have 
\bigmaths{\mbox{\tt choose s}\nottight{\nottight=}\mbox{\tt choose s}}, but 
\bigmath{{\tt s}\nottight={\tt t}}
does not imply
\bigmaths{\mbox{\tt choose s}\nottight{\nottight=}\mbox{\tt choose t}}.
In an implementation where some parallel reordering of lists may take
place, even 
\bigmath{\mbox{\tt choose s}\nottight{\nottight=}\mbox{\tt choose s}}
may be wrong.
\end{example}

\yestop\noindent
From this example we may learn that 
the question of
\bigmath{\mbox{\tt choose s}\nottight{\nottight=}\mbox{\tt choose s}}
may be indefinite until the choice steps have actually been performed.
\emph{This is exactly how we will treat our \nlbmath\varepsilon.} \ 
The steps that are performed in logic are proof steps.

\yestop\noindent
Thus, on the one hand, when we want to prove 
\par\noindent\LINEmath{
  \varepsilon x.\,\truepp
  \nottight{\nottight =}
  \varepsilon x.\,\truepp
}\par\noindent
we can choose \zeropp\ for both occurrences of 
\bigmaths{\varepsilon x.\,\truepp},
get \bigmaths{\zeropp\tightequal\zeropp}, and the proof is successful.
On the other hand, when we want to prove 
\par\noindent\LINEmath{
  \varepsilon x.\,\truepp
  \nottight{\nottight{\not=}}
  \varepsilon x.\,\truepp
}\par\noindent
we can choose \zeropp\ for one occurrence and \onepp\ for the other,
get \bigmaths{\zeropp\tightnotequal\onepp}, 
and the proof is successful again.
This procedure may seem wondrous again, but is very similar to something
quite common with \fev s, \cf\ \sectref{section free}:
On the one hand, when we want to prove 
\par\noindent\LINEmath{\existsvari x{}\tightequal\existsvari y{}
}\par\noindent
we can choose \zeropp\ to substitute for both \existsvari x{} and 
\existsvari y{},
get \bigmaths{\zeropp\tightequal\zeropp}, and the proof is successful.
On the other hand, when we want to prove 
\par\noindent\LINEmath{\existsvari x{}\tightnotequal\existsvari y{}}
\par\noindent
we can choose \zeropp\ to substitute for \existsvari x{} and \onepp\
to substitute for \existsvari y{},
get \bigmaths{\zeropp\tightnotequal\onepp}, 
and the proof is successful again.

\subsection
{Replacing \math\varepsilon-terms with \SFUV s}
\label{section replacing epsilon}\noindent
There is an important difference between the inequations 
\bigmath{\varepsilon x.\,\truepp\nottight{\nottight{\not=}}
  \varepsilon x.\,\truepp} 
and 
\bigmath{\existsvari x{}\tightnotequal\existsvari y{}}
at the end of the previous \sectref{section do not be afraid}: \ 
The latter does not violate the reflexivity axiom! \ 
And we are going to cure the violation of the former
immediately with the help of a special kind of free variables, 
namely our\emph{\sfuv s}, \cf\ \sectref{section free}. \ 
Now, instead of 
\bigmath{\varepsilon x.\,\truepp\nottight{\nottight{\not=}}
  \varepsilon x.\,\truepp} 
we write \bigmath{\sforallvari x{}\tightnotequal\sforallvari y{}}
and remember what these \sfuv s stand for by storing this into a function
\nlbmath C, called a\emph\cc: 
\\\noindent\LINEmaths{\begin{array}{l l l@{}l}
  \app C{\sforallvari x{}}
 &:=
 &\truepp
 &,
\\\app C{\sforallvari y{}}
 &:=
 &\truepp
 &. 
\\\end{array}
}{}\par\yestop\noindent
For a first step, suppose that our \math\varepsilon-terms 
are not subordinate to 
any outside binder, \cfnlb\ \defiref{definition subordinate}.
% that binds variables inside the \math\varepsilon-term.
Then, we can replace an \math\varepsilon-term
\ \mbox{\math{\varepsilon z.\, A}} \ with a new \sfuv\ 
\nlbmath{\sforallvari z{}} and 
extend the partial function \nlbmath C \nolinebreak by 
\par\noindent\LINEmaths{\begin{array}{l l l@{}l}
  \app C{\sforallvari z{}}
 &:=
 &A\{z\tight\mapsto\sforallvari z{}\}
 &.
\\\end{array}
}{}\par\noindent
By this procedure we can eliminate all \math\varepsilon-terms
without loosing any syntactical information.

As a first consequence of this elimination,
the substitutability and reflexivity axioms
are immediately regained,
and the problems discussed in \sectref{section do not be afraid} disappear.

A second reason for replacing the \math\varepsilon-terms
with \sfuv s 
is that the latter can solve the question whether a committed choice is
required:
We can express---on the one hand---a committed choice
by using a single \sfuv\
and---on the other hand---a choice without commitment 
by using several variables with 
the same \cc. 

Indeed, this also solves our problems with committed choice
of \examref{example committed choice} of 
\sectref{section committed choice}: 
Now, again using (\math{\varepsilon_1}), \ \ 
\bigmaths{\exists x.\,\inpit{x\not= x}}{} reduces to
\bigmaths{\sforallvari x{}\not=\sforallvari x{}}{} with 
\par\noindent\LINEmaths
{\begin{array}{l l l@{}l}
  \app C{\sforallvari x{}}
 &:=
 &\inpit{\sforallvari x{}\not=\sforallvari x{}}
\\\end{array}
}{}\par\noindent
and the proof attempt immediately fails due to the 
now regained reflexivity axiom.

\begin{sloppypar}\yestop\yestop\noindent
As the second step, we still have to explain what to do with 
subordinate \math\varepsilon-terms. \ 
If the \mbox{\math\varepsilon-term \ \math{\varepsilon z\stopq A}} \ 
contains free occurrences of exactly the distinct variables 
\math{v_0}, \ldots, \math{v_{l-1}}, 
%that are not bound by quantifiers or other binders inside
%the \math\varepsilon-term, 
then we have to replace this \math\varepsilon-term
with the application term
\nlbmath{\sforallvari z{}(v_0)\cdots(v_{l-1})}
of the same type as \nlbmath z
(for a new \sfuv\ \nolinebreak\sforallvari z{})
and to extend the \cc\ \nlbmath C by
\par\noindent\LINEmaths{\begin{array}{l l l @{}l}
  \app C{\sforallvari z{}}
 &:=
 &\lambda v_0\stopq\ldots\lambda v_{l-1}\stopq\inparentheses
  {A\{z\mapsto\sforallvari z{}(v_0)\cdots(v_{l-1})\}}
 &.
\\\end{array}
}{}\end{sloppypar}

\yestop\yestop\begin{example}[Higher-Order \CC]
\hfill{\em(continuing \examref{example subordinate} of 
 \sectref{section quantifier elimination and subordinate})}\label
{example higher-order choice-condition}\par\noindent
In our framework, the complete elimination of \math\varepsilon-terms in 
(\ref{example subordinate}.1) of \examref{example subordinate} 
% turns 
% \par\noindent\LINEmaths{
%   \forall x\stopq\exists y\stopq\forall z\stopq\Ppppdrei x y z}{}
% \\into
results in 
\par\noindent\phantom
{(\cf\ (\ref{example subordinate}.1)!)}\LINEmaths{
  \Ppppdrei{\sforallvari x a}{\sforallvari y c}{\sforallvari z d}
}{}(\cf\ (\ref{example subordinate}.1)!)\par\halftop\noindent
% and the subordinate \math\varepsilon-terms are replaced 
with the following higher-order \cc:
\par\halftop\noindent\mbox{}\hfill\math{\begin{array}{@{}l l l l@{}}
\mbox{~~~~~~~~~~~~~~~~~~~~~~~~~~~~~~~~~~~~~~~~~}\hfill
 &\app{C}{\sforallvari z d}
 &:=
 &\neg\Ppppdrei{\sforallvari x a}{\sforallvari y c}{\sforallvari z d}
  \hfill\mbox{(\cf\ (\ref{example subordinate}.2)!)}
\\\headroom
 &\app{C}{\sforallvari y c}
 &:=
 &\Ppppdrei
    {\sforallvari x a}
    {\sforallvari y c}
    {\app{\sforallvari z c}{\sforallvari y c}}
  \hfill\mbox{(\cf\ (\ref{example subordinate}.3)!)}
\\\headroom
 &\app{C}{\sforallvari z c}
 &:=
 &\lambda y\stopq\neg\Ppppdrei{\sforallvari x a}{y}{\app{\sforallvari z c}{y}}
  \hfill\mbox{(\cf\ (\ref{example subordinate}.4)!)}
\\\headroom
 &\app{C}{\sforallvari x a}
 &:=
 &\neg\Ppppdrei
    {\sforallvari x a}
    {\app{\sforallvari y a}{\sforallvari x a}}
    {\app{\sforallvari z b}{\sforallvari x a}}
  \hfill\mbox{(\cf\ (\ref{example subordinate}.5)!)}
\\\headroom
 &\app{C}{\sforallvari z b}
 &:=
 &\lambda x\stopq\neg\Ppppdrei
    {x}
    {\app{\sforallvari y a}{x}}
    {\app{\sforallvari z b}{x}}
  \hfill\mbox{(\cf\ (\ref{example subordinate}.6)!)}
\\\headroom
 &\app{C}{\sforallvari y a}
 &:=
 &\lambda x\stopq\Ppppdrei
   {x}
   {\app{\sforallvari y a}{x}}
   {\app{\app{\sforallvari z a}{x}}{\app{\sforallvari y a}{x}}}
  \hfill\mbox{(\cf\ (\ref{example subordinate}.7)!)}
\\\headroom
 &\app{C}{\sforallvari z a}
 &:=
 &\lambda x\stopq\lambda y\stopq
  \neg\Ppppdrei{x}{y}{\app{\app{\sforallvari z a}{x}}{y}}
  \hfill\mbox{(\cf\ (\ref{example subordinate}.8)!)}
\\\multicolumn{4}{@{}l@{}}{\makebox[\textwidth]{}}\end{array}}\\
Notice that this representation of (\ref{example subordinate}.1)
is smaller and easier to understand than all previous ones.
Indeed, by  combination of \math\lambda-abstraction and term sharing
via \sfuv s, in our framework
the \nlbmath\varepsilon\ becomes practically feasible
for the first time.
\end{example}
\vfill\pagebreak

\subsection{Instantiating \SFUV s (``\math\varepsilon-Substitution'')}\label
{section Instantiating Strong Free Universal Variables}\noindent
Having realized \requirementeins\ of 
\sectref{section requirement specification} in the previous 
\nolinebreak\sectref{section replacing epsilon},
in this \sectref{section Instantiating Strong Free Universal Variables}
we are now going to explain how to satisfy \requirementzwei\@.
To this end, 
we have to explain how to replace \sfuv s with terms that 
satisfy their \cc s.

The first thing to know about \sfuv s is:
Just like the 
\fev s and contrary to \wfuv s, the \sfuv s 
are\emph{rigid} in the sense that the only way to replace a \sfuv\ is to 
do it\emph{globally}, \ie\ in all formulas and all \cc s 
in an atomic transaction.

In\emph{reductive} theorem proving such as
in sequent, tableau, or matrix calculi
we are in the following situation:
While
a \fev\ \existsvari x{} 
can be replaced with nearly everything,
the replacement of a \sfuv\ \nolinebreak \sforallvari y{}
requires some proof work, and
a \wfuv\ \nlbmath{\wforallvari z{}} cannot be instantiated at all.
 
Contrariwise, when formulas are used as tools instead of tasks,
 \wfuv s can indeed be replaced---and 
 this even locally (\ie\ non-rigidly). 
 This is the case not only
 for purely\emph{generative} calculi, 
 such as resolution and paramodulation calculi
 and \hilbert-style calculi such as the predicate calculus of 
 \nolinebreak\cite{grundlagen}, but also for the 
 lemma and induction hypothesis application
 in the otherwise reductive calculi of \cite{wirthcardinal},
 \cfnlb\ \cite[\litsectref{2.5.2}]{wirthcardinal}.

More precisely---again considering\emph{reductive} theorem proving, 
where formulas are proof tasks---a 
\fev\ \nlbmath{\existsvari x{}} may be instantiated with
any term (of appropriate type) that does not violate the current \vc,
\cf\ \sectref{section substitutions} for details.
The instantiation of a \sfuv\ \nlbmath{\sforallvari y{}}
additionally requires some proof work depending on the current \cc\ \nlbmath C,
which also puts some requirements on the \vc\ \nlbmath R and thus is
formally called an\emph{\math R-\cc}, \ 
\cf\ \defiref{definition choice condition} for the formal details. \ 
In general, if a substitution \nlbmath\sigma\
replaces---possibly among other \fev s and \sfuv s---the 
\sfuv\ \nolinebreak\sforallvari y{} in the 
domain of the \math R-\cc\ \nlbmath C, \ 
% namely if
% \par\noindent\LINEmaths{\app{C}{\sforallvari y{}}
% =\lambda v_0\stopq\ldots\lambda v_{l-1}\stopq B
% }{}\par\halftop\noindent
% is given,
then---to know that the global instantiation of the whole proof forest
with \nlbmath\sigma\
preserves its soundness---we have to prove 
\bigmaths{\inpit{\app{Q_C}{\sforallvari y{}}}\sigma}, 
where \math{Q_C} is given as follows:

\begin{definition}[\math{Q_C}]\label{definition Q}\\\noindent
For an \math R-\cc\ \nlbmath C, 
we let \nlbmath{Q_C} be a total function from \nlbmath
{\DOM C}
into the set of single-formula sequents
such that for each \math{\sforallvari y{}\in\DOM C} \ 
with 
\bigmaths{\app C{\sforallvari y{}}=
               \lambda v_0\stopq\ldots\lambda v_{l-1}\stopq 
               B}{} for a formula \math B,
we \nolinebreak have
\par\noindent\LINEmath{\app{Q_C}{\sforallvari y{}}\hfill=\hfill
  \forall v_0\stopq\ldots\forall v_{l-1}\stopq
  \inparentheses{
    ~\exists y\stopq
    \inpit{B\{\sforallvari y{}(v_0)\cdots(v_{l-1})
    \mapsto y\}}
    \nottight{\nottight\implies}
    B~}
    }\par\noindent
for an arbitrary fresh bound variable \math
{y\in\Vbound\tightsetminus\VAR{\app C{\sforallvari y{}}}}.
\end{definition}

\yestop\noindent
Note that \app{Q_C}{\sforallvari y{}} \ is 
nothing but 
a formulation of axiom\,(\math{\varepsilon_0})
from \sectref{section Why} in our framework,
and \lemmref{lemma Q valid} states its validity.
% where (\cf\ \defiref{definition Q}) \app{Q_C}{\sforallvari y{}}
% is the single-formula sequent
% \par\yestop\noindent\LINEmath{ 
% \forall v_0\stopq\ldots\forall v_{l-1}\stopq 
%   \inparentheses{\smallheadroom\smallfootroom
%     ~\exists y\stopq
%     \inpit{B\{\sforallvari y{}(v_0)\cdots(v_{l-1})\mapsto y\}}
%     \nottight{\nottight\implies}
%     B~}}\par\yestop\noindent
% for a new bound variable \nlbmath{y} of the same type as 
% \math{\sforallvari y{}(v_0)\cdots(v_{l-1})}\@. 
%
% For doing so, 
% we need means of expressing the requirement on 
% \aquasiexRsub\ to replace the 
% \sfuv s in accordance with the 
% compatibility requirement of \cc s in
% \defiref{definition compatibility}(\ref
% {item 2 definition compatibility}). 

It is an essential\footnote
{If the occurrences of \nlbmath{\sforallvari y{}} in 
 \nlbmath{\app C{\sforallvari y{}}} could differ in their arguments,
 there could be irresolvable conflicts on special arguments.
 And, in these conflicts, the choice of a\emph{function as a whole} 
 would essentially violate 
 \hilbert's axiomatizations:
 As only terms and no functions are considered in \cite{grundlagen},
 the axiom schemes (\math{\varepsilon_0})
 and ({\em\math\varepsilon-formula})
 (\cfnlb\ \sectrefs{section Why}{section proof-theoretic origin})
 seem to require us to choose the values of this function individually.
 For example, in case of 
 \\\LINEmaths{\headroom\footroom
 \app C{\sforallvari y{}} 
 \nottight{\nottight{\nottight{\nottight=}}}
 \lambda b\stopq\inparentheses{\smallheadroom\smallfootroom
 \app{\sforallvari y{}}b \und 
 \neg\inpit
 {\app{\sforallvari y{}}\truepp\und\app{\sforallvari y{}}\falsepp}}
 },\\
 for choosing \nlbmath{\sforallvari y{}},
 we are in conflict between 
 \bigmaths{\lambda b'\stopq\inpit{b'\tightequal\falsepp}}{}
 (\ie\ \math{\lambda b'.\,\neg b'},
  for \math{\app{\app C{\sforallvari y{}}}\falsepp} to be \truepp)
 and
 \ \mbox{\maths{\lambda b'\stopq\inpit{b'\tightequal\truepp}}{}} \ 
 (\ie\ \math{\lambda b'.\,b'},
 for \math{\app{\app C{\sforallvari y{}}}\truepp} to be \truepp).
 \vfill\pagebreak}
property of our \cc s that 
all occurrences of \nlbmath{\sforallvari y{}} in \nlbmath B
necessarily are of the form 
\bigmaths{\sforallvari y{}(v_0)\cdots(v_{l-1})},
\cfnlb\ \defiref
{definition choice condition}(\ref
{item two definition choice condition})\@.
Therefore, the formula 
\bigmaths{\app{Q_C}{\sforallvari y{}}}{}
is logically equivalent to the formula
\\\noindent\LINEmath{
\forall v_0\stopq\ldots\forall v_{l-1}\stopq 
  \inparentheses{\smallheadroom\smallfootroom
    ~\exists z\stopq
    \inpit{B\{\sforallvari y{}\tight\mapsto z\}}
    \nottight{\nottight\implies}
    B~}}\\\noindent
for a new bound variable \nlbmath{z}
of the same type as \nlbmath{\sforallvari y{}}\@. \ 

\yestop
\begin{example}[Predecessor Function\halftop]\label
{example choice-conditions and predecessor}\\\noindent
Suppose that our domain is natural numbers and that
\sforallvari y{\inpit{\psymbol1}} has the \cc\
\par\noindent\LINEmaths
{\app{C}{\sforallvari y{\inpit{\psymbol1}}}
=\lambda v\stopq
 \inpit
   {v\tightequal\sforallvari y{\inpit{\psymbol1}}(v)\tight+\onepp}}.
\par\noindent
Then, before we may instantiate \nlbmath{\sforallvari y{\inpit{\psymbol1}}} 
with the symbol \nlbmath\psymbol\ for the predecessor function
specified by \bigmaths
{\forall x\stopq\inpit{\ppp{\tightplusppnoparentheses x 1}
\tightequal x}},
we have to prove \bigmaths{\inpit{\app Q{\sforallvari y{\inpit{\psymbol1}}}}
\{\sforallvari y{\inpit{\psymbol1}}\mapsto\psymbol\}}, which reads as
\par\noindent\LINEmaths
{\forall v\stopq\inparentheses{\smallheadroom\smallfootroom
 \exists y\stopq\inpit{v\tightequal\tightplusppnoparentheses y\onepp}
 \implies
 \inpit{v\tightequal\tightplusppnoparentheses{\ppp v}\onepp}}
},\par\noindent and is valid in arithmetic.\end{example}

\yestop\yestop\begin{example}[Canossa 1077]\hfill{\em
(continuing \examref{example Pope better})}\label
{example Canossa}\par\noindent
The situation of \examref{example Pope better} now reads 
\par\halftop\noindent\phantom{(\ref{example Canossa}.1)}\LINEmath{
  \HG
  \nottight{\nottight =}
  \sforallvari z 0
  \nottight{\nottight{\nottight{\nottight\und}}}
  \ident{Joseph}
  \nottight{\nottight =}
  \sforallvari z 1
}(\ref{example Canossa}.1)\par\halftop\noindent
with\LINEmaths
{\app C{\sforallvari z 0}=\Fatherpp{\sforallvari z 0}{\ident{Jesus}},}{}
\phantom{with}\\and\LINEmaths
{\app C{\sforallvari z 1}=\Fatherpp{\sforallvari z 1}{\ident{Jesus}}.}{}
\phantom{and}\par\halftop\noindent 
This does not bring us into the old trouble with the Pope 
because nobody knows whether
\bigmaths{\sforallvari z 0=\sforallvari z 1}{} holds.
\par\halftop\noindent
On the one hand, knowing (\ref{example iota}.2) from \examref{example iota}
of \sectref{section from iota to epsilon},
we can prove
(\ref{example Canossa}.1) as follows:
We \nolinebreak first substitute \sforallvari z 0 with \nlbmath\HG\
because, for \math{\sigma_0:=\{\sforallvari z 0\tight\mapsto\HG\}},
we have \bigmaths{\inpit{\app C{\sforallvari z 0}}\sigma_0}{}
and---a fortiori---\bigmaths{\inpit{\app{Q_C}{\sforallvari z 0}}\sigma_0},
which reads
\par\noindent\LINEmaths{
 {\exists z\stopq\Fatherpp{z}{\ident{Jesus}}
  \nottight{\nottight{\implies}}
  \Fatherpp\HG{\ident{Jesus}}}
};\par\noindent
and, analogously, substitute \sforallvari z 1 with \math{\ident{Joseph}}
because, for \math{\sigma_1:=\{\sforallvari z 1\tight\mapsto\ident{Joseph}\}},
we have \bigmaths{\inpit{\app C{\sforallvari z 1}}\sigma_1}{}
and---a fortiori---\bigmaths{\inpit{\app{Q_C}{\sforallvari z 1}}\sigma_1}.
After these substitutions, (\ref{example Canossa}.1) becomes the tautology
\par\noindent\LINEmath{
  \HG
  \nottight{\nottight =}
  \HG
  \nottight{\nottight{\nottight{\nottight\und}}}
  \ident{Joseph}
  \nottight{\nottight =}
  \ident{Joseph}
}\par\yestop\noindent
On the other hand, if we want to have trouble, we can 
apply the substitution  
\par\halftop\noindent\LINEmaths{\sigma'\ =\ \{
  \sforallvari z 0\mapsto\ident{Joseph}
,\ 
  \sforallvari z 1\mapsto\ident{Joseph}
\}}{}\par\halftop\noindent
to (\ref{example Canossa}.1) because of 
\bigmaths{\inpit{\app{Q_C}{\sforallvari z 0}}\sigma'
         =\inpit{\app{Q_C}{\sforallvari z 1}}\sigma_1
         =\inpit{\app{Q_C}{\sforallvari z 1}}\sigma'}.
\\Then our task is to show 
\par\noindent\LINEmath{
  \HG
  \nottight{\nottight =}
  \ident{Joseph}
  \nottight{\nottight{\nottight{\nottight\und}}}
  \ident{Joseph}
  \nottight{\nottight =}
  \ident{Joseph}}\par\noindent
Note that this procedure is stupid already under the aspect of theorem proving
alone.\vfill\pagebreak\end{example}

\section{Formal Presentation of Our Indefinite Semantics}\label
{section formal discussion}\noindent
To satisfy \requirementdrei\ of 
\sectref{section requirement specification},
in this \sectref{section formal discussion}
we present our novel semantics for the \nlbmath\varepsilon\emph{formally}.
This is required for precision and consistency.
As consistency of our new semantics is not trivial at all, 
technical rigor cannot be avoided.
From \sectref{section new indefinite}
the reader should have a good intuition of our intended 
representation and semantics of the \nlbmath\varepsilon,
\sfuv s, and \cc s in our framework. \ 
\mbox{\Sectref{section formal discussion} organizes} as follows:
In \sectref{section substitutions} and 
\sectref{section existential valuations} 
we formalize \vc s and explain how to
deal with \fev s syntactically and semantically.
In \sectref{section validity} we introduce a preliminary semantics
that does not treat \sfuv s properly,
and in \sectref{section strong validity} the proper semantics.
Only between these two 
\nolinebreak\sectrefs{section validity}{section strong validity},
we can discuss \cc s (\sectref{section choice-conditions}).
Our interest goes beyond soundness in that we want 
``\tightemph{preservation of solutions}\closequotefullstop
By this we mean the following:
All\emph{closing substitutions} for the \fev s and \sfuv s---\ie\ all solutions 
that transform a proof attempt (to which a proposition has been reduced)
into a closed proof---are also solutions of the original proposition.
This is similar to a proof in Prolog, 
computing answers to a query proposition that contains \fev s.
Therefore, 
in \nolinebreak\sectref{section reduction}
we discuss this solution-preserving notion of\emph{reduction}, especially 
under the aspect of global instantiation of \sfuv s.
Finally, in \nolinebreak\sectref{subsection On the Design of Similar Operators}
we give some hints on the design of operators similar to our
\nlbmath\varepsilon. \ 
All in all, in this \nolinebreak\sectref{section formal discussion},
we extend and simplify the presentation of \cite{wirthcardinal},
which, however, additionally contains comparative discussions,
compatible extensions for\emph\descenteinfinie,
and those proofs that are \mbox{omitted here}.\enlargethispage{0.2ex}%
\newcommand\basicssectiontitle{Basic Notions and Notation}%
\subsection\basicssectiontitle\label\basicssectiontitle
`\N' denotes the set of natural numbers 
and `\math\prec' the ordering on \nolinebreak\N\@. 
Let \bigmaths{\posN:=\setwith{n\tightin\N}{0\tightnotequal n}}.
% `\ZZ' denotes the set of integers.
We \nolinebreak 
use `\math\uplus' for the union of disjoint classes and `\id' for the
identity function.
For classes \nlbmath R, \math A, and \math B we define:\smallfootroom
\\\noindent\math{\begin{array}{@{\indent}l@{\ }l@{\ }l@{~~~~~~}l@{}}
   \DOM R
  &:=
  &\setwith{\!a}{\exists b\stopq          (a,b)\tightin R\!}
  &\mbox{\tightemph{domain}} 
 \\\domres R A
  &:=
  &\setwith{\!(a,b)\tightin R}{a\tightin A\!}
  &\mbox{\tightemph{restriction to }}
   A
 \\\relapp R A
  &:=
  &\setwith{\!b}{\exists a\tightin A\stopq(a,b)\tightin R\!}
  &\mbox{\tightemph{image of }}
   A
   \mbox{, \ \ie\ \ }
   \relapp R A=\RAN{\domres R A}
 \\\multicolumn{4}{@{}l@{}}{\mbox
 {And the dual ones:}\smallheadroom\smallfootroom}
 \\\RAN R
  &:=
  &\setwith{\!b}{\exists a\stopq          (a,b)\tightin R\!}
  &\mbox{\tightemph{range}}
 \\\ranres R B
  &:=
  &\setwith{\!(a,b)\tightin R}{b\tightin B\!}
  &\mbox{\tightemph{range-restriction to \math B}}
 \\\revrelapp R B
  &:=
  &\setwith{\!a}{\exists b\tightin B\stopq(a,b)\tightin R\!}
  &\mbox{\tightemph{reverse-image of }}
   B
   \mbox{, \ \ie\ \ }
   \revrelapp R B=\DOM{\ranres R B}
 \\\end{array}}
\\\noindent\smallheadroom
Furthermore, we use `\math\emptyset' to denote the empty set as well as the
empty function.
Functions are (right-) unique relations and
the meaning of `\math{f\tight\circ g}' is extensionally given by
\bigmaths{\app{\inpit{f\tight\circ g}}x=\app g{\app f x}}.
% Note that we take the operator `\math\circ' to have 
% higher priority than the operators `\math\cup' and \nolinebreak`\math\uplus'.
The\emph{class of total functions from \math A to \math B}
is denoted as \nolinebreak\FUNSET A B.
The\emph{class of (possibly) partial functions from \math A to \math B}
is denoted as \nolinebreak\PARFUNSET A B. \ 
Both \FUNSET{}{} and \PARFUNSET{}{} associate to the right, 
\ie\ \PARFUNSET A{\FUNSET B C} reads 
\PARFUNSET A{\inpit{\FUNSET B C}}.

Let \math R be a binary relation. \ 
\math R is said to be a relation\emph{on \math A} \ \udiff\ \ 
\bigmath{\DOM R\nottight\cup\RAN R\nottight{\nottight\subseteq} A.} \ 
\math R \nolinebreak is\nolinebreak\emph{irreflexive} \udiff\  
\bigmath{
  \id\cap R=\emptyset
.} 
It is \math A{\em-reflexive\/} \udiff\  
\bigmath{
  \domres\id A\subseteq R
.} 
Speaking of a\emph{reflexive} relation
we refer to the largest \math A that is appropriate in the local context,
and referring to this \math A
we write \math{R^0} 
%or \redindexn 0{} 
to ambiguously denote 
\math{\domres\id A}. \ 
With \math{R^1:=R}, and
\math{R^{n+1}:=R^{n}\tight\circ R} for \math{n\in\posN}, \ 
\math{R^m} \nolinebreak 
denotes the \math m-step relation
for \nlbmath R. \ 
The\emph{transitive closure} of \nlbmath R
%\red\ 
is 
\bigmaths{\transclosureinline R:=\bigcup_{n\in\posN}R^n}. \ 
%\bigmath{\trans:=\bigcup_{n\in\posN}\redindexn n{}.}
The\emph{reflexive \& transitive closure} of \nlbmath R
%\red\ 
is 
\bigmath{
  \refltransclosureinline R
  :=
  \bigcup_{n\in\N}R^n
.}
% The\emph{reverse} of \math R is
% \nolinebreak\math
% {\reverserelation R:=\setwith{\pair b a}{\pair a b\tightin R}}. \ 
% A sequence \bigmath{\inpit{s_i}_{i\in\N}}
% is\emph{non-terminating in \math R}
% \udiff\ \bigmath{s_i\nottight R s_{i+1}} for all \math{i\in\N}.
% \ \math R \nolinebreak is\emph{terminating} 
% \udiff\ there are no non-terminating sequences in \nlbmath R. \ 
\mbox{A relation \math R (on \math A)} is \mbox{\em\wellfounded}\/ 
\udiff\ 
any non-empty class \math B (\math{\tightsubseteq A}) has an
\math R-minimal element, \ie\ \math
{\exists a\tightin B\stopq\neg\exists a'\tightin B\stopq a' R\,a}.\pagebreak

\subsection{Variables and \protect\RSubs}\label
{section substitutions}
We assume the following four sets of symbols to be disjoint:
\footroom\\\noindent\LINEnomath{\begin{tabular}{l l@{}}
  \Vsome
 &\tightemph{\fev s},
%\\&
  \ie\ 
  the free variables of \cite{fitting}
\\\Vall
 &\tightemph{\fuv s},
%\\&
  \ie\ 
  nullary parameters, instead of \skolem\ functions
\\\Vbound
 &\tightemph{bound variables},
%\\&
  \ie\ variables to be bound, \cf\ below
\\\math\Sigmaoffont
 &\tightemph{constants},
%\\&
  \ie\ 
  the function and predicate symbols from the signature
\end{tabular}}\par\noindent
As explained in \sectref{section free},
we partition the \fuv s into\emph{\wfuv s} 
and\emph{\sfuv s}: \bigmaths{\Vall=\Vwall\uplus\Vsall}.
We define the\emph{free variables} by \bigmaths{\Vfree:=\Vsome\uplus\Vall}{}
and the\emph{variables} by \bigmaths{\V:=\Vbound\uplus\Vfree}. \ 
Finally, the\emph{rigid} variables by \bigmaths{
\Vsomesall:=\Vsome\uplus\Vsall}. \ 
\sentenceonsetofvariables

Let \math\sigma\ be a substitution. \ 
\math\sigma\ \nolinebreak is a\emph{substitution on \math X} 
\udiff\ \mbox{\math{\DOM\sigma\subseteq X}.} \ \ 
We \nolinebreak denote with `\math{\Gamma\sigma}' 
the result of replacing 
each occurrence of a variable \nlbmath{x\in\DOM\sigma} in \nlbmath{\Gamma} 
with \nlbmath{\app\sigma x}. \ 
(Actually, we may have to rename some 
of the bound variables in \nlbmath{\app\sigma x} when we 
exclude the binding of a variable within the scope of a bound variable 
of the same name.) \ 
Unless otherwise stated,
we tacitly assume that all occurrences of variables from 
\nlbmath\Vbound\ 
in a term or formula or in the range of a substitution
are\emph{bound occurrences} 
(\ie\ \nolinebreak that a variable \math{x\in\Vbound}
occurs only in the scope of a binder on \nlbmath x)
and that each substitution \math\sigma\
satisfies \bigmaths{\DOM\sigma\subseteq\Vfree},
so that no bound occurrences of variables can be replaced 
and no additional variable occurrences can become bound (\ie\ \nolinebreak 
captured) when applying \nlbmath\sigma.

Several binary relations on free variables will be introduced in this
and the following \nolinebreak\sects.
The overall idea is that when \pair x y occurs in such a relation
this means something like 
\ ``\math x \nolinebreak is \nolinebreak necessarily older than \nlbmath y'' \ 
or \ ``the value of \math y
depends on \nlbmath x or is described in terms of \nlbmath x\closequotefullstop 
\begin{definition}[\VC]\label{definition variable condition}\noindent
A\emph\vc\ is a subset of \bigmath{\Vfree\times\Vfree.}
\end{definition}
\begin{definition}[\math\sigma-Update]\label{definition update}
Let \math R be a \vc\ and \math\sigma\ be a substitution.
\\The\emph{\math\sigma-update of \math R} \ is 
\LINEmaths{R\quad\cup\quad\setwith
    {\pair{\freevari z{}}{\freevari x{}}}
    {\freevari x{}\tightin\DOM\sigma
     \und
     \freevari z{}\tightin\VARfree{\app\sigma{\freevari x{}}}
    }}.\end{definition}
\begin{definition}[\RSub]\label
{definition ex r sub}\label{definition quasi-existential}
Let \math R be a \vc.
\\\math\sigma\ is an\emph{\Rsub}
\udiff\  
\math\sigma\ is a substitution 
and the \math\sigma-update of \nlbmath R is \wellfounded. 
\end{definition}
Syntactically, 
\bigmath{\pair{\freevari x{}}{\freevari y{}}\tightin R} 
is to express that an \math R-substitution \nlbmath\sigma\ 
must not replace \nlbmath{\freevari x{}} with a term in which 
\math{\freevari y{}} could ever occur. 
This is guaranteed when
the \math\sigma-updates \nlbmath{R'} 
of \nlbmath R are always required to
be \wellfounded.
For \math{\freevari z{}\in\VARfree{\app\sigma{\freevari x{}}}}, \ 
we get \bigmaths{
\freevari z{}\nottight{R'}\freevari x{}\nottight{R'}\freevari y{}},
blocking \nlbmath{\freevari z{}} against terms containing \freevari y{}. \ 
Note that in practice a \math\sigma-update of \nlbmath R
can always be chosen to be finite. In this case, it \nolinebreak
is \wellfounded\ \uiff\
it \nolinebreak is acyclic.

\newcommand\semanticssectiontitle{Semantical Requirements}\subsection
{\protect\math R-Validity}\label{section validity}\label\semanticssectiontitle
\noindent
Instead of defining validity from scratch,
we require some abstract properties 
typically holding in two-valued semantics. 
Validity is given relative to some \semanticobject\ 
\nolinebreak\salgebra,
assigning a non-empty universe (or \nolinebreak``carrier'')
to each type.
For \math{\X\subseteq\V} we denote 
the set of total \salgebra-valuations of \X\
(\ie\ \nolinebreak
 functions mapping variables to objects 
 of the universe of \nolinebreak\salgebra\
 (respecting \nolinebreak types))
with
\\\noindent\LINEmath{\FUNSET\X\salgebra}\\\noindent
and the set of (possibly) partial \salgebra-valuations of \nlbmath\X\
with 
\\\noindent\LINEmath{\PARFUNSET\X\salgebra}\\\noindent
For \math{\,\FUNDEF\delta\X\salgebra\,} we denote with
`\math{\salgebra\tightuplus\delta}'
the extension of \salgebra\ to the variables of \X\@.
%which are then treated as nullary constants.
More precisely, we assume some
evaluation function `\EVALSYM' such that \EVAL{\salgebra\tightuplus\delta}
\nolinebreak 
maps any term whose constants and freely occurring variables are from 
\nlbmath{\Sigmaoffont\tightuplus\X} into the universe
of \nlbmath\salgebra\ (respecting types) such that for all \math{x\in\X}:
\ \bigmaths{
  \EVAL{\salgebra\tightuplus\delta}
  \funarg x
  {\nottight{\nottight{=}}}
  \delta
  \funarg x
}. \ \
Moreover, \EVAL{\salgebra\tightuplus\delta} \nolinebreak maps
any formula \math B whose constants and freely occurring variables are from
\math{\Sigmaoffont\tightuplus\X} to \TRUEpp\ or \FALSEpp,
such that \ \math B is valid in \math{\salgebra\tightuplus\delta} 
 \ \uiff\ \ \math{\EVAL{\salgebra\tightuplus\delta}(B)\tightequal\TRUEpp}.

Notice that we leave open what our formulas and what our 
\mbox{\math\Sigmaoffont-structures} exactly are. 
The latter can range from a \firstorder\ \math\Sigmaoffont-structure to
a higher-order modal \math\Sigmaoffont-model,
provided that the following two standard textbook lemmas hold for
a term or formula \nlbmath B
(possibly with some\emph{unbound} occurrences of variables from \nlbmath\Vbound)
and a \semanticobject\ \nlbmath\salgebra\ 
with valuation \nlbmath{\,\PARFUNDEF\delta\V\salgebra\,}\@. \ 
\par\halftop\noindent\sloppy{\sc Explicitness Lemma}\\\noindent
The value of the evaluation function on \nlbmath B 
depends only on the valuation of those variables 
that actually occur freely in \nlbmath B; \ formally: \ 
For \X\ being the set of variables that occur freely in \nlbmath B, \ if
\bigmaths{\X\subseteq\DOM\delta}:
\bigmaths{\quad\quad
  \app
  {\EVAL{\salgebra\tightuplus\delta}}
  B 
  \nottight{\nottight{\nottight{=}}}
  \app
  {\EVAL{\salgebra{{\nottight{\nottight\uplus}}}\domres\delta\X}}
  B}.
\par\halftop\noindent\sloppy{\sc Substitution \opt{Value} Lemma}\\\noindent
Let \math\sigma\ be a substitution.
 If the variables
 that occur freely in \nlbmath{B\sigma} 
 belong to\/\, \DOM\delta, \ then:
 \\\noindent\LINEmath{
   \app
     {\app
        \EVALSYM
        {\salgebra\tightuplus\delta}}
     {B\sigma}
  \nottight{\nottight{\nottight{\nottight=}}}
  \displayapp
    {\displayapp
       \EVALSYM
       {\mediumheadroom\salgebra
        \nottight{\nottight{\nottight{\nottight\uplus}}}
        \inparentheses{
        \inparenthesesinline{
           \sigma
           \nottight{\nottight\uplus}
           \domres\id{\V\setminus\DOM\sigma}}
        \nottight{\nottight\circ}
        \EVAL{\salgebra\tightuplus\delta}}}}
    {\mediumheadroom B}
.}

\yestop\noindent
We are now going to define a new notion of validity 
of sets of sequents, \ie\ sets of lists of formulas.
As \nolinebreak this new kind of validity depends on a \vc\ \nlbmath R,
it is called ``\math R-validity\closequotefullstop
It provides the \fev s with an existential semantics given by their valuation 
\bigmaths{\hastype{\app{\app\epsilon e}\delta}{\FUNSET\Vsome\salgebra}},
and the \fuv s with a universal semantics by 
\bigmaths{\FUNDEF\delta\Vall\salgebra}.
The definition is top-down and the function \nlbmath\epsilon\ 
(having nothing to do with \hilbert's \nlbmath\varepsilon)
and the notion of an \strongexRval\ 
are to be explained in \sectref{section existential valuations},
which also contains examples illustrating \math R-Validity.
\begin{definition}[\math R-Validity, \K]\label{definition weak validity} \
Let \math R be a \vc.
Let \salgebra\ be a \semanticobject\ with valuation 
\PARFUNDEF\delta\V\salgebra. 
Let \math G be a set of sequents.
\\\noindent
\math G is\emph{\math R-valid in \nolinebreak\salgebra} 
\udiff\
there is \astrongexRval\ \bigmath e such that
\math G is \pair e\salgebra-valid.
\\\noindent
\math G 
is\emph{\pair e\salgebra-valid} \udiff\
\math G is \trip{\delta'}e\salgebra-valid
for all \math{\,\FUNDEF{\delta'}\Vall\salgebra\,}.
\\\noindent
\math G 
is\emph{\trip\delta e\salgebra-valid} \udiff\ 
\ \math G is valid in
\math
{\salgebra\nottight\uplus\epsilon\funarg e\funarg\delta\nottight\uplus\delta
}.
\\\noindent
\math G is\emph{valid in \math{\salgebra\tightuplus\delta}} 
\udiff\ \math\Gamma\ is valid in \math{\salgebra\tightuplus\delta}
for all \math{\Gamma\in G}.
\\\noindent
A sequent \math\Gamma\ is\emph{valid in \math{\salgebra\tightuplus\delta}} 
\udiff\
there is some formula listed in \nlbmath\Gamma\ that is valid in 
\nlbmath{\salgebra\tightuplus\delta}.
\\\noindent
Validity in a class of \semanticobject s
is understood as validity in each of the \semanticobject s of that class.
If we omit the reference to a special \semanticobject\
we mean validity in some fixed class \nolinebreak\K\
of \semanticobject s, such as the class of all \math\Sigmaoffont-structures 
or the class of  \herbrand\ \math\Sigmaoffont-structures.
\end{definition}
\vfill\pagebreak

\subsection{\protect\StrongexRVals}
\label{section existential valuations}
Let \salgebra\ be some \semanticobject.
We now define semantical counterparts of our \exRsubs, 
which we will call 
``\strongexRvals\closequotefullstop
As \astrongexRval\
plays the \role\ of a\emph{raising function}
(a dual of a \skolem\ function as defined in \cite{miller}), \ 
it \nolinebreak does not simply map each \fev\ 
directly to an object of \nlbmath\salgebra\ 
(of \nolinebreak the \nolinebreak same \nolinebreak type),
but may additionally read the values of some \fuv s under
an \salgebra-valuation \math{\FUNDEF\delta\Vall\salgebra}. \ 
More precisely,\,
\astrongexRval\
\,\nlbmath e\,
takes 
some restriction of \nlbmath\delta\
as a second argument, \ say
\math{\PARFUNDEF{\delta'}\Vall\salgebra} \ 
with \bigmaths{\delta'\subseteq\,\delta}. \ 
In \nolinebreak short:\\\noindent\LINEmaths{
  \FUNDEF e\Vsome{\PARFUNSET{\inpit{\PARFUNSET\Vall\salgebra}}\salgebra}
}.\par\noindent
Moreover, for each \fev\ \existsvari x{}, we require that 
the set \nlbmath{\DOM{\delta'}}
of \fuv s read by \nlbmath{\app e{\existsvari x{}}} is
identical for all \nlbmath\delta. \ 
This identical set will be denoted with 
\nlbmath{\revrelappsin{S_e}{\existsvari x{}}} below.
Technically, we require that there is some ``semantical relation''
\math{S_e\subseteq\Vall\tighttimes\Vsome} \ 
such that for all \math{\existsvari x{}\in\Vsome}: 
\\\noindent\LINEmath{
  \FUNDEF
    {\app e{\existsvari x{}}\ }
    {\ \inpit
       {\FUNSET
          {\revrelappsin{S_e}{\existsvari x{}}}
          \salgebra}}
    \salgebra     
.}\par\noindent
This means that \app e{\existsvari x{}} can read the value of
\nlbmath{\forallvari y{}} if and only if \bigmaths{
  \pair{\forallvari y{}}{\existsvari x{}}\tightin S_e
}.
Note that, for each   
\bigmath{
  \FUNDEF 
    e
    \Vsome
%    {\inpit
      {\PARFUNSET
        {\inpit
          {\PARFUNSET\Vall\salgebra}
        }
        \salgebra}%}
,}
at most one semantical relation
exists, namely
\par\noindent
\LINEmaths{
  S_e
  \nottight{\nottight{:=}}
  \setwith
    {\pair{\forallvari y{}}{\existsvari x{}}}
    {\existsvari x{}\tightin\Vsome
     \und
     \forallvari y{}\tightin\DOM
     {\app\bigcup{\DOM{\app e{\existsvari x{}}}}}}
}.\par\noindent
In some of the following definitions we are slightly more general because
we want to apply the terminology not only to \fev s but also to
\sfuv s.

\begin{definition}[Semantical Relation (\math{S_e})]
\label{definition semantical relation} \ 
The\emph{semantical relation for \math e} is
\par\noindent\LINEmaths{
  S_e
  \nottight{\nottight{:=}}
  \setwith
    {\pair y x}
    {x\tightin\DOM e
     \und
     y\tightin\DOM{\app\bigcup{\DOM{\app e x}}}}
}.\par\noindent
\math e \ is\emph{semantical}
\udiff\
\bigmath e is a partial function on \nlbmath\V\ such that 
for all \math{x\in\DOM e}:\par\noindent\LINEmaths{
  \FUNDEF{e\funarg x}{\inpit{\FUNSET{\revrelappsin{S_e}x}\salgebra}}\salgebra
}.\end{definition}\begin{definition}[\StrongexRVal]
\label{definition exval}\\Let \math R be a \vc\
and let \salgebra\ be a \semanticobject. \ \ 
\math e \ is an\emph{\strongexRval}
\udiff\
\bigmaths{\FUNDEF e\Vsome{\PARFUNSET{\inpit{\PARFUNSET\Vall\salgebra}}\salgebra}},
\ \math e is semantical, \ and
\math{R\cup S_e} is \wellfounded.
\end{definition}
Finally, we need the technical means 
to turn \astrongexRval\ \,\nlbmath e\,
together with a valuation \nlbmath{\delta} of the \fuv s
into
a valuation \nlbmath{\app{\app\epsilon e}\delta} of the \fev s:
\par\noindent\parbox{\textwidth}{%
\begin{definition}[\math\epsilon]\\
We define the function \ \ 
\noindent\math{\footroom
  \FUNDEF
    \epsilon
    {~~(\PARFUNSET\V{\PARFUNSET{(\PARFUNSET\V\salgebra)}\salgebra})~~}
    {~~\FUNSET{(\PARFUNSET\V\salgebra)~~}{~~\PARFUNSET{\V~~}{~~\salgebra}}~~}
}\par\noindent
for \LINEmath{~~~~~~~~~~~
 \PARFUNDEF e\V{\PARFUNSET{\inpit{\PARFUNSET\V\salgebra}}\salgebra},
 ~~~~~~\math{\PARFUNDEF\delta\V\salgebra}, 
 \,~~~\math{x\in\V}}
\par\noindent by\LINEmaths{\headroom
  \app{\app{\app\epsilon e}\delta}x
  :=
  \app{\app e x}{\domres\delta{\revrelappsin{S_e}x}}
}.
\end{definition}}
\vfill\pagebreak

\begin{example}[\math R-Validity]\label{ex validity}\sloppy
\newcommand\outdent{\hskip 12.7em\mbox{}}For 
\,\maths{\existsvari x{}\,\tightin\,\Vsome}, 
\,\maths{\forallvari y{}\,\tightin\,\Vall},\,
the sequent \bigmaths{\existsvari x{}\boldequal\forallvari y{}}{}
is \math\emptyset-valid in any \nlbmath\salgebra\ because we can choose 
\mbox{\math{S_e:=\Vall\tighttimes\Vsome}} and \math{
  e
  \funarg{\existsvari x{}}
  \funarg\delta
  :=
  \delta
  \funarg{\forallvari y{}}
}
for \math{\FUNDEF\delta\Vall\salgebra},
resulting in 
%\par\noindent\LINEmaths{
\bigmaths{
   \app
     {\app
        {\app\epsilon e} 
        \delta}
     {\existsvari x{}}
   =
   \app
     {\app e{\existsvari x{}}}
     {\domres\delta{\revrelappsin{S_e}{\existsvari x{}}}}
   =
   \app
     {\app e{\existsvari x{}}}
     {\domres\delta\Vall}
   =
   \app\delta{\forallvari y{}}
}.%\par\noindent
This means that \math\emptyset-validity of 
\math{
  \existsvari x{}\boldequal \forallvari y{}
} 
is the same as validity of 
\ \mbox{\math{
  \forall y
  \stopq
  \exists x
  \stopq
  x\boldequal y
}.} \ \ 
Moreover, note that
\math{
  \epsilon(e)(\delta)
}
has access to the \math\delta-value of 
\math{\forallvari y{}}
just as a raising function \nlbmath f for \nlbmath x 
in the raised (\ie\ 
dually Skolemized) version 
\math{ 
  f(\forallvari y{})
  \boldequal 
  \forallvari y{}
}
of
\bigmath{
  \forall y
  \stopq
  \exists x
  \stopq
  x\boldequal y
.}

Contrary to this, for \math{R:=\Vsome\tighttimes\Vall},
the same formula
\math{
  \existsvari x{}
  \boldequal 
  \forallvari y{}
} 
is not
\math R-valid in general
because then the required \wellfoundedness\ of
\math{R\cup S_e} (\cf\ \defiref{definition exval}) implies 
\bigmath{S_e\tightequal\emptyset,}
and the value of \nlbmath{\existsvari x{}} cannot depend on 
\math{\delta\funarg{\forallvari y{}}}
anymore, due to
\bigmaths{
  \app
    {\app 
       e
       {\existsvari x{}}}
    {\domres\delta{\revrelappsin{S_e}{\existsvari x{}}}}
  =
  \app
    {\app
       e
       {\existsvari x{}}}
    {\domres\delta\emptyset}
  =
  \app
    {\app
       e
       {\existsvari x{}}}
    \emptyset
}.
This means that \math{(\Vsome\tighttimes\Vall)}-validity of 
\math{
  \existsvari x{}
  \boldequal 
  \forallvari y{}
} 
is the same as validity of 
\ \mbox{\maths{
  \exists x
  \stopq
  \forall y
  \stopq
  x\boldequal y
}.} \ 
Moreover, note that
\math{
  \epsilon(e)(\delta)
}
has no access to the \math\delta-value of 
\math{\forallvari y{}}
just as a raising function 
\nlbmath c for \nlbmath x
in the raised version 
\math{ 
  c\boldequal \forallvari y{}
}
of
\mbox{\math{\exists x\stopq\!\forall y\stopq\!x\boldequal y.}}

For a more general example let
\math{G=
  \setwith
    {A_{i,0}\ldots A_{i,n_i-1}}
    {i\tightin I}
},
where for \math{i\in I} and \math{j\tightprec n_i} the \nlbmath{A_{i,j}} 
are formulas with \fev s from \math{\vec e} and \fuv s from \nlbmath{\vec u}.
Then \math{(\Vsome\tighttimes\Vall)}-validity of \nlbmath G 
means \hfill\math{
  \exists\vec e
  \stopq
  \forall\vec u
  \stopq
  \forall i
  \tightin I
  \stopq
  \exists j
  \tightprec n_i
  \stopq
  A_{i,j}
}\outdent\\
whereas \math\emptyset-validity of \nlbmath G 
means \hfill\math{
  \forall\vec u
  \stopq
  \exists\vec e
  \stopq
  \forall i
  \tightin I
  \stopq
  \exists j
  \tightprec n_i
  \stopq
  A_{i,j}
}\outdent\par
Also any other sequence of universal and existential quantifiers can
be represented by a \vc\ \nlbmath R, starting from the empty set and 
applying the \math\delta-rules from \sectref{subsection Rules}. \ 
A \nolinebreak
translation of a \vc\ \nlbmath R into a sequence of quantifiers may,
however, require a strengthening of dependences, in the sense that
a backwards translation would result in a \vc\ \nlbmath{R'}
with \bigmaths{R\subsetneq R'}.
This means that our framework can express logical dependences more
fine-grained than standard quantifiers.
\end{example}

\yestop\subsection{\CC s}\label{section choice-conditions}
\yestop\begin{definition}[\CC]\label{definition choice condition}
\par\noindent\math C is an\emph{\math R-\cc} \udiff\
\math R \nolinebreak is a \wellfounded\ \vc\ and 
\math C is a partial function from \nlbmath\Vsall\ 
into the set of formula-valued \mbox{\math\lambda-terms}, 
such that for all \math{\sforallvari y{}\in\DOM C}:
\noitem\begin{enumerate}\item
\label{item one definition choice condition}%
\bigmath{\freevari z{}\nottight{\refltransclosureinline R}\sforallvari y{}}
for all \math{\freevari z{}\in\VARfree{C\funarg{\sforallvari y{}}}}, \ and
\item
\label{item two definition choice condition}%
\bigmaths{\app C{\sforallvari y{}}}{} is of the form 
\bigmaths{\lambda v_0\stopq\ldots\lambda v_{l-1}\stopq B}, where
\par\noindent\LINEnomath{
\math{B} \nolinebreak 
 is a formula whose freely occurring variables from \nlbmath\Vbound\ 
 \nlnomath
are among
 \bigmaths{\{v_0,\ldots,v_{l-1}\}\subseteq\Vbound}{}}
\par\noindent
and where,
for 
\ \bigmaths{\hastype{v_0    }{\alpha_0    }}, \ \ldots, 
\ \bigmaths{\hastype{v_{l-1}}{\alpha_{l-1}}}, \ 
 we have
\par\noindent\LINEnomath{
\bigmaths{\hastype
   {\sforallvari y{}}
   {\FUNSET{\alpha_0}{\FUNSET\cdots{\FUNSET{\alpha_{l-1}}{\alpha_l}}}}}{} \ 
 for some type \nlbmath{\alpha_l},}
\par\noindent
and any occurrence of  \sforallvari y{} in \math B is of the form
 \bigmaths{\sforallvari y{}(v_0)\cdots(v_{l-1})}.
\end{enumerate}\end{definition}
\vfill\pagebreak

\yestop
\newcommand\itemelementaryexample{(c)}%
\begin{example}[\CC]\hfill{\em 
(continuing \examref{example higher-order choice-condition})}
\label{example choice-condition}\begin{enumerate}
\noitem\item[(a)] If \math R is a \wellfounded\ \vc\ that satisfies
\par\noindent\LINEnomath
{\sforallvari z a \math R \sforallvari y a \math R \sforallvari z b \math R 
 \sforallvari x a \math R \sforallvari z c \math R \sforallvari y c \math R 
 \sforallvari z d,}\par\noindent
then the \math C \nolinebreak of 
\examref{example higher-order choice-condition}
is an \math R-\cc, indeed.
\noitem\item[(b)] If some clever person would like to do the complete
quantifier elimination of \examref{example higher-order choice-condition}
by 
\\\LINEmath{\begin{array}[t]{l l r}
  \app{C'}{\sforallvari z d}
 &:=
 &\neg\Ppppdrei{\sforallvari x a}{\sforallvari y c}{\sforallvari z d}
\\\app{C'}{\sforallvari y c}
 &:=
 &\Ppppdrei{\sforallvari x a}{\sforallvari y c}{\sforallvari z d}
\\\app{C'}{\sforallvari x a}
 &:=
 &\neg\Ppppdrei
    {\sforallvari x a}{\sforallvari y c}{\sforallvari z d}
\\\end{array}}\par\noindent
then he would---among other things---need 
\sforallvari z d \transclosureinline R \sforallvari y c \transclosureinline R 
\sforallvari z d, by 
\defiref
{definition choice condition}(\ref{item one definition choice condition})
due to the values of \math{C'} at \math{\sforallvari y c} and 
\nlbmath{\sforallvari z d}.
This renders \nlbmath R \nonwellfounded.
Thus, this \math{C'} cannot be an \math R-\cc\ for any \nlbmath R. \ 
Note that the choices required by \nlbmath{C'} for 
\math{\sforallvari y c} 
and 
\nlbmath{\sforallvari z d} 
are in an unsolvable conflict, indeed.
\noitem\item[\itemelementaryexample] For a more elementary example, take 
\par\noindent\LINEmath{\begin{array}{l l l l l l l}
  \app{C''}{\sforallvari x{}}
 &:=
 &\inpit{\sforallvari x{}\tightequal\sforallvari y{}}
 &~~~~~~~~~
  \app{C''}{\sforallvari y{}}
 &:=
 &\inpit{\sforallvari x{}\tightnotequal\sforallvari y{}}
\\\end{array}}\par\noindent
Then \sforallvari x{} and \sforallvari y{} form a vicious circle of
conflicting choices for which no valuation can be found that is compatible
with \math{C''},
\cf\ \defiref{definition compatibility}, 
\lemmref{lemma compatible exists}. \ 
But \math{C''} is no \cc\ at all because there is no \wellfounded\ \vc\ \nlbmath R
that could turn it into an \math R-\cc.\end{enumerate}\end{example}

\begin{sloppypar}
\yestop\yestop\yestop\yestop\noindent
We now split our valuation \FUNDEF\delta\Vall\salgebra;\hskip.7em% 
while 
\FUNDEF\tau\Vwall\salgebra\ valuates the \wfuv s,\, 
\math\pi\ valuates the remaining \sfuv s. \ 
As the choices of \nlbmath\pi\ may depend on \nlbmath\tau,
the technical realization is similar to that of the dependence of
the \strongexRvals\ on the \fuv s,
as described in \sectref{section existential valuations}.
\end{sloppypar}

\yestop\yestop
\begin{definition}[Compatibility]
\label{definition compatibility}\par\noindent
Let \math C be an \math R-\cc,
\salgebra\ a \semanticobject, and \math e \astrongexRval. \ 
\math\pi\nolinebreak\ is \emph{\pair e\salgebra-compatible with \pair C R}
\udiff\begin{enumerate}\noitem\item
\label{item 1 definition compatibility}\FUNDEF
\pi\Vsall{\PARFUNSET{\inpit{\PARFUNSET\Vwall\salgebra}}\salgebra}
is semantical (\cf\ \defiref{definition semantical relation}) and
\\\math{R\cup S_e\cup S_\pi} is \wellfounded.
\noitem\item\label{item 2 definition compatibility}\headroom
For all \bigmaths{\sforallvari y{}\in\DOM C}{}
with \bigmaths{
  \app C{\sforallvari y{}}
  =
  \lambda v_0\stopq\ldots\lambda v_{l-1}\stopq B
}{} for a formula \math B,\\ 
for all \math{\FUNDEF\tau\Vwall\salgebra}, \  
for all \math{\FUNDEF\eta{\{\sforallvari y{}\}}\salgebra}, \
and 
for all \math{\FUNDEF\chi{\{v_0,\ldots,v_{l-1}\}}\salgebra},
\\setting 
\bigmath{\delta:={\app{\app\epsilon\pi}\tau\uplus\tau\uplus\chi}} and
\bigmath{\delta':=\eta\uplus\domres\delta{\V\setminus\{\sforallvari y{}\}}} 
(\ie\ \math{\delta'} is the \math\eta-variant of \nlbmath\delta):
\par\mbox{}\hskip2.5em
If \math{B} is \trip{\delta'}e\salgebra-valid, then 
   \math{B} is also \trip\delta e\salgebra-valid.
\end{enumerate}\end{definition}

\yestop\yestop\noindent
Roughly \nolinebreak speak\-ing, 
\itemref{item 1 definition compatibility} 
of this definition requires---for similar reasons as before---that 
the flow of information 
between variables expressed in \math R, \math e, and \math\pi\ is acyclic. 
%We need this to be able to instantiate the \wfuv s in lemma applications. 
\vfill\pagebreak\par
To understand \itemref{item 2 definition compatibility},
consider an \math R-\cc\ 
\nlbmath{C:=\{
\pair
{\sforallvari y{}}
{\ \lambda v_0\stopq\ldots\lambda v_{l-1}\stopq B}
\}},
which restricts the value of \nlbmath{\sforallvari y{}}
with the formula-valued \mbox{\math\lambda-term}
\ \mbox{\math{\lambda v_0\stopq\ldots\lambda v_{l-1}\stopq B}}. \ 
Then \math C simply requires 
that a different choice for the \app{\app\epsilon\pi}\tau-value of 
\sforallvari y{} cannot give rise to 
%a previously not given 
the validity of the formula \nlbmath B in
\bigmaths{
  \salgebra
  \nottight\uplus
  \app
    {\app\epsilon e}
    \delta
  \nottight\uplus
  \delta%\uplus\chi
}. \ 
Or---in other words---that \app{\app{\app\epsilon\pi}\tau}{\sforallvari y{}} 
is chosen such that
\math B becomes valid, whenever such a choice is possible. \ 
This is closely related to 
\label{cc s and Hilbert's epsilon}\hilbert's \math\varepsilon-operator
in the sense that \sforallvari y{} is given the value of 
\par\noindent\LINEmath{
  \lambda v_0\stopq\ldots\lambda v_{l-1}\stopq\varepsilon y\stopq
  \inpit{B\{\sforallvari y{}(v_0)\cdots(v_{l-1})\mapsto y\}}
}\par\noindent for a fresh bound variable \nlbmath y. 

As the choice for \nolinebreak\sforallvari y{}
depends on the other free variables of 
\maths{\lambda v_0.\,\ldots\lambda v_{l-1}.\,B}{} \ 
(\ie\nolinebreak\ the free variables of 
 \ \mbox{\math{
  \lambda v_0\stopq\ldots\lambda v_{l-1}\stopq\varepsilon y\stopq
  \inpit{B\{\sforallvari y{}(v_0)\cdots(v_{l-1})\mapsto y\}
  }
 }}~),
we included this dependence 
into the transitive closure of the \vc\ \nlbmath R
in \nolinebreak\defiref
{definition choice condition}(\ref{item one definition choice condition}). \ 
Therefore, the \wellfoundedness\ of \nlbmath R avoids the
conflict of \examref{example choice-condition}\itemelementaryexample.

%%%%%%%%%%%%%%%%%%%%%%%%%%%%%%%%%%%%%%%%%%%%%%%%%%%%%%%%%%%%%%%%%%%%%%%%%%%%
Note that the empty function \math\emptyset\ is an \math R-\cc\ for any 
\wellfounded\ \vc\ \nlbmath R. \ 
Furthermore, any \nlbmath\pi\ with 
\bigmaths{\FUNDEF\pi\Vsall{\FUNSET{\{\emptyset\}}\salgebra}}{}
is \pair e\salgebra-compatible with \nlbmath{\pair\emptyset R} due to 
\bigmaths{S_\pi\tightequal\emptyset}. \ 
Indeed, as stated in the following lemma, a compatible \nlbmath\pi\ 
always exists. This is due to \defiref
{definition choice condition}(\ref{item one definition choice condition})
and the \wellfoundedness\ of \bigmaths{R\cup S_e}{} (according to 
\defiref{definition exval}) \ 
and due to the restriction on the 
occurrence of \sforallvari y{} in \nlbmath B in \defiref
{definition choice condition}(\ref{item two definition choice condition}).
\begin{lemma}
\label{lemma compatible exists}
If\/ \math C is an \math R-\cc, \ 
\salgebra\ a \semanticobject, \  
and \bigmath e \astrongexRval, \ 
then there is some \math\pi\ 
that is \pair e\salgebra-compatible with \pair C R.
\end{lemma}

\yestop\yestop\noindent Just like the \vc\ \nlbmath R, 
the \math R-\cc\ \nlbmath C may grow during proofs.
This kind of extension together with a simple soundness condition plays
an important \role\ in inference:
\begin{definition}[Extension]
\pair{C'}{R'} is an\emph{extension of} \pair C R \udiff\
\math{C} is an \math{R}-\cc,
\math{C'} is an \math{R'}-\cc,
\bigmath{C\tightsubseteq C',}
and 
\bigmath{R\tightsubseteq R'.}
\end{definition}
\begin{lemma}[Extension]\label{lemma extension and compatibility}
Let \pair{C'}{R'} be an extension of \pair C R.
\\If\/ \math e is \astrongexRprimeval\ 
and \math\pi\nolinebreak\ 
is \pair{e}\salgebra-compatible with \nlbmath{\pair{C'}{R'}},
\\then \math e is also \astrongexRval\ and \math\pi\
is also \pair{e}\salgebra-compatible with \nlbmath{\pair{C}{R}}.
\end{lemma}

\yestop\yestop\noindent After global application of an 
\math R-substitution \nlbmath\sigma\ 
 we now have to update both \math R and \nlbmath C:
\begin{definition}[Extended \math\sigma-Update]\label{definition ex str s up}
Let \math C be an \math R-\cc\ and let \math\sigma\ be a substitution. \ 
The\emph{extended \math\sigma-update 
\bigmath{\pair{C'}{R'}} of \bigmath{\pair C R}}
is given by:
\\\LINEmaths{\begin{array}{l@{\ \ }l}
  C'
 &:=\ \ 
  \setwith{\pair x{B\sigma}}{\pair x B\tightin C\und x\tightnotin\DOM\sigma}, 
\\R'
 &\mbox
  {is the \strongsigmaupdate\ of \math R, \ \cf\ \defiref{definition update}.}
\\\end{array}}{}
\end{definition}
\begin{lemma}[Extended \math\sigma-Update]\label{lemm ex str s up}
If\/ \math C is an \math R-\cc, \ 
\math\sigma\ an \math R-substitu\-tion, \ 
and if \pair{C'}{R'} is the extended 
\math\sigma-update of \nlbmath{\pair C R}, \ 
then \math{C'} is an \math{R'}-\cc.\end{lemma}\vfill\pagebreak
%%%%%%%%%%%%%%%%%%%%%%%%%%%%%%%%%%%%%%%%%%%%%%%%%%%%%%%%%%%%%%%%%%%%%%%%%
\subsection{\protect\Validity C R}\label{section strong validity}
While the notion of \math R-validity 
(\cfnlb\ \defiref{definition weak validity})
already provides the \fev s with an existential semantics,
it fails to give the \sfuv s the proper semantics according to 
an \math R-\cc\ \nlbmath C. This deficiency is overcome in the following
notion of \mbox{``\validity C R\closequotecomma}
which---roughly speaking---requires
the following:
For arbitrary values of the \wfuv s, 
we must be able to choose values for
the \sfuv s satisfying \nlbmath C, and then we must be able to choose
values for the \fev s,
such that the sequents become valid.
Note that the dependences of these choices 
are restricted by \nlbmath R.
In \nolinebreak a \nolinebreak formal top down representation, this reads:

\begin{definition}[\Validity C R]\label{definition strong validity}
\noindent
Let \math C be an \math R-\cc, let \salgebra\ be a \semanticobject, 
and let \math G be a set of sequents. \ 
\noindent
\math G is\emph{\valid C R in \nolinebreak\salgebra} \udiff\
\math G is \validstrongtrip\pi e\salgebra\ for 
some \strongexRval\ \bigmath e and \ some
\bigmath\pi\ that is \pair e\salgebra-compatible with \nolinebreak\pair C R. \ 
\noindent
\math G is\emph{\validstrongtrip\pi e\salgebra} \udiff\ 
\math G is \displaytrip{\app{\app\epsilon\pi}\tau\uplus\tau}e\salgebra-valid
for each \math{\FUNDEF\tau\Vwall\salgebra}.
\end{definition}

\noindent
Notice that the notion of \validitystrongtrip\pi e{\salgebra} with 
\bigmaths
%{\DOM\pi\tightequal\Vsall}
{\FUNDEF\pi\Vsall{\PARFUNSET{\inpit{\PARFUNSET\Vwall\salgebra}}\salgebra}}{}
differs from \trip\delta e\salgebra-validity with 
\bigmaths
%{\DOM\delta\tightequal\Vall}
{\PARFUNDEF\delta\V\salgebra}
{}
as given in \defiref{definition weak validity}. \ 
Notice that \validity C R treats the \sfuv s properly, 
whereas \math R-validity 
of \defiref{definition weak validity} does not.

\yestop\yestop\yestop\noindent In our framework the formula \nolinebreak (E2) 
 of \sectref{section E2} looks like (E2\math') in the following lemma.\label
{section where E2' is}\begin{lemma}[\Validity C R of (E2\math')]\label
{lemma where E2' is}\label{lemma E2' valid} \ 
Let\/ \nlbmath C be an \mbox{\math R-\cc}. \\ 
\mbox{For \math{i\in\{0,1\}}}, \
let\/ \math{A_i} be a formula and\/ \math{\sforallvari x i\in\Vsall} \ with
\bigmaths{\app C{\sforallvari x i}={A_i\{x\tight\mapsto\sforallvari x i\}}},\\
\ \mbox{\math{\sforallvari x i\not\in\VAR{A_0,A_1}}}, \ 
\ \maths{\sforallvari x i\tightnotin\DOM R}, \ 
\bigmaths{\sforallvari x i\not\in
\VAR{\relapp C{\V\tightsetminus\{\sforallvari x i\}}}}. \ 
The formula\par\noindent\LINEmath{
  \forall x\stopq\inparenthesesinlinetight{A_0\tightequivalent A_1}
  \nottight{\nottight\implies}
  \sforallvari x 0
  \tightequal
  \sforallvari x 1}{\rm(E2\math')}\\\noindent is \valid C R.\end{lemma}

\yestop\noindent
Note that the conditions of \lemmref{lemma E2' valid} may simply be achieved by
taking\emph{fresh} \sfuv s \sforallvari x 0 and \sforallvari x 1 and adding
\bigmaths{\inpit{\VARfree{A_i\{x\tight\mapsto\sforallvari x i\}}
\tightsetminus\{\sforallvari x i\}}\times\{\sforallvari x i\}}{} 
to the current \vc.
Very roughly speaking, \lemmref{lemma E2' valid} holds 
because after choosing a value for 
\nolinebreak\sforallvari x 0
we can take the same value for \sforallvari x 1, simply because 
\sforallvari x 1 is new and can read all \wfuv s,
and especially those that \sforallvari x 0 reads.
We will actually do two proofs of \lemmref{lemma E2' valid}.
First, as an exercise for the reader, a semantical one right now, which is
complicated and ugly. And then in \examref{example formal proof}
a formal, nice, and short one in our calculus.

\yestop
\begin{proofqed}{\lemmref{lemma E2' valid}} \ Formally,
in this proof we would have to apply the 
Explicitness and the Substitution \opt{Value} Lemma
from \sectref\semanticssectiontitle\ several times, but we just
argue informally in a straightforward and intuitively clear manner.
Otherwise the proof would be even longer and more ugly. 
\vfill\pagebreak

Let \bigmaths
{B\nottight{\nottight{:=}}
  \inpit{\forall x\stopq\inparenthesesinlinetight{A_0\tightequivalent A_1}
  \nottight{\implies}
  \sforallvari x 0
  \tightequal
  \sforallvari x 1}}.
Let \salgebra\ be an arbitrary \semanticobject.
As universes are non-empty, there is some \strongexRval\ \bigmaths e{} with 
\bigmaths{S_e=\emptyset}. \\ 
By \lemmref{lemma compatible exists} there is some \math\pi\ 
that is \pair e\salgebra-compatible with \pair C R. \ 
Define \math{\pi'} by \\\LINEmaths{\app{\app{\pi'}{\sforallvari x 1}}\tau:=
\left\{\begin{array}{@{}l l@{}}
  \app{\app{\app\epsilon\pi}\tau}{\sforallvari x 0}
 &\mbox{if~~}\forall x\stopq\inparenthesesinlinetight{A_0\tightequivalent A_1}
  \mbox{ is \trip{\app{\app\epsilon{\pi}}\tau\uplus\tau}e\salgebra-valid.}
\\\app{\app{\app\epsilon\pi}\tau}{\sforallvari x 1}
 &\mbox{otherwise}
\\\end{array}\right\}
\mbox{ for }\FUNDEF\tau\Vwall\salgebra
},\\and \bigmaths{\app{\pi'}{\sforallvari y{}}:=
\app{\pi}{\sforallvari y{}}}{} for all \math
{\sforallvari y{}\in\Vsall\tightsetminus\{\sforallvari x 1\}}. \ 
By \bigmaths{\sforallvari x 1\not\in\VAR{A_0,A_1}},
we have \bigmaths{
 \app{\app{\app\epsilon{\pi'}}\tau}{\sforallvari x 0}
=\app{\app{\app\epsilon{\pi'}}\tau}{\sforallvari x 1}}{}
for all \FUNDEF\tau\Vwall\salgebra\ for which \ \mbox{\math
{\forall x\stopq\inparenthesesinlinetight{A_0\tightequivalent A_1}}} \ 
is \trip{\app{\app\epsilon{\pi'}}\tau\uplus\tau}e\salgebra-valid. \ 
Thus, \math B is \trip{\app{\app\epsilon{\pi'}}\tau\uplus\tau}e\salgebra-valid.
It remains to show that \math{\pi'} 
is \pair e\salgebra-compatible with \pair C R, too. \ 
As \math{\pi'} is obviously semantical, 
for \itemref{item 1 definition compatibility} of 
\defiref{definition compatibility} it suffices to show that 
\math{R\cup S_e\cup S_{\pi'}} is \wellfounded.
By \bigmaths{\sforallvari x 1\tightnotin\DOM R},
due to \bigmaths{S_e=\emptyset}{} and 
\bigmaths{\DOM{S_{\pi'}}\subseteq\Vwall},
we have \bigmaths{\sforallvari x 1\tightnotin\DOM{R\cup S_e\cup S_{\pi'}}}.
Therefore, it suffices to show that 
\math{R\cup S_e\cup \ranres{S_{\pi'}}{\V\setminus\{\sforallvari x 1\}}}
is \wellfounded. But this is \wellfounded\ as a subrelation of 
\math{R\cup S_e\cup S_{\pi}}, which is \wellfounded\ because \math\pi\ 
is \pair e\salgebra-compatible with \pair C R. \ 
It remains to show that \itemref{item 2 definition compatibility} of 
\defiref{definition compatibility} holds.
By \bigmaths{\sforallvari x 1\not\in
\VAR{\relapp C{\V\tightsetminus\{\sforallvari x 1\}}}},
as \math\pi\ 
is \pair e\salgebra-compatible with \pair C R, it suffices to show 
\itemref{item 2 definition compatibility} only for \sforallvari x 1. \ 
Let \math{\FUNDEF\tau\Vwall\salgebra} and 
\math{\FUNDEF\eta{\{\sforallvari x 1\}}\salgebra} be arbitrary. \ 
Set 
\bigmaths{\delta_0:={\app{\app\epsilon{\pi}}\tau\uplus\tau}},
\bigmaths{\delta_1:={\app{\app\epsilon{\pi'}}\tau\uplus\tau}}, 
\bigmaths{\delta'_i:=\eta\uplus\domres{\delta_i}
{\V\setminus\{\sforallvari x 1\}}}, and
\bigmaths{\delta''
:=\{\sforallvari x 0\mapsto\app\eta{\sforallvari x 1}\}\uplus\domres{\delta_0}
{\V\setminus\{\sforallvari x 0\}}}.
Suppose that \math{A_1\{x\tight\mapsto\sforallvari x 1\}} is 
\trip{\delta'_1}e\salgebra-valid. 
We have to show the claim that \math{A_1\{x\tight\mapsto\sforallvari x 1\}} is 
\trip{\delta_1}e\salgebra-valid. \ 
As \bigmaths{\delta'_0\tightequal\delta'_1},
\math{A_1\{x\tight\mapsto\sforallvari x 1\}} is 
\trip{\delta'_0}e\salgebra-valid. \ 
As \math\pi\ is \pair e\salgebra-compatible with \pair C R,
we have that \math{A_1\{x\tight\mapsto\sforallvari x 1\}} is 
\trip{\delta_0}e\salgebra-valid. \ 
If \ \mbox{\math
{\forall x\stopq\inparenthesesinlinetight{A_0\tightequivalent A_1}}} \ 
is not \trip{\delta_0}e\salgebra-valid,
then \ \mbox{\math{\delta_0\tightequal\delta_1}}, \ and the claim holds.
Otherwise, 
as \bigmaths{\sforallvari x 1\not\in\VAR{A_0,A_1}}{}
and \bigmaths{
 \domres{\delta_1'}{\V\setminus\{\sforallvari x 1\}}
=\domres{\delta_0 }{\V\setminus\{\sforallvari x 1\}}
=\domres{\delta_1 }{\V\setminus\{\sforallvari x 1\}}
}, we know that
\math{A_0\{x\tight\mapsto\sforallvari x 1\}} is 
\trip{\delta'_1}e\salgebra-valid and it suffices to show that 
\math{A_0\{x\tight\mapsto\sforallvari x 1\}} is 
\trip{\delta_1}e\salgebra-valid. \ 
By \bigmaths{\sforallvari x i\not\in\VAR{A_0,A_1}},
\math{A_0\{x\tight\mapsto\sforallvari x 0\}} 
is \trip{\delta''}e\salgebra-valid 
and it suffices to show 
(note that \math{
 \app{\app{\app\epsilon{\pi'}}\tau}{\sforallvari x 1}
=\app{\app{\app\epsilon{\pi }}\tau}{\sforallvari x 0}}) that 
\math{A_0\{x\tight\mapsto\sforallvari x 0\}} is 
\trip{\delta_0}e\salgebra-valid,
but this is the case indeed,
because \math\pi\ is \pair e\salgebra-compatible with \pair C R.\end{proofqed}

\yestop\yestop\yestop\yestop\noindent As already noted in \sectref
{section Instantiating Strong Free Universal Variables},
the single-formula sequent \app{Q_C}{\sforallvari y{}} 
of \defiref{definition Q} is a formulation of axiom\,(\math{\varepsilon_0})
of \sectref{section Why} in our framework.

\yestop\begin{lemma}[\Validity C R of \app{Q_C}{\sforallvari y{}}]\label
{lemma Q valid} \ 
Let \math C be an \math R-\cc. \\
Let \math{\sforallvari y{}\in\DOM C}. \ 
The formula\/ \nlbmath{\app{Q_C}{\sforallvari y{}}} is \valid C R.\\
Moreover, \math{\app{Q_C}{\sforallvari y{}}} is \validstrongtrip\pi e\salgebra\
for any \semanticobject\ \nlbmath\salgebra,
any \strongexRval\ \bigmaths e, and
any \bigmath\pi\ that is \pair e\salgebra-compatible with 
 \nolinebreak\pair C R.\end{lemma}
\begin{proofqed}{\lemmref{lemma Q valid}} \ 
Let \bigmaths{\app C{\sforallvari y{}}=
               \lambda v_0\stopq\ldots\lambda v_{l-1}\stopq 
               B}{} for a formula \math B. \ 
Then we have
\bigmaths{\app{Q_C}{\sforallvari y{}}\nottight{\nottight=}
  \forall v_0\stopq\ldots\forall v_{l-1}\stopq
  \inparentheses{
    ~\exists y\stopq
    \inpit{B\{\sforallvari y{}(v_0)\cdots(v_{l-1})
    \mapsto y\}}
    \nottight{\nottight\implies}
    B~}
    }{}
for an arbitrary \math
{y\in\Vbound\tightsetminus\VAR{\app C{\sforallvari y{}}}}. \ \ 
For \bigmath\pi\ being \pair e\salgebra-compatible with \nolinebreak\pair C R,
the \validitystrongtrip\pi e\salgebra\ follows now directly from
\itemref{item 2 definition compatibility} of 
\defiref{definition compatibility}, \
according to the short discussion following \defiref{definition Q}. \ 
The rest is trivial.\end{proofqed}\vfill\pagebreak

\subsection{Reduction}\label{section reduction}
%\yestop\noindent 
Reduction is the reverse of consequence. 
It is the backbone of logical reasoning, 
especially of abduction and goal-directed deduction.
Our version of reduction does not only reduce a set of problems
to another set of problems but also guarantees that the solutions of
the latter also solve the former; where ``solutions'' means
the valuations for the rigid variables, \ie\ for the \fev s and the \sfuv s.

\begin{definition}[Reduction]
\label{definition strong reduction}\\
Let \math C be an \math R-\cc. \\Let \salgebra\ be a \semanticobject, 
and let \math{G_0} and \math{G_1} be sets of sequents.\\
\noindent{\em \math{G_0} \stronglyreduces C R < to \nolinebreak
\math{G_1} in \salgebra\/} \udiff\\
for any \strongexRval\ \math e
and any \math\pi\ that is \pair e\salgebra-compatible with \pair C R: \
\\\LINEnomath
{if   \math{G_1} is \validstrongtrip\pi e\salgebra,
 then \math{G_0} is \validstrongtrip\pi e\salgebra.}
\end{definition}

\begin{theorem}[Reduction]\label{theorem strong reduces to}\\\headroom
Let \math C be an \math R-\cc; \ 
\salgebra\ a \semanticobject; \ 
\math{G_0}, \math{G_1}, \math{G_2}, 
and \math{G_3} sets of sequents.
\\\headroom{\bf 1.\,(Validity) }
\begin{tabular}[t]{@{}l@{}}
If \math{G_0} \stronglyreduces C R < to \math{G_1} 
in \nolinebreak\salgebra\ 
and \math{G_1} is \valid C R in \nolinebreak\salgebra,
\\then \math{G_0} is \valid C R in \nolinebreak\salgebra, too.
\\\end{tabular}
\\\headroom{\bf 2.\,(Reflexivity) }
In case of
\ \math{
  G_0\tightsubseteq G_1
}:
\ \math{G_0} \stronglyreduces C R < to \math{G_1} 
in \nolinebreak\salgebra.
\\\headroom{\bf 3.\,(Transitivity) }
\begin{tabular}[t]{@{}l@{}}
If \math{G_0} \stronglyreduces C R < to \math{G_1} in 
\nolinebreak\salgebra\
\ and \ \math{G_1} \stronglyreduces C R < to \math{G_2} in 
\nolinebreak\salgebra,
\\then \math{G_0} \stronglyreduces C R < to \math{G_2} 
in \nolinebreak\salgebra.
\\\end{tabular}
\\\headroom{\bf 4.\,(Additivity) }
\begin{tabular}[t]{@{}l@{}}
If \math{G_0} \stronglyreduces C R < to \math{G_2} 
in \nolinebreak\salgebra\ 
\ and \ \math{G_1} \stronglyreduces C R < to \math{G_3} 
in \nolinebreak\salgebra,
\\then \math{G_0\tightcup G_1} \stronglyreduces C R < 
to \math{G_2\tightcup G_3} in \nolinebreak\salgebra.
\\\end{tabular}
\\\headroom{\bf 5.\,(Monotonicity) }
For \pair{C'}{R'} being an extension of\/ \pair C R:
\\\headroom\mbox{}\hfill
\begin{tabular}[t]{@{}l@{}}
(a) If   \math{G_0} is \valid{C'}{R'} in \nolinebreak\salgebra,
then \math{G_0} is \valid C R     in \nolinebreak\salgebra.
\\\headroom(b)
If \math{G_0} \stronglyreduces C R < to \math{G_1} in \nolinebreak\salgebra,
then \math{G_0} \stronglyreduces{C'}{R'}{<'} to \math{G_1} 
in \nolinebreak\salgebra.
\\\end{tabular}
\\\headroom{\bf 6.\,(Instantiation) }
\begin{tabular}[t]{@{}l@{}}
For an \Rsub\ \nlbmath{\sigma}\ on \math{\Vsomesall}, \ 
the extended \math\sigma-update\/ \nolinebreak\pair{C'}{R'} 
\\of\/ \nolinebreak\pair C R, \ 
and for \bigmaths{O:=
\DOM C\cap\DOM\sigma\cap\revrelapp{\refltransclosureinline R}{\VARsall{G_0,G_1}}
}:
\\\end{tabular}
\\\headroom\mbox{}\hfill
\begin{tabular}[t]{@{}l@{}}
  (a) If   \math{G_0\sigma\cup\math{\inpit{\relapp{Q_C}O}\sigma}} 
is \valid{C'}{R'} in \nolinebreak\salgebra, \ 
then \math{G_0} is \valid C R in \nolinebreak\salgebra.
\\\headroom(b)
If \math{G_0} \stronglyreduces C R{} to \math{G_1} in \nolinebreak\salgebra, \
then \math{G_0\sigma} \stronglyreduces{C'}{R'}{} to 
\math{G_1\sigma\cup\math{\inpit{\relapp{Q_C}O}\sigma}}
in \nolinebreak\salgebra.
\\\end{tabular}
% \\\headroom{\bf 6.\,(Instantiation) }
% \begin{tabular}[t]{@{}l@{}}
% For an \Rsub\ \nlbmath{\sigma}\ on \math{\Vsome\cup\Vsall}, \ 
% the extended \math\sigma-update\/ \nolinebreak\pair{C'}{R'} 
% \\of\/ \nolinebreak\pair C R, \ 
% and for \math O, \math N with 
% \bigmaths{O\subseteq\DOM C\cap\DOM\sigma\subseteq O\uplus N},
% \\\maths{N\subseteq\DOM C\setminus O},
% \bigmaths{\DOM C\cap\relapp{\transclosureinline R}N\subseteq N}, 
% and
% \ \maths{N\cap\VAR{G_0,G_1}=\emptyset}:
% \\\end{tabular}
% \\\headroom\mbox{}\hfill
% \begin{tabular}[t]{@{}l@{}}
%   (a) If   \math{G_0\sigma\cup\math{\inpit{\relapp{Q_C}O}\sigma}} 
% is \valid{C'}{R'} in \nolinebreak\salgebra, \ 
% then \math{G_0} is \valid C R in \nolinebreak\salgebra.
% \\\headroom(b)
% If \math{G_0} \stronglyreduces C R{} to \math{G_1} in \nolinebreak\salgebra, \
% then \math{G_0\sigma} \stronglyreduces{C'}{R'}{} to 
% \math{G_1\sigma\cup\math{\inpit{\relapp{Q_C}O}\sigma}}
% in \nolinebreak\salgebra.
% \\\end{tabular}
\end{theorem}

\begin{proofqed}{\theoref{theorem strong reduces to}}\\
\lititemfromtoref 1 5 are the \lititemfromtoref 1 5 of 
\litlemmref{2.31} of \cite{wirthcardinal}.
\\\lititemref 6 follows from \litlemmref{B.6} of \cite{wirthcardinal}
when we set the meta variable \math N of \litlemmref{B.6} to 
\bigmaths{\DOM C\cap\relapp
  {\refltransclosureinline R}
  {\inpit{\DOM C\cap\DOM\sigma}\setminus O}}.
\end{proofqed}

\yestop\noindent
\lititemfromtoref 1 5 of 
\theoref{theorem strong reduces to} are straightforward.
\lititemref 6 is only technically complicated.
Roughly speaking, 
the idea behind \lititemref 6 is that reduction stays invariant
under global application of the substitution \nlbmath\sigma\ on rigid variables,
provided that we change from \pair C R to its 
extended \math\sigma-update \pair{C'}{R'}
and that, 
in case that \nlbmath\sigma\ replaces some \sfuv\ \nlbmath{\sforallvari y{}}
constrained by the \cc\ \nlbmath C, \ 
we can establish that this is a proper
choice by showing \nlbmath{\inpit{\app{Q_C}{\sforallvari y{}}}\sigma}, \ 
\cf\ \defiref{definition Q}\@. \ 
The rest of this \sectref{section reduction} 
will give further explanation on the 
application of \theoref{theorem strong reduces to} and 
especially of \lititemref 6.

\yestop\begin{example}
[Instantiation with \ \math{\DOM C\cap\DOM\sigma=\emptyset}]\label
{example Ackermann}\\
For a simple application of \theoref{theorem strong reduces to}(6b),
where no \sfuv s occur and only a \fev\ is instantiated,
let us have a glimpse at
the example proof of \cite[\litsectref{3.3}]{wirthcardinal}\@. \ \ 
Let \math{G_0} be the proposition we want to prove, namely
\bigmaths{\{\ 
    {\app{\app{\existsvari z 0}{\wforallvari x 0}}{\wforallvari y 0}}
    \prec
    {\ackpp{\wforallvari x 0}{\wforallvari y 0}}
\ \}},
which says that \ackermann's function has a lower bound that is to be
determined during the proof.
Moreover, let \math{G_1}\,---together with \vc\ \nlbmath R and 
\mbox{\math R-\cc\ \nlbmath\emptyset}\,---represent 
the current state of the proof. \ 
Then \math{G_0} \pair\emptyset R-reduces to \math{G_1}. \ 
Moreover, in the example, 
\math{G_1} reduces to a known lemma when we apply the substitution \bigmaths
{\sigma:=\mbox{\math{\{\existsvari z 0\mapsto\lambda x.\,\lambda y.\,y\}}}}. \
Now, \theoref{theorem strong reduces to}(6b) says that 
the instantiated (and \math{\lambda\beta}-reduced) theorem \bigmaths{
\{\ {\wforallvari y 0}
    \prec
    {\ackpp{\wforallvari x 0}{\wforallvari y 0}}
\ \}}{}
\pair\emptyset{R}-reduces to the instantiated proof state \bigmaths{G_1\sigma}{}
and thus is \pair\emptyset{R}-valid by 
\theoref{theorem strong reduces to}(3,1). \ 
Note that in this case the extended \mbox{\math\sigma-update} 
of \nlbmath{\pair\emptyset R} is 
\pair\emptyset R itself, and we have \bigmaths{O\tightequal\emptyset}{}
due to \bigmaths{\DOM C\cap\DOM\sigma=\emptyset}.
Moreover, by \theoref{theorem strong reduces to}(6a),
also the original 
\bigmaths{\{\ 
    {\app{\app{\existsvari z 0}{\wforallvari x 0}}{\wforallvari y 0}}
    \prec
    {\ackpp{\wforallvari x 0}{\wforallvari y 0}}
\ \}}{} 
is known to be \pair\emptyset{R}-valid, 
but who would be interested in this weaker result now?
\end{example}
\begin{example}[\Validity C R of (E2\math')]\hfill{\em (continuing
\lemmref{lemma where E2' is})}\label
{example formal proof}\sloppy\\\noindent
Instead of the ugly semantical
proof of (E2\math') of \lemmref{lemma where E2' is} 
in \sectref{section where E2' is},
let us give a formal proof of (E2\math') in our framework on a very abstract
level by applying \theoref{theorem strong reduces to}. 
We will reduce the set containing
the single-formula sequent of the formula (E2\math') to a valid set.
This will complete our proof by 
\lititemref 1 of \theoref{theorem strong reduces to}. \ 
In the following, be aware of
the requirements on occurrence of the variables as described 
in  \lemmref{lemma where E2' is}. \ 
We extend \pair C R with a fresh variable \nlbmath{\sforallvari y{}}
with \bigmaths{\app{C'}{\sforallvari y{}}\nottight{\nottight{:=}}
\inparenthesesoplist{\inpit{
        \forall x\stopq\inparenthesesinlinetight{A_0\tightequivalent A_1}
      \implies\sforallvari y{}\tightequal\sforallvari x 0}
  \oplistund\inpit{
        \neg\forall x\stopq\inparenthesesinlinetight{A_0\tightequivalent A_1}
      \implies A_1\{x\mapsto\sforallvari y{}\}}}}.
Of \nolinebreak course, to satisfy \defiref
{definition choice condition}(\ref{item one definition choice condition}),
the current \vc\ \nlbmath R must be extended to
\bigmaths{R'\nottight{\nottight{:=}}
R\nottight\cup\inpit{\VARfree{A_0,A_1}\cup\{\sforallvari x 0\}}
\times\{\sforallvari y{}\}}.
Note that, if we had done this extension during the proof,
we would have needed \lititemref{5b} to keep reduction invariant,
but as there is no reduction sequence given yet, it suffices to use
\lititemref{5a} instead. Similarly, instead of \lititemref{6b},
we apply \lititemref{6a}, with 
\bigmaths{\sigma:=\{\sforallvari x 1\mapsto\sforallvari y{}\}}.
Then we have \bigmaths{O\subseteq\DOM C\cap\DOM\sigma=\{\sforallvari x 1\}}.
For \pair{C''}{R''} being the extended \math\sigma-update of 
\nlbmath{\pair{C'}{R'}},
\lititemref{6a} says that it suffices to show 
\pair{C''}{R''}-validity of the set with the two single-formula
sequents \ \mbox{\math{
  \forall x\stopq\inparenthesesinlinetight{A_0\tightequivalent A_1}
  \implies
  \sforallvari x 0
  \tightequal
  \sforallvari y{}
}} \ 
and \bigmaths{\inpit{\app{Q_{C'}}{\sforallvari x 1}}\sigma}.
The latter sequent reads
\ \mbox{\math{\inpit{\exists x\stopq
    A_1\{x\mapsto\sforallvari x 1\}\{\sforallvari x 1\mapsto x\}
    \implies
    A_1\{x\mapsto\sforallvari x 1\}}\sigma
}}, \ \ie\nolinebreak\ 
\ \mbox{\math{\exists x\stopq
    A_1\implies A_1\{x\mapsto\sforallvari y{}\}
}}. \ 
But a simple case analysis on \bigmaths{
\forall x\stopq\inparenthesesinlinetight{A_0\tightequivalent A_1}}{}
shows that 
the whole set \pair{C''}{R''}-reduces to 
\\\displayset{
\exists x\stopq A_0\nottight\implies A_0\{x\mapsto\sforallvari x 0\};
~~~~~\hfill
\inparenthesesoplist{
\exists\boundvari y{}\stopq
\inparenthesesoplist{
    \inpit{
        \forall x\stopq\inparenthesesinlinetight{A_0\tightequivalent A_1}
      \implies
        \boundvari y{}\tightequal\sforallvari x 0}
  \oplistund
    \inpit{
        \neg\forall x\stopq\inparenthesesinlinetight{A_0\tightequivalent A_1}
      \implies
        A_1\{x\mapsto y\}}}\!\!\!
\oplistimplies
\inparenthesesoplist{
    \inpit{
        \forall x\stopq\inparenthesesinlinetight{A_0\tightequivalent A_1}
      \implies
        \sforallvari y{}\tightequal\sforallvari x 0}
  \oplistund
    \inpit{
        \neg\forall x\stopq\inparenthesesinlinetight{A_0\tightequivalent A_1}
      \implies
        A_1\{x\mapsto\sforallvari y{}\}}}}\!\!\!},\\
\ie\nolinebreak\ to \bigmaths{\{\app{Q_{C''}}{\sforallvari x 0}; \ \ 
\app{Q_{C''}}{\sforallvari y{}}\}},
which is \pair{C''}{R''}-valid by \lemmref{lemma Q valid}. \ 
(Note that by \lititemref 4 of \theoref{theorem strong reduces to}
 it would have been sufficient to show that each of the the formulas
 of the set \pair{C''}{R''}-reduces to some \pair{C''}{R''}-valid set.) \ 
Thus, \math{\mbox{(E2\math')}\sigma} is \pair{C''}{R''}-valid.
By \lititemref{6a} this means that (E2\math') is \pair{C'}{R'}-valid, and
by \lititemref{5a} this means that (E2\math') is \pair{C}{R}-valid, as
was to be shown.
Note that we have \bigmaths
{R'':=R'\cup\{\pair{\sforallvari y{}}{\sforallvari x 1}\}},
so that \math\sigma\ is an \math{R'}-substitution. Indeed, the graph 
of \nlbmath{R''} is acyclic:\\\LINEmaths{\xymatrix{
   {\sforallvari x 0}
   \ar[rrr]_{R'}
&&&{\sforallvari y{}}
   \ar[rrr]_{R''}
&&&{\sforallvari x 1}
\\{\VARfree{A_0}}\ar[u]_{\transclosureinline R}\ar[urrr]_{R'}
&&&&&&{\VARfree{A_1}}\ar[u]_{\transclosureinline R}\ar[ulll]_{R'}
}}{}\pagebreak\end{example}

\yestop\yestop\thepeinsexample

\subsection{On the Design of Similar Operators}\label
{subsection On the Design of Similar Operators}

In \nolinebreak\sectref{section requirement specification}
we already mentioned that the semantic free-variable framework 
for our \nlbmath\varepsilon\ may 
serve as the paradigm for the design of other operators
similar to our version of the \nlbmath\varepsilon. \ 
In this \nolinebreak\sectref{subsection On the Design of Similar Operators},
we give some general hints on the two screws which may be turned to
achieve the intended properties of such new operators.

The one screw to turn is the definition of \pair C R-validity.
For instance, 
the ``some \nlbmath\pi'' in \defiref{definition strong validity}
is something we can play around with.
Indeed, in \cite
[\litdefiref{5.7} (\litdefiref{4.4} in short version)]{wirthgreen}, 
we can read ``any \nlbmath\pi''
instead, which is just the opposite extreme; \ 
for which (E2\math') of \lemmref{lemma E2' valid}
is valid \uiff\ \mbox{\math
{\exists!x.\,A_0\oder\exists x.\forall y.\inpit{x\boldequal y}}}. \ 
In \nolinebreak 
between of both extremes, we could design operators tailored for 
generalized quantifiers (\eg\nolinebreak\ with cardinality specifications) 
or for the 
special needs of specification and computation of semantics of discourses in 
natural language. 
Note that the changes of our general framework
for these operators would be quite moderate: In any case, it is
``any \nlbmath\pi'' what we read in the 
important \lemmref{lemma Q valid}
and the crucial \defiref{definition strong reduction}. \ 
{Roughly speaking, only 
 \theoref{theorem strong reduces to}(6a) 
 for the case of \bigmath{O\tightnotequal\emptyset} 
 as well as
 \theoref{theorem strong reduces to}(5a) 
 would become false for a different 
 choice on the quantification of \nlbmath\pi\
 in \defiref{definition strong validity}.} \ 
The reason why we prefer ``some \nlbmath\pi'' to ``any \nlbmath\pi'' here 
and in \cite{wirthcardinal} is that ``some \nlbmath\pi'' 
results in more valid formulas (\eg\ \nolinebreak(E2\math')) 
and makes theorem proving easier. \ 
Contrary to ``any \nlbmath\pi'' and to all semantics in the literature, \ 
``some \nlbmath\pi'' frees us from considering all possible choices:
We just have to pick a single arbitrary one and fix it in a proof step. \
Moreover, ``some \nlbmath\pi'' is very close to \hilbert's intentions
on \math\varepsilon-substitution as described best in 
\cite[\Vol\,II, \litsectref{2.4}]{grundlagen}.

The other screw to turn is the definition of compatibility.
For instance, by modifying \itemref{item 2 definition compatibility} of
\defiref{definition compatibility} we can strengthen the notion of
compatibility in such a way that \nlbmath{\app\delta{\sforallvari y{}}}
has to pick the\emph{smallest} value such that \math B becomes
\trip\delta e\salgebra-valid. \ 
%As problems with substitutability within \math\varepsilon-terms
%do not appear in our framework (where committed choice is represented
%by term-sharing), 
With that modification of compatibility it would 
be interesting to model the failed trials of \hilbert's group to 
show termination of \math\varepsilon-substitution in arithmetic
before \cite{ackermann-consistency-of-arithmetic}
as described in \cite[\Vol\,II, \nolinebreak\litsectref{2.4}]{grundlagen}.

All in all, in our conceptually disentangled framework for the
\nlbmath\varepsilon, there are at least these two well-defined and conceptually
simple screws to turn for a convenient 
adjustment to achieve similar operators for different purposes.

\vfill\pagebreak
%%%%%%%%%%%%%%%%%%%%%%%%%%%%%%%%%%%%%%%%%%%%%%%%%%%%%%%%%%%%%%%%%%%%%%%%%%%
\section{Examples and Discussion on Philosophy of Language}\label
{section philosophy of language}

\subsection{Motivation and Overview}\label{section motive}
In this \sectref{section philosophy of language},
we exemplify our version of \hilbert's \nlbmath\varepsilon\
with several linguistic standard examples. 
The reason for choosing 
philosophy of language and the semantics of sentences in natural language
as the field for our examples is threefold:%
\newcommand\itemreflinguists{(\math\daleth)}%
\begin{enumerate}

\noitem\item[(\math\aleph)] These examples are simple and easily comprehensible,
even without linguistic expertise.
Moreover, they provide interesting and relevant test cases
for descriptive terms and their logical frameworks.

\noitem\item[(\math\beth)] The choice of our examples is natural due to 
the close relation of our \nlbmath\varepsilon\ to semantics for indefinite 
(and definite) articles and anaphoric pronouns in some natural languages.
\par
(We ignore, however, the generic, qualitative, metaphoric, and pragmatic effects
 of these indefinite determiners; \cf\ \sectref{section proviso}.)
\noitem\item[\itemreflinguists]
We hope that linguists find our solutions to these standard examples 
interesting enough to evaluate our semantics on its usefulness for 
developing tools that may help to represent and compute the semantics
of sentences and discourses in natural language.
\par
(Although the careful reader will find some method in our
 preference for certain representations, 
 it would go far beyond the scope of this \daspaper\ to present
 concrete procedures for generating different representational variants and 
 to decide on which of them to prefer.)
\noitem\end{enumerate}
We will proceed as follows:
In \sectref{section russell meinong} we introduce to the 
description of the semantics of determiners in natural languages, and
show that the \math\varepsilon\ is useful for it.
In \sectref{section linguistic literature}
we have a brief look at
the linguistic literature on \hilbert's \nlbmath\varepsilon.
In \sectref{section critical cases} we discuss cases that are 
difficult to model with our \nlbmath\varepsilon, such as 
\henkin\ quantifiers and 
cyclic choice in \bachpeterssentence s.
We look at problems with right-unique
\nlbmath\varepsilon\ in \nlbsectref{section geurts}, 
at donkey sentences in \sectref{section donkey}, 
and at the difficulty of 
capturing semantics of natural language with quantifiers
in \nolinebreak\sectref{section quantifiers adequate}.

To speed our hope expressed in \itemreflinguists\nolinebreak\ above,
we try to make this \sectref{section philosophy of language}
accessible without reading the 
formally involved previous 
\sectref{section formal discussion}. \ 
Accordingly, we remind or inform the reader of the following: \ 
We apply \smullyan's classification (\cfnlb\ \cite{smullyan}) 
of problem-reduction rules into
\math\alpha, \math\beta, \math\gamma, and \math\delta, and 
call the quantifiers eliminated 
and the variables introduced by \math\gamma- and \math\delta-steps,\emph
{\math\gamma-} and\emph{\math\delta-quantifiers} and\emph
{free \math\gamma-} and\emph{free \math\delta-variables}, respectively. \ 
\Fev s (written \existsvari x{}) are implicitly existentially quantified. \ 
\mbox{\Wfuv s (\wforallvari x{})} are implicitly universally quantified. \ 
The structure of the quantification is represented in a \vc. \ 
A\emph\vc\ is  a directed acyclic graph on free variables. \ 
The value of a free variable may transitively 
depend on the predecessors in the \vc, with the exception of the 
\wfuv s that may always take arbitrary values. \ 
Moreover, a \sfuv\ such as \nlbmath{\sforallvari x{}} 
is existentially quantified
but must take a value that makes its \cc\ \nlbmath{\app C{\sforallvari x{}}}
true---if \nolinebreak such a choice is possible. \ 
In \nolinebreak problem reduction, \wfuv s behave as constant parameters,
\fev s may be globally instantiated with any term that does not violate
the current \vc, and the instantiation of \sfuv s must additionally satisfy
the current \cc. \ %provided that it can be satisfied at all.
Furthermore, a sequent is a list of formulas which
denotes the disjunction of these formulas.
\vfill\pagebreak

\newcommand\reductiontheorem
[1]{\theoref{theorem strong reduces to}(#1)}
\newcommand\myepsisection{\subsection}
\newcommand\myepsisubsection{\subsubsection}
\sectionproandcontrareference
\vfill\pagebreak
\subsection
{A brief look at the Linguistic Literature on the \math\varepsilon}\label
{section linguistic literature}
In this \sectref{section linguistic literature},
we have a brief look at
the linguistically motivated literature on \hilbert's \nlbmath\varepsilon,
which goes beyond our discussion in \litsectrefstwoandthreepartone\@. \ 
The usefulness of \hilbert's \nlbmath\varepsilon\ for the description of
the semantics of natural language is simultaneously threatened 
by\emph{right-uniqueness} and\emph{uncommitted choice}, which 
seem to be opposite threats like \scylla\ and \charybdis,
hard to pass by in between even for brave \ulysses.
\begin{description}\item[Right-Uniqueness: ]
A right-unique behavior of the \math\varepsilon\ 
is a problem in natural language.
For example, the 
same phrase modeled as an \math\varepsilon-term 
does not necessarily denote the same object.
Indeed, it may necessarily denote two different ones as in 
``If\emph{a bishop} meets\emph{a bishop}, \ldots\closequotefullstop 
\end{description}
Based on \ND\ (\cf\ \cite{gentzen}, \cite{prawitznatural}), \ 
\meyerviolname\ presents in his \PhDthesis\ \cite{meyervioldiss} 
most interesting results on 
the \math\varepsilon\ in intuitionistic logic and a lot of 
fascinating ideas on how to use it for computing the 
semantics of sentences in natural language. \ 
The latter ideas, however, 
suffer from a right-unique behavior of the \nlbmath\varepsilon. \ 
We will discuss more problems with 
the right-uniqueness requirement in 
\sectref{section geurts} along \cite{geurts-one}.
\begin{description}
\item[Uncommitted Choice: ]
A major advantage of reference in natural language is the possibility to
refer to an object a second time.
Thus, 
the \math\varepsilon\ can hardly be of any use in semantics of natural 
language without the possibility to express committed choice; \hskip.2em
\cfnlb\ \sectref{section committed choice}. \hskip.4em 
Note, however, that
---~to express committed choice~---
we need right-uniqueness
unless we replace the \math\varepsilon-terms with \sfuv s; \hskip.2em
\cfnlb\ \sectrefs{section committed choice}{section my assumption}.
\end{description}
Already in 1993,
\eijckname\ addressed the double problem of \scylla\ and \charybdis\
in the first part of
the following sentence:
\begin{quote}
``{%
What we want, instead, 
is to employ different choice functions as we go along,
and to let the interpretation process fail in case no appropriate choice 
of \math\varphi\ is possible because there are no \nlbmath\varphi s.%
}''
\getittotheright{\cite[\p\,242\f]{eijck-epsilon}}
\end{quote}
The second part of this sentence, however, is a judgment
contra the \math\varepsilon,
which we cannot accept: 
If we want to model a natural language discourse,
we have to introduce a reference object even if currently no salient 
object satisfies the \cc\ of its \sfuv;
moreover, even for the round quadrangle we have to introduce
an object because we cannot talk about it otherwise.

\heusingername\
seems to take the  
first part of \eijck's sentence as a task instead of a problem
description:
In \citet{heusingerepsilon},
the right-uniqueness of the 
\nlbmath\varepsilon\ is kept, but the usefulness for describing the
semantics of natural language is improved by adding a situational index
to the \mbox{\math\varepsilon-symbol} that makes it possible to 
denote different choice functions explicitly; \hskip.2em
\cfnlb\ (19a\math') in our 
\sectref{section donkey} for an example. \ 
We will refer to this indexed \nlbmath\varepsilon\ as 
{\em``\heusinger's indexed 
 \mbox{\math\varepsilon-operator}\closequotefullstop} \ 
It \nolinebreak
already occurs in the English draft 
paper \cite{heusingerindefinitesold}. \ 
The book \cite{heusingerepsilon}, however, 
is a German monograph on 
applying \hilbert's epsilon to the semantics 
of noun phrases and pronouns in natural language,
with a focus on salience.
\heusinger's indexed \math\varepsilon-operator is used
to describe the definite as well as the indefinite article  
in specific as well as non-specific contexts, resulting in four different
representations.\footnote
{{\bf(Do Salience, Specifity, and Uniqueness Determine Definiteness?)}
\par\noindent\label
{footnote indexed epsilon operator one}{\em Salience}\/
is the property of being known and prominent in discourse.
{\em Specifity} is a property concerning 
the referential status for a speaker,
expressing that he has a specific object in mind.
Salience, specifity, and uniqueness are important aspects
immanent in the distinction of definite and indefinite forms.
 In \cite[\p 1]{heusingerepsilon},
 we find the thesis that 
 definiteness of articles expresses salience.
 This thesis is opposed to others emphasizing the aspects of\emph{uniqueness}
 (as in the tradition of \cite{denoting})
 or specifity instead of salience.
 The thesis is supported by the following two examples:
 \noitem\begin{quote}\mbox{}\hfill\emph
 {(definite, salient, specific, but not unique)}\\
 ``\underline{The dog} got in a fight with another dog.'' 
 %(\cite{lewis}, \p 178; after 
 \getittotheright{\cite[\p\,20; our underlining]{heusingerepsilon}}
 \noitem\end{quote}\notop\begin{quote}\mbox{}\hfill
 \emph
 {(indefinite, not salient, but specific)}\\{\fraknomath``\germantexttwo''}
 \getittotheright{\cite[\p 16]{heusingerepsilon}}
 \noitem\end{quote}\notop\begin
 {quote}``\englishtexttwo\underline''\getittotheright
 {(our translation, our underlining)}
 \noitem\end{quote}\noindent
 Nevertheless, indefiniteness is typically unspecific: 
 \notop\begin{quote}\mbox{}\hfill
 \emph{(indefinite, not salient, unspecific)}\\
 {\fraknomath``\germantextten''}\getittotheright
 {\cite[\p 16]{heusingerepsilon}}
 \noitem\end{quote}\notop\begin
 {quote}``\englishtextten\underline''\getittotheright
 {(our translation, our underlining)}
 \noitem\end{quote}\noindent
 \heusinger's thesis is not consistent, however, with the following example:
 \mannname\ \mannlifetime\
 starts his narration ``{\fraknomath\germantextseven}'' 
 as follows:
 \noitem\begin{quote}\mbox{}\hfill\emph
 {(definite, specific, but not salient)}
 \\{\fraknomath``\germantextsix''}\getittotheright{\cite{friedemann}}
 \noitem\end{quote}\notop\begin
 {quote}``\englishtextsix\underline''\getittotheright
 {(our translation, our underlining)}
 \noitem\end{quote}\noindent
Obviously, none of uniqueness, salience, or specifity alone determines
definiteness of articles: For uniqueness this becomes obvious from the
first example already. 
For salience and specifity the following table may be helpful:
\par\noindent\getittotheright{\begin{tabular}[t]{l||l|l}
 &\sc Salient
 &\sc Not Salient
\\\hline\hline\sc Specific
 &\footroom\headroom\math{\{\mbox{\underline{The dog}, \ldots}\}}
 &\math{\{\mbox{\underline{\em a book}, \ldots}\}\nottight{\nottight\uplus}
        \{\mbox{\underline{The nurse}, \ldots}\}}
\\\hline\sc Unspecific
 &\headroom\math\emptyset
 &\math{\{\mbox{\underline{{\em a}\/(n arbitrary)\emph{book}}, \ldots}\}}
\\\end{tabular}}\par\noindent
 If---as I conjecture---examples for ``indefinite, but salient'' 
%which are not merely\emph{quantificational}
%(\ie\ generic or qualitative, \cf\ \sectref{section proviso})
 do not exist, 
 salience indeed requires
 definite forms; but not vice versa.
 In the technical treatment of salience
 with \heusinger's indexed \math\varepsilon-operator
 in \cite{heusingerepsilon}, however, salience and definite forms indeed
 require each other.} \ 
\Cfnlb\ \cite{heusingerepsilon}
for further reference on the \nlbmath\varepsilon\
in the semantics of natural language. \ 
\par\halftop\noindent
The possible advantage of our semantics for the \math\varepsilon\ 
is that it is not right-unique but admits commitment to choices.
Thus, it may help brave \ulysses\ to avoid both threats.
\vfill\pagebreak
%%%%%%%%%%%%%%%%%%%%%%%%%%%%%%%%%%%%%%%%%%%%%%%%%%%%%%%%%%%%%%%%%%%%%%%%%%%%%%%
\subsection{Problematic Aspects of Our \math\varepsilon}\label
{section critical cases}
In this \sectref{section critical cases}, we discuss some aspects
whose modeling in our free variable framework with our \nlbmath\varepsilon\
may fail when we take the straightforward way.
The reason for this partial failure is that the posed representational demands
are in conflict with 
our requirement of \wellfoundedness\ or acyclicity on the \vc\ \nlbmath R of our 
\nlbmath R-\cc s, \cfnlb\ \sectrefs{subsection Rules}{section substitutions},
 \defiref{definition choice condition}, and 
 \examref{example choice-condition}. \
These representational demands are\emph
{\henkin\ quantification} (\sectref{section quantifiers}) and\emph
{cyclic choice} in \bachpeterssentence s (\sectref{section cyclic choices}). \ 
We \nolinebreak also show how to overcome these two weaknesses in our framework 
by simple deviations, namely by\emph{raising} and by\emph{parallel choice}.
%%%%%%%%%%%%%%%%%%%%%%%%%%%%%%%%%%%%%%%%%%%%%%%%%%%%%%%%%%%%%%%%%%%%%%%%%%%%%%%
\subsubsection{\henkin\ Quantification}\label
{section quantifiers}
In \cite{hintikkaquantification}, quantifiers in \firstorder\ logic were
found insufficient to give the precise semantics of some English sentences. \ 
In \cite{hintikkaprinciples}, \emph{IF logic}, \ 
\ie\ \underline Independence-\underline Friendly logic---a \firstorder\ logic 
with more flexible quantifiers---is presented to overcome this weakness. \ 
% We, however, will suggest to get rid of the quantifiers at all and 
% restrict the dependences to a minimum. \ 
In \cite{hintikkaquantification}, we find the following sentence:
\begin{quote}
Some relative of each villager and some relative of each townsman 
\\hate each other.\hfill (H0)\hspace*{-\rightmargin}
\end{quote}
Let us first change to a lovelier subject:
\begin{quote}
Some loved one of each woman and some loved one of each man 
\\love each other.\hfill (H1)\hspace*{-\rightmargin}
\end{quote}
For our purposes here,
we consider (H1) to be equivalent to the following sentence, which may
be easier to understand and more meaningful:
\noitem\begin{quote}Every woman would love someone and every man would love 
someone,\\such that these loved ones would love each other.\noitem\end{quote}
(H1) can be represented by the following
 \henkin-quantified IF-logic formula:
\newcommand\innerpartofhenkinquantifiedformula[5]
{\inparenthesesoplist{\Femaleppp{#1}\oplistund\Maleppp{#2}}
 \nottight{\nottight\implies}#5
 \inparenthesesoplist{\Lovespp{#1}{#3}\oplistund\Lovespp{#2}{#4}
   \oplistund\Lovespp{#3}{#4}\oplistund\Lovespp{#4}{#3}}}
\newcommand\fullhenkinquantifiedformula[6]{#6\inparentheses
{\!\!\innerpartofhenkinquantifiedformula{#1}{#2}{#3}{#4}{#5} \!\!}}
\par\noindent\LINEmaths{\fullhenkinquantifiedformula
{\boundvari x 0}
{\boundvari y 0}
{\boundvari y 1}
{\boundvari x 1}
{ \exists\boundvari y 1/\boundvari y 0\stopq
 \exists\boundvari x 1/\boundvari x 0\stopq}
{\forall\boundvari x 0\stopq\forall\boundvari y 0\stopq}}{}(H2)
\par\noindent
Note that Formula~(H2) is already close to anti-prenex form; \hskip.3em
so we cannot reduce the dependences of its quantifiers by moving 
them closer toward the leaves of the formula tree.

Let us refer to the standard game-theoretic semantics for quantifiers
(\cf\ \eg\ \cite{hintikkaprinciples}), \hskip.2em
which is defined as follows: \hskip.3em
Witnesses have to be picked for the quantified variables outside-in. \hskip.3em
We have to pick the witnesses for the \math\gamma-quantifiers
(\ie, in \nolinebreak(H2), for the existential quantifiers), \hskip.2em
and our opponent in the game picks the witnesses for the 
\mbox{\math\delta-quantifiers}
(\ie\ \nolinebreak for the universal quantifiers in \nolinebreak(H2)). \hskip.3em
We win iff the resulting quantifier-free 
formula evaluates to true. \hskip.3em
A formula is true iff we have a winning strategy.

Then a \henkin\ quantifier such as \hskip.2em
``\maths{\exists\boundvari y 1/\boundvari y 0}{}.'' \hskip.2em
in (H2) \hskip.1em
is a special quantifier, \hskip.1em
which is a bit different from 
``\maths{\exists\boundvari y 1}{}.''\@. \ \ 
Game-theoretically, \hskip.1em
it 
% the \henkin\ quantifier ``\maths{\exists\boundvari x 1/\boundvari x 0}{}.''\
has the following semantics: \
It asks us to pick the loved one \nlbmath{\boundvari y 1} independently from
the choice of the man \nlbmath{\boundvari y 0} \hskip.1em
(by our opponent in the game), \hskip.2em
although the \henkin\ quantifier occurs in the scope of the quantifier 
``\nlbmath{\forall\boundvari y 0.}''.

An alternative way to define the semantics of \henkin\ quantifiers 
is by describing their effect on the logically equivalent 
{\em raised} \hskip.15em
forms of the formulas in which they occur. \hskip.3em
{\em Raising} is a dual of \skolemization, \cf\ \cite{miller}. \hskip.35em
The raised version is defined as usual, beside that 
a \math\gamma-quantifier, \hskip.2em
say \hskip.1em
``\maths{\exists\boundvari y 1}.'', \hskip.35em
followed by
a slash as in \hskip.2em
``\maths{\exists\boundvari y 1/\boundvari y 0}.'', \hskip.35em
are raised in such a form that \boundvari y 0 \nolinebreak does not appear 
as an argument to the raising function for \boundvari y 1.%

According to this, 
{\it\mutatismutandis}, \hskip.3em
(H2) is logically equivalent to its following raised form
\nolinebreak (H3), \hskip.3em
where \boundvari y 0 does not occur as an argument to the raising function
\nlbmath{\app{\boundvari y 1}{\boundvari x 0}}, \hskip.2em
which, \hskip.1em
however, \hskip.1em
would be the case if we had a usual
\math\gamma-quantifier \hskip.2em
``\maths{\exists\boundvari y 1}.'' \hskip.2em
instead of \hskip.2em
``\maths{\exists\boundvari y 1/\boundvari y 0}.'' \hskip.2em
in \nolinebreak (H2).
\par\noindent\LINEmaths{
\fullhenkinquantifiedformula
{\boundvari x 0}
{\boundvari y 0}
{\app{\boundvari y 1}{\boundvari x 0}}
{\app{\boundvari x 1}{\boundvari y 0}}
{}
{\exists\boundvari y 1\stopq\exists\boundvari x 1\stopq
 \forall\boundvari x 0\stopq\forall\boundvari y 0\stopq}}{}(H3)
\par\noindent Now, (H3) looks already very much like the following 
tentative representation of (H1) in our framework of free variables:
\par\notop\noindent\LINEmaths{\mbox{}~~~~~~~~~~~~~~
     \inparenthesesoplist{
         \Femaleppp{\wforallvari x 0}
       \oplistund
         \Maleppp{\wforallvari y 0}} 
     \implies
     \inparenthesesoplist{
         \Lovespp{\wforallvari x 0}{\sforallvari y 1}
       \oplistund
         \Lovespp{\wforallvari y 0}{\sforallvari x 1}
       \oplistund
         \Lovespp{\sforallvari y 1}
                 {\sforallvari x 1}
       \oplistund
         \Lovespp{\sforallvari x 1}
                 {\sforallvari y 1}}
}{}(H1\math')\par\noindent
with \cc\ \nlbmath C given by 
\LINEmaths{\begin{array}[t]{l l l}
  \app C{\sforallvari y 1}
 &:=
 &\Femaleppp{\wforallvari x 0}
  \implies\Lovespp{\wforallvari x 0}{\sforallvari y 1}
\\\app C{\sforallvari x 1}
 &:=
 &\Maleppp{\wforallvari y 0}
  \implies\Lovespp{\wforallvari y 0}{\sforallvari x 1}
\\\end{array}}{}\par\noindent
which requires the \vc\ to contain
\bigmaths{R_C:=\{\pair{\wforallvari x 0}{\sforallvari y 1},
                 \pair{\wforallvari y 0}{\sforallvari x 1}\}}{} by \defiref
{definition choice condition}(\ref{item one definition choice condition}).
Note that we can add \bigmaths{\pair{\sforallvari y 1}{\wforallvari y 0}}{}
to our \vc\ \nlbmath R 
here to express that \sforallvari y 1 must not read \nlbmath{\wforallvari y 0},
which results in a logical equivalence to the original formula 
\nolinebreak (H2) but with a standard \math\gamma-quantification
``\nlbmaths{\exists\boundvari x 1.}{}'' 
instead of the \henkin\ quantification
``\maths{\exists\boundvari x 1/\boundvari x 0.}{}''.

If we tried to model the \henkin\
quantifier by adding \bigmaths{\pair{\sforallvari x 1}{\wforallvari x 0}}{}
to \nlbmath R in addition, 
our \cc\ \nlbmath C would not be an \math R-\cc\ anymore 
by \defiref{definition choice condition} due to the following cycle:\\
\LINEmath
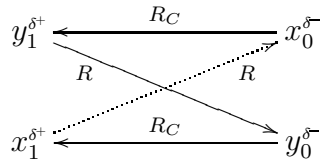
{\xymatrix{
   {\sforallvari y 1}
   \ar[drrr]_<<<<<<{R}
 &&&{\wforallvari x 0}
   \ar[lll]_{R_C}
 \\{\sforallvari x 1}
   \ar@{.>}[urrr]_>>>>>>{R}
 &&&{\wforallvari y 0}
   \ar[lll]_{R_C}
 }} \ \ 
%\begin{minipage}[t]{19em}
\par\noindent As shown in \litexamref{2.9} of \cite{wirthcardinal}, 
the \deltaplus-rules from \sectref{section where delta rules are}
become unsound when we admit such cycles. \ 
Without the \deltaplus-rules we could argue
that \nlbmath{R_C} means something
%\linebreak\vspace*{-1.6ex}\end{minipage}
like ``is \nolinebreak read by'' and 
that \nlbmath R means something like ``must not read\closequotecomma
so that it would be sufficient to require only
the given irreflexivity of \nlbmaths{R_C\tight\circ R},
instead of the irreflexivity of the transitive closure of 
\nlbmaths{R_C\tightcup R},
which is nothing but the acyclicity of \nlbmaths{R_C\tightcup R}. \ \ 
Such ``weak forms'' 
are indeed sound for \deltaminus-rules
(\cfnlb\ \cite[\litnoteref 9]{wirthcardinal}), \ 
but the 
price of abandoning the \mbox{\deltaplus-rules} (\esp\ in a framework
for \hilbert's \math\varepsilon) \ 
is ridiculously high in comparison to an increased order of some variables, \ 
such as of \nlbmath{\boundvari x 1} and
\nlbmath{\boundvari y 1} in \nolinebreak (H3).

Let us compare the failure of our approach to represent
\henkin\ quantifiers without raising on the one hand,
with the situation in 
% \cite[\p\,85]{heusingerindefinitesold},
% where (H0) is given the label\,(25), 
\cite[\p\,85]{heusingerindefinitesold}
(where (H0) has the label\,(25))
on the other hand. \ 
It may be interesting to see that it is well possible to model \henkin\
quantifiers with a right-unique version of \hilbert's \nlbmath\varepsilon, \ 
\cf\ \cite{heusingerindefinitesold}, \p\,85, \nolinebreak(25c). \ 
After replacing both ``hating'' and ``being a relative'' with 
``loving\closequotecomma 
adding the fact that the loved ones
are not chosen from empty sets of candidates
(\ie\ the presupposition that they exist), using \wfuv s for the 
outermost universal bound variables, 
correcting a flaw\commanospace\footnote
{Note that our modeling of (H1) as (H3) of \sectref{section quantifiers}
 is correct, whereas 
 the modeling of (H0) as (25c) of \cite{heusingerindefinitesold}
 is flawed: {\em Mutatis mutandis}, 
 both ``hating'' and ``being a relative'' replaced with 
 ``loving\closequotecomma 
 already the less complex (25b) of \cite{heusingerindefinitesold}
 has this flaw and \nolinebreak reads:
 \par\noindent\LINEmaths{
    \exists\boundvari x 1,\boundvari y 1\stopq
    \forall\boundvari x 0,\boundvari y 0\stopq
    \inparentheses{
      \inparenthesesoplist{
          \Femaleppp{\boundvari x 0}
        \oplistund
          \Maleppp{\boundvari y 0}
        \oplistund
          \Lovespp{\boundvari x 0}{\app{\boundvari y 1}{\boundvari x 0}}
        \oplistund
          \Lovespp{\boundvari y 0}{\app{\boundvari x 1}{\boundvari y 0}}} 
      \implies
      \inparenthesesoplist{
          \Lovespp{\app{\boundvari y 1}{\boundvari x 0}}
                  {\app{\boundvari x 1}{\boundvari y 0}}
        \oplistund
          \Lovespp{\app{\boundvari x 1}{\boundvari y 0}}
                  {\app{\boundvari y 1}{\boundvari x 0}}}
 \!\!}}{}(25b\math')\par\noindent
 Indeed, it is easy to see from \nolinebreak (H2) that the polarity of the
 first two (negative) occurrences of the \math\Lovessymbol-predicate 
 in (25b\math') must actually be positive.}
and enhancing readability by introducing two more 
\wfuv s \wforallvari x 1 and \wforallvari y 1, \ \ 
\mbox{(25c) of \cite{heusingerindefinitesold}} 
reads:
\par\noindent\LINEmaths{
     \inparenthesesoplist{
         \Femaleppp{\wforallvari x 0}
       \oplistund
         \Maleppp{\wforallvari y 0}
       \oplistund
         \wforallvari x 1=
         \varepsilon\boundvari x 1.\,\Lovespp{\wforallvari y 0}{\boundvari x 1}
       \oplistund
         \wforallvari y 1=
         \varepsilon\boundvari y 1.\,\Lovespp{\wforallvari x 0}{\boundvari y 1}}
     \implies
     \inparenthesesoplist{
         \Lovespp{\wforallvari x 0}{\wforallvari y 1}
       \oplistund
         \Lovespp{\wforallvari y 0}{\wforallvari x 1}
       \oplistund
         \Lovespp{\wforallvari y 1}{\wforallvari x 1}
       \oplistund
         \Lovespp{\wforallvari x 1}{\wforallvari y 1}}
}{}(H5)\par\noindent 
To model the \henkin\ quantifier correctly,
an \nlbmath\varepsilon-term such as \ ``\maths{
  \varepsilon\boundvari x 1.\,\Lovespp{\wforallvari y 0}{\boundvari x 1}
}{}'' \  in \nolinebreak (H5) must not depend on \nlbmath{\wforallvari x 0}. \ 
This is contrary to 
\sforallvari x 1 in (H1\math'), \ 
whose value may well depend on that of \nlbmath{\wforallvari x 0}, \ 
unless \bigmaths{\pair{\sforallvari x 1}{\wforallvari x 0}}{}
is included in \nlbmath{\transclosureinline R}\@. \ 
To \nolinebreak achieve this independence, 
it is not necessary that the \math\varepsilon\ gets an 
extensional semantics. \ 
%, \cf\ \sectref{section E2}. \ 
It \nolinebreak
suffices that the semantics of the \mbox{\math\varepsilon-term}
does not depend on anything not named in its formula, namely
\ ``\math{\Lovespp{\wforallvari y 0}{\boundvari x 1}}'' \ in our case. \ 
On the one hand, 
any of the semantics of \sectrefsectionrightuniquesemantics\
satisfies this independence,
but---due to its right-uniqueness---is not suitable 
for describing the semantics of determiners in natural languages,
\cfnlb\ \sectref{section linguistic literature}, 
\lititemref{``Right-Uniqueness''}. \ 
On \nolinebreak the other hand,
\heusinger's indexed \math\varepsilon-operator, however, does not necessarily
satisfy this independence, 
because it may get information on \nlbmath{\wforallvari x 0}
out of its situational index,
\cfnlb\ \sectref{section linguistic literature}, 
below \lititemref{``Uncommitted Choice''}, and 
\nolinebreak\sectref{section geurts}.

Thus, 
the inability of our framework to capture \henkin\ quantifiers without raising
is also implicitly present in all other known approaches suitable 
for describing the semantics of determiners in natural languages.

Moreover, raising cannot be avoided in the presence of 
explicit \mbox{\math\varepsilon-terms} because these terms are an 
equivalent to raising already.

Furthermore, in natural language, \henkin\ quantification is typically ambiguous
and the \henkin-quantified versions are always logically stronger than
the ones with usual \mbox{\math\gamma-quantifiers} instead. \ 
Thus, it \nolinebreak appears to be advantageous to have more flexibility 
in computing the semantics of sentences in natural language 
by starting with possibly weaker formulations such as \nolinebreak(H1\math').
While we cannot represent the \henkin\ quantification in our framework without
raising, we could start with the following raised version of 
\nolinebreak(H1\math').\par\noindent\LINEmaths{
     \inparenthesesoplist{
         \Femaleppp{\wforallvari x 0}
       \oplistund
         \Maleppp{\wforallvari y 0}
       \oplistund
         \wforallvari x 1=
         \app{\app{\sforallvari x 3}{\wforallvari x 0}}{\wforallvari y 0}
       \oplistund
         \wforallvari y 1=
         \app{\sforallvari y 2}{\wforallvari x 0}}
     \implies
     \inparenthesesoplist{
         \Lovespp{\wforallvari x 0}{\wforallvari y 1}
       \oplistund
         \Lovespp{\wforallvari y 0}{\wforallvari x 1}
       \oplistund
         \Lovespp{\wforallvari y 1}{\wforallvari x 1}
       \oplistund
         \Lovespp{\wforallvari x 1}{\wforallvari y 1}}
}{}(H2\math')\par\noindent
with \math R-\cc\ \nlbmath C given by 
\par\noindent\LINEmaths{\begin{array}{l l l}
  \app C{\sforallvari y 2}
 &:=
 &\lambda\boundvari x{}\stopq\inpit{
  \Femaleppp{\boundvari x{}}
  \nottight\implies\Lovespp{\boundvari x{}}
  {\app{\sforallvari y 2}{\boundvari x{}}}}
\\\app C{\sforallvari x 2}
 &:=
 &\lambda\boundvari y{}\stopq\inpit{
  \Maleppp{\boundvari y{}}\nottight\implies\Lovespp{\boundvari y{}}
  {\app{\sforallvari x 2}{\boundvari y{}}}}
\\\app C{\sforallvari x 3}
 &:=
 &\lambda\boundvari x{}\stopq\lambda\boundvari y{}\stopq\inpit{
  \Femaleppp{\boundvari x{}}\und\Maleppp{\boundvari y{}}
  \nottight\implies\Lovespp{\boundvari y{}}
  {\app{\app{\sforallvari x 3}{\boundvari x{}}}{\boundvari y{}}}}
\\\end{array}}{}\par\noindent
which requires no extension of the \vc\ \nlbmath R\@. \ 
When we then find out that the sentence is actually meant to be 
\henkin\ quantified, we can apply the substitution \nolinebreak\mbox{\math
{\sigma:=\{\sforallvari x 3\mapsto\lambda\boundvari u{}.\,
 %\lambda\boundvari y{}\stopq\inpit{\app{
 \sforallvari x 2
 %}{\boundvari y{}}}
 \}}}. \ 
This turns (H2\math') into a form equivalent to (H3),
reflecting the intended semantics of \nolinebreak(H1). \ 
Note that the condition \nlbmath{\inpit{\app{Q_C}{\sforallvari x 3}}\sigma}
(\cf\ \defiref{definition Q}),
which is required for invariance of reduction under instantiation in
\theoref{theorem strong reduces to}(6),
is
\par\noindent\LINEmaths{
\forall\boundvari x{}\stopq\forall\boundvari y{}\stopq\inparenthesesoplist{
  \exists\boundvari z{}\stopq\inpit{
  \Femaleppp{\boundvari x{}}\und\Maleppp{\boundvari y{}}
  \nottight\implies\Lovespp{\boundvari y{}}{\boundvari z{}}}
  \oplistimplies
  \inpit{
  \Femaleppp{\boundvari x{}}\und\Maleppp{\boundvari y{}}
  \nottight\implies\Lovespp{\boundvari y{}}
  {\app{\app{\inpit{\lambda\boundvari u{}.\,\sforallvari x 2}}
            {\boundvari x{}}}
       {\boundvari y{}}}}}}{}\par\noindent and simplifies to
\\\LINEmaths{\forall\boundvari y{}\stopq\inparenthesesoplist{
  \exists\boundvari z{}\stopq\inpit{
  \Maleppp{\boundvari y{}}
  \nottight\implies\Lovespp{\boundvari y{}}{\boundvari z{}}}
  \oplistimplies
  \inpit{
  \Maleppp{\boundvari y{}}
  \nottight\implies\Lovespp{\boundvari y{}}
  {\app{\sforallvari x 2}{\boundvari y{}}}}}},\par\noindent 
which is just \app{Q_C}{\sforallvari x 2}, which is valid according to 
\lemmref{lemma Q valid}.

A solution without raising will be found in \cite{SR--2011--01}.

%%%%%%%%%%%%%%%%%%%%%%%%%%%%%%%%%%%%%%%%%%%%%%%%%%%%%%%%%%%%%%%%%%%%%%%%%%%%%%
\subsubsection{Cyclic Choices and \bachpetersSentence s}\label
{section cyclic choices}
As an example where references of an anaphor and a cataphor cross
(\ie\ a so-called ``\bachpeterssentence'' after \bachname\ and \petersname),
consider
\begin{quote}A man who loves her marries a woman who, 
however, does not love him.
\hfill(B0)\hspace*{-\rightmargin}\par
\end{quote}
If we start with  
\\\noindent\LINEmath{\Marriespp{\sforallvari y 0}{\sforallvari x 0}
}(B1)\par\halftop\noindent with \math R-\cc\hfill\math{
\begin{array}[t]{@{}l@{~~~}l@{~~~}l@{}}
  \app C{\sforallvari y 0}
 &:=
 &\Maleppp{\sforallvari y 0}
  \und\Lovespp{\sforallvari y 0}{\app{\existsvari f 1}{\sforallvari y 0}}
  \und\Femaleppp{\app{\existsvari f 1}{\sforallvari y 0}},
\\\app C{\sforallvari x 0}
 &:=
 &\Femaleppp{\sforallvari x 0}
  \und\neg\Lovespp{\sforallvari x 0}{\app{\existsvari f 2}{\sforallvari x 0}}
  \und\Maleppp{\app{\existsvari f 2}{\sforallvari x 0}},
\\\end{array}}\par\noindent
then \nlbmath{\transclosureinline R} has to contain \bigmaths{\{
  \pair{\existsvari f 1}{\sforallvari y 0}
  \comma
  \pair{\existsvari f 2}{\sforallvari x 0}
\}}{} according to \defiref
{definition choice condition}(\ref{item one definition choice condition}).
This says that the substitution \nlbmath{\sigma:=}
\bigmaths{\{
\existsvari f 1\mapsto\lambda z.\,\sforallvari x 0
\comma
\existsvari f 2\mapsto\lambda z.\,\sforallvari y 0
\}},
which binds the pronouns ``her'' (\app{\existsvari f 1}{\sforallvari y 0})
and ``him'' ({\app{\existsvari f 2}{\sforallvari x 0}}) 
to their intended referents \sforallvari x 0 and \sforallvari y 0, \resp,
is not an \math R-substitution, however. \ 
This is due to the following cycle; \ 
\cfnlb\ \defiref{definition ex r sub}:
\par\noindent\LINEmath{\xymatrix{
   {\sforallvari y 0}
   \ar[drrr]_<<<<<<{R_\sigma}
 &&&{\existsvari f 1}
   \ar[lll]_{R}
 \\{\sforallvari x 0}
   \ar[urrr]_>>>>>>{R_\sigma}
 &&&{\existsvari f 2}
   \ar[lll]_{R}
 }}\par\noindent
Indeed, the (extended) \math\sigma-updated (and \math{\lambda\beta}-reduced)
\cc\ \nlbmath{\pair{C'}{R'}} 
of \nlbmath{\pair C R}
(\cfnlb\ \defiref{definition ex str s up}), namely \hfill\LINEmath
{\begin{array}[t]{@{}l@{~~~~}l@{~~~~}l@{}}
  \app{C'}{\sforallvari y 0}
 &:=
 &\Maleppp{\sforallvari y 0}
  \und\Lovespp{\sforallvari y 0}{\sforallvari x 0}
  \und\Femaleppp{\sforallvari x 0},
\\\app{C'}{\sforallvari x 0}
 &:=
 &\Femaleppp{\sforallvari x 0}
  \und\neg\Lovespp{\sforallvari x 0}{\sforallvari y 0}
  \und\Maleppp{\sforallvari x 0},
\\\end{array}}\\\noindent
cannot be an \math{R'}-\cc\ for any (acyclic) \vc\ \nlbmath{R'},
\cf\ \defiref{definition choice condition}.

\yestop\noindent
As we cannot choose \sforallvari y 0 before \sforallvari x 0
nor  \sforallvari x 0 before \sforallvari y 0, we have to choose them 
in parallel.
Thus, the only way to overcome this failure within our framework
seems to be to start with
\par\halftop\noindent\LINEmath{
  \Marriespp{\app{\nth 1}{\sforallvari z{}}}
            {\app{\nth 2}{\sforallvari z{}}}
}(B2)\\with \cc\\\noindent\mbox{}\hfill\math{\begin
{array}{@{}l@{~~~~~~}l@{~~~~~~}l@{}}
  \app{C_1}{\sforallvari z{}}
 &:=
 &\inparenthesesoplist{
      \Maleppp{\app{\nth 1}{\sforallvari z{}}}
    \und\Lovespp
      {\app{\nth 1}{\sforallvari z{}}}
      {\app{\existsvari f 1}{\sforallvari z{}}}
    \und\Femaleppp{\app{\existsvari f 1}{\sforallvari z{}}}
    \oplistund
      \Femaleppp{\app{\nth 2}{\sforallvari z{}}}
    \und\neg\Lovespp
      {\app{\nth 2}{\sforallvari z{}}}
      {\app{\existsvari f 2}{\sforallvari z{}}}
    \und\Maleppp{\app{\existsvari f 2}{\sforallvari z{}}}
  }
\\\end{array}}\par\noindent
where \sforallvari z{} has the type of a pair and
\math{\nth 1} and \math{\nth 2} 
are its projections to the \nth 1 and \nth 2 component,
respectively.
This requires the \vc\ \nlbmath R to contain
\bigmaths{\{
  \pair{\existsvari f 1}{\sforallvari z{}}
  \comma
  \pair{\existsvari f 2}{\sforallvari z{}}
\}}, which admits the substitution \nlbmath{\sigma':=}
\bigmaths{\{
\existsvari f 1\tight\mapsto{\nth 2}
\comma
\existsvari f 2\tight\mapsto{\nth 1}
\}}{}
to be an \math R-substitution.
Now, (B2) together with 
the (extended) \math{\sigma'}-update of \nlbmath{C_1}
(\cf\ \defiref{definition ex str s up})
captures the intended semantics of (B0) correctly. \ 
\par\yestop\noindent
Finally, note that a \cc\ of\par\noindent\mbox{}\hfill\math{
\begin{array}{@{}l@{~~~}l@{~~~}l@{}}
  \app{C_2}{\sforallvari z{}}
 &:=
 &\inparenthesesoplist{
      \Maleppp{\app{\nth 1}{\sforallvari z{}}}
    \und\Lovespp
         {\app{\nth 1}{\sforallvari z{}}}
         {\app{\existsvari f 1}{\app{\nth 1}{\sforallvari z{}}}}
    \und\Femaleppp{\app{\existsvari f 1}{\app{\nth 1}{\sforallvari z{}}}}
    \oplistund
      \Femaleppp{\app{\nth 2}{\sforallvari z{}}}
    \und\neg\Lovespp
         {\app{\nth 2}{\sforallvari z{}}}
         {\app{\existsvari f 2}{\app{\nth 2}{\sforallvari z{}}}}
    \und\Maleppp{\app{\existsvari f 2}{\app{\nth 2}{\sforallvari z{}}}}}
\\\end{array}}\par\noindent
requires the substitution \bigmaths{\{
\existsvari f 1\mapsto\lambda u.\,\inpit{\app{\nth 2}{\sforallvari z{}}}
\comma
\existsvari f 2\mapsto\lambda u.\,\inpit{\app{\nth 1}{\sforallvari z{}}}
\}},
which is still no \mbox\Rsub\ because of the cycles between
\sforallvari z{} and \existsvari f i. \ \ 
This means that---within cyclic choices---we should not
restrict or project before all ambiguities have been resolved.

%%%%%%%%%%%%%%%%%%%%%%%%%%%%%%%%%%%%%%%%%%%%%%%%%%%%%%%%%%%%%%%%%%%%%%%%%%%%%%%%
\subsubsection{Conclusion}
We have managed to overcome the two weaknesses of our framework 
exhibited in \sectrefs{section quantifiers}{section cyclic choices} 
by simple deviations. 
For the \henkin\ quantifiers we had to increase the 
order of variables by\emph{raising}. 
For the \bachpeterssentence s
we had to replace a cycle of choices with a single\emph{parallel choice}.
As these problems are somehow unavoidable without paying high prices,
this appears to be acceptable, especially because
the partially ordered quantification required for 
natural languages in \cite{hintikkaquantification} is available for free 
in our framework of free variables of \sectrefs
{section free}{section substitutions}.

Indeed, these inelegant aspects 
of our framework should not lead us to the conclusion
to open \pandora's box by admitting cyclic choices.
This would let most of the famous antinomies break into our system. 
If we admitted cyclic choices, we could not even say anymore 
whether a \cc\ can be satisfied for a certain \sfuv\ or not. \ 
\examref{example choice-condition} in \sectref{section choice-conditions}
makes the essential problem obvious.
\vfill\pagebreak
%%%%%%%%%%%%%%%%%%%%%%%%%%%%%%%%%%%%%%%%%%%%%%%%%%%%%%%%%%%%%%%%%%%%%%%%%%%%%%%%
%%%%%%%%%%%%%%%%%%%%%%%%%%%%%%%%%%%%%%%%%%%%%%%%%%%%%%%%%%%%%%%%%%%%%%%%%%%%%%%
\subsection{More Problems with a Right-Unique \math\varepsilon}\label
{section geurts}In \cite{geurts-one}, the use of \hilbert's \math\varepsilon\ 
in form of choice functions for the semantics of indefinites 
is attacked in several ways; and it is proposed that 
there is no way to interpret indefinites\emph{in situ},
but that some form of ``movement'' is necessary,
which, roughly speaking, may be interpreted as changing scopes of quantifiers.
Although the examples given in \cite{geurts-one} are perfectly convincing
in the given setting, 
we would like to point out that all the presented problems with the \nlbmath
\varepsilon\ disappear when one uses a non-right-unique version such as ours. 
The following three example sentences and their labels are the ones 
of \cite{geurts-one}.
%%%%%%%%%%%%%%%%%%%%%%%%%%%%%%%%%%%%%%%%%%%%%%%%%%%%%%%%%%%%%%%%%%%%%%%%%%%%%%%%
\subsubsection{All bicycles were stolen by a German.\hfill\rm(1a)}
We model this as
\\\noindent\LINEmath{
  \Bicycleppp{\wforallvari x{}}\nottight{\nottight\implies}
  \StolenBypp{\wforallvari x{}}{\sforallvari y{}}
}\\\noindent with \cc\\\noindent\LINEmath{
\app C{\sforallvari y{}}
\nottight{\nottight{\nottight{:=}}}\Germanppp{\sforallvari y{}}}\par\noindent
If---in a first step---we find a model for this sentence with an empty \vc, 
then---in a second step---we can check whether it also satisfies a \vc\ 
that contains
\nlbmath{\pair{\sforallvari y{}}{\wforallvari x{}}} in addition. \ 
A \nolinebreak success 
of the first step provides us with a model for the weaker reading;
a success of the second step with one for the stronger reading, too; \ 
\ie\ that all bicycles were stolen by the same German. \ 
And this without ``moving'' any quantifiers or the like; \ 
which is, however, required when changing from \par\noindent\LINEmaths{
\forall\boundvari x{}\stopq\inparentheses{\Bicycleppp{\boundvari x{}}
\nottight{\nottight\implies}
\exists\boundvari y{}\stopq\inpit{
\Germanppp{\boundvari y{}}
\nottight\und
\StolenBypp{\boundvari x{}}{\boundvari y{}}
}}}{}~~(1a-weak)\\to\\\LINEmaths{\exists\boundvari y{}\stopq\inparentheses{
\Germanppp{\boundvari y{}}\nottight\und
\forall\boundvari x{}\stopq\inpit{\Bicycleppp{\boundvari x{}}
\nottight{\nottight\implies}
\StolenBypp{\boundvari x{}}{\boundvari y{}}}}}{}(1a-strong)%
\par\yestop\noindent
For a more interesting problem with right-unique \math\varepsilon,
let us consider the following example.\notop

\subsubsection{Every girl gave a flower to a boy she fancied.\hfill\rm(5)}
Ignoring past tense, we model this as
\par\noindent\LINEmath{
  \Girlppp{\wforallvari x{}}\nottight{\nottight\implies}
  \Givepp{\wforallvari x{}}{\sforallvari z{}}{\sforallvari y{}}
}\\\noindent
with \cc\\\noindent\LINEmath{~~~~~~~~~~~~~~~~~~~~~~~~~~~~~~~~~~~~~~~
\begin{array}{l@{~~~~~~}l@{~~~~~~}l}
  \app C{\sforallvari y{}}
 &:=
 &\Girlppp{\wforallvari x{}}\nottight{\nottight\implies}
  \Boyppp{\sforallvari y{}}\und\Lovespp{\wforallvari x{}}{\sforallvari y{}}
\\\app C{\sforallvari z{}}
 &:=
 &\Flowerppp{\sforallvari z{}}
\\\end{array}}\par\noindent 
As a choice function must 
pick the identical element from an identical extension,
in \cite{geurts-one} 
there is a problem with two girls who love all boys,
but give their flowers to two different ones.
This problem does not
appear in our modeling because our semantical relation 
(\cf\ \nolinebreak\defiref{definition semantical relation})
does not depend on the common extension of their love,
but only has to contain \nlbmath{\pair{\wforallvari x{}}{\sforallvari y{}}},
which is in accordance with our \vc, which also has to contain
\nlbmath{\pair{\wforallvari x{}}{\sforallvari y{}}}
due to our above \cc\ for \nlbmath{\sforallvari y{}}, 
\cfnlb\ \defiref{definition choice condition}.

\yestop\noindent
The same problem of a common extension but a different choice object---but
now in all possible worlds and intensions---of the following example is 
again no problem for us.\pagebreak\vfill

\subsubsection
{Every odd number is followed by an even number\\that is not equal to it.
\hfill\rm(7)}
We model this as
\\\noindent\LINEmath{
  \Oddpp{\wforallvari x{}}\nottight{\nottight\implies}
  \tightplusppnoparentheses{\wforallvari x{}}\onepp\tightequal{\sforallvari y{}}
}\\\noindent with \cc\\\noindent\LINEmath{
~~~~~~~~~~~~~~~~~~~~~~~~~~~~~~~~~~~~~~~~~~~~
\begin{array}{l@{~~~~~~}l@{~~~~~~}l}
  \app C{\sforallvari y{}}
 &:=
 &\Oddpp{\wforallvari x{}}\nottight{\nottight\implies}
  \Evenpp{\sforallvari y{}}\und{\sforallvari y{}}\boldunequal{\wforallvari x{}}
\\\end{array}}\par\noindent All in all, 
there was no real reason to ``move''
quantifiers or the like and the arguments of 
\cite{geurts-one} are not justified 
in the absence of a right-unique behavior of the \nlbmath\varepsilon. \ 
Moreover, the moving of the quantifiers as from (1a-weak) to (1a-strong)
above is more complex and less intuitive than adding 
\nlbmath{\pair{\sforallvari y{}}{\wforallvari x{}}} to the current \vc.
\vfill\pagebreak

\subsection
{Donkey Sentences and \heusinger's Indexed \math\varepsilon-Operator}\label
{section donkey}
\subsubsection{If a man has a donkey, he beats it.\hfill\rm(D)}
The word ``syntax'' in the modern sense seems to have 
its first occurrence in the
voluminous writings of \chrysipposname\ \chrysipposlifetime, 
not the son of Pelops in the Oedipus mythos, 
but, of course, after \zenozweiname\ and \cleanthesname,
the third leader of the Stoic school.
So-called\emph{\chrysippossentence s} and\emph{donkey sentences}
demonstrate the difficulties of 
interaction of indefinite noun phrases in a conditional 
(``a \nolinebreak man\closequotecomma ``a \nolinebreak donkey'') 
and anaphoric pronouns 
referring to them in the conclusion (``he\closequotecomma ``it''). \ \
\Cfnlb\ \eg\ \cite[\litsectref{7}]{heusingerepsilon} for references
 on donkey and \chrysippossentence s. \ 
If semantics is represented with the help of quantification,
donkey sentences reveal difficulties resulting from quantifiers and 
their scopes. \ 
Let us have a closer look at two examples.\notop

\subsubsection{If a man loves a woman, she loves him.\hfill\rm(L0)}\label
{section If a man loves a woman}
If we start by modeling this tentatively as 
\par\noindent\LINEmath{\noparenthesesoplist{
    \exists\boundvari y 0\stopq
    \inparentheses{\Maleppp{\boundvari y 0}
        \und
       \exists\boundvari x 0\stopq\inparentheses{\Femaleppp{\boundvari x 0}
            \und\Lovespp{\boundvari y 0}{\boundvari x 0}}}
  \oplistimplies{
      \Femaleppp{\existsvari x 1}
    \und
      \Lovespp{\existsvari x 1}{\existsvari y 1}
    \und
      \Maleppp{\existsvari y 1}
}}}(L1)\par\noindent
we have no chance to resolve the reference of the pronouns ``she'' and
``him'' ({\existsvari x 1} and {\existsvari y 1}) 
before we get rid of the quantifiers.
If we apply \deltaminus-rules (\cf\ \sectref{subsection Rules})
(besides \math\alpha- and \math\beta-rules)
we end up with the three sequents
\par\noindent\LINEmath{\begin{array}[c]{l}
  \neg\Maleppp{\wforallvari y 0}\comma
\neg\Femaleppp{\wforallvari x 0}\comma
\neg\Lovespp{\wforallvari y 0}{\wforallvari x 0}\comma
\Femaleppp{\existsvari x 1}
\\\neg\Maleppp{\wforallvari y 0}\comma
\neg\Femaleppp{\wforallvari x 0}\comma
\neg\Lovespp{\wforallvari y 0}{\wforallvari x 0}\comma
\Lovespp{\existsvari x 1}{\existsvari y 1}
\\\neg\Maleppp{\wforallvari y 0}\comma
\neg\Femaleppp{\wforallvari x 0}\comma
\neg\Lovespp{\wforallvari y 0}{\wforallvari x 0}\comma
\Maleppp{\existsvari y 1}
\\\end{array}}(L2)\par\noindent
and a \vc\ \nlbmath R including \bigmaths{
\{\existsvari x 1,\existsvari y 1\}
\times
\{\wforallvari x 0,\wforallvari y 0\}
}, which says that the substitution 
\par\noindent\LINEmaths{\sigma^-\nottight{\nottight{:=}}
\{\existsvari x 1\tight\mapsto\wforallvari x 0\comma
\existsvari y 1\tight\mapsto\wforallvari y 0\}}{}\par\noindent
which turns the first and last sequents into tautologies and the middle one
\nolinebreak(L2) into the intended reading of \nolinebreak(L0),
is not an \math R-substitution and must not be applied, 
\cfnlb\ \defiref{definition ex r sub}. 
\\Using \deltaplus-rules instead of 
the \deltaminus-rules we get
\par\noindent\LINEmath{\begin{array}[c]{l}
  \neg\Maleppp{\sforallvari y 0}\comma
\neg\Femaleppp{\sforallvari x 0}\comma
\neg\Lovespp{\sforallvari y 0}{\sforallvari x 0}\comma
    \Femaleppp{\existsvari x 1}
\\\neg\Maleppp{\sforallvari y 0}\comma
\neg\Femaleppp{\sforallvari x 0}\comma
\neg\Lovespp{\sforallvari y 0}{\sforallvari x 0}\comma
      \Lovespp{\existsvari x 1}{\existsvari y 1}
\\\neg\Maleppp{\sforallvari y 0}\comma
\neg\Femaleppp{\sforallvari x 0}\comma
\neg\Lovespp{\sforallvari y 0}{\sforallvari x 0}\comma
      \Maleppp{\existsvari y 1}
\\\end{array}}\par\noindent
and a \vc\ \nlbmath R including 
\nlbmath{\{\pair{\sforallvari y 0}{\sforallvari x 0}\}}
instead. \ 
After application of the 
\\\math R-substitution \bigmaths{\sigma^+\nottight{\nottight{:=}}\{
\existsvari x 1\tight\mapsto\sforallvari x 0\comma
\existsvari y 1\tight\mapsto\sforallvari y 0\}}, \ 
the instance \mbox{of (L1) reduces to}\par\halftop\noindent\LINEmath{
\neg\Maleppp{\sforallvari y 0}\comma
\neg\Femaleppp{\sforallvari x 0}\comma
\neg\Lovespp{\sforallvari y 0}{\sforallvari x 0}\comma
\Lovespp{\sforallvari x 0}{\sforallvari y 0}
}(L3)\par\halftop\noindent
which is valid in a utopia where love is symmetric. \ 
A closer look reveals that our \math{\sigma^+}-updated \vc\ \nlbmath R 
now looks like
\bigmaths{\begin{array}[c]{@{}l@{}}\xymatrix{
   {\existsvari x 1}
 &{\sforallvari x 0}
   \ar[l]%_{R}
 &{\sforallvari y 0}
   \ar[r]%_{R}
   \ar[l]%_{R}
 &{\existsvari y 1}
}
\\\end{array}},
while our (\math{\sigma^+}-updated) \math R-\cc\ \nlbmath C is
\par\noindent\LINEmath{\begin{array}{l l l}
  \app C{\sforallvari y 0}
 &:=
 &\Maleppp{\sforallvari y 0}
\nottight{\nottight\und}
\exists\boundvari x 0\stopq\inparentheses{\Femaleppp{\boundvari x 0}
\und\Lovespp{\sforallvari y 0}{\boundvari x 0}}
\\\app C{\sforallvari x 0}
 &:=
 &\Femaleppp{\sforallvari x 0}
\und\Lovespp{\sforallvari y 0}{\sforallvari x 0}
\\\end{array}}\par\noindent
But even if (L3) may be valid, this 
is not what we wanted to say in \nolinebreak (L0),
where ``she'' and ``he'' are
obviously meant to be universal (strong, \math{\delta^-}).

\yestop\noindent
Thus, we had better start\emph{without quantifiers from the very beginning}, 
namely directly with 
\par\noindent\LINEmath{\noparenthesesoplist{
    {\Maleppp{\wforallvari y 0}
        \und\Lovespp{\wforallvari y 0}{\wforallvari x 0}
        \und\Femaleppp{\wforallvari x 0}
 }\oplistimplies{
      \Femaleppp{\existsvari x 1}
    \und\Lovespp{\existsvari x 1}{\existsvari y 1}
    \und\Maleppp{\existsvari y 1}
}}}(L4)\par\noindent
and empty \vc\ \nlbmath{R'}, and then apply the \math{R'}-substitution
\nlbmath{\sigma^-} from above to reduce its instance to
\\\noindent\LINEmath{
\neg\Maleppp{\wforallvari y 0}\comma
\neg\Lovespp{\wforallvari y 0}{\wforallvari x 0}\comma
\neg\Femaleppp{\wforallvari x 0}\comma
\Lovespp{\wforallvari x 0}{\wforallvari y 0}
}(L5)\par\noindent
which captures the universal meaning of (L0) properly.

\yestop\noindent Instead of a donkey sentence such as (L0) 
that prefers a genuinely universal reading as in \nolinebreak (L5), 
the following donkey sentence prefers a 
partial switch to an existential reading:\notop

\subsubsection{If a bachelor loves a woman, he marries her.\hfill\rm(M0)}\label
{section If a bachelor loves a woman, he marries her.}
If I love three utopian women, I am loved by all of them, but may marry
at most one. Thus\par\noindent\LINEmath{\noparenthesesoplist{
    {\Maleppp{\wforallvari y 0}
        \und\Lovespp{\wforallvari y 0}{\wforallvari x 0}
        \und\Femaleppp{\wforallvari x 0}
 }\oplistimplies{
      \Maleppp{\existsvari y 1}
    \und\Marriespp{\existsvari y 1}{\existsvari x 1}
    \und\Femaleppp{\existsvari x 1}
}}}(M1)\par\noindent
should be refined by application of \bigmaths{\{
\existsvari x 1\mapsto\sforallvari x 1\comma
\existsvari y 1\mapsto\wforallvari y 0\}}{} 
and simplification to\smallfootroom\par\noindent\LINEmath{\begin{array}[b]{l}
\neg\Maleppp{\wforallvari y 0}\comma
\neg\Lovespp{\wforallvari y 0}{\wforallvari x 0}\comma
\neg\Femaleppp{\wforallvari x 0}\comma
\Femaleppp{\sforallvari x 1}
\\\neg\Maleppp{\wforallvari y 0}\comma
\neg\Lovespp{\wforallvari y 0}{\wforallvari x 0}\comma
\neg\Femaleppp{\wforallvari x 0}\comma
\Marriespp{\wforallvari y 0}{\sforallvari x 1}
\\\end{array}}\begin{tabular}[b]{@{}r@{}}(M2a)\\(M2b)
\\\end{tabular}\par\noindent
with \cc\par\noindent\LINEmath{\begin{array}{l@{~~~~~~}l@{~~~~~~}l}
  \app C{\sforallvari x 1}
 &:=
 &\Maleppp{\wforallvari y 0}\nottight{\nottight\implies}
  \Lovespp{\wforallvari y 0}{\sforallvari x 1}\und\Femaleppp{\sforallvari x 1}
\\\end{array}}(C2)\par\noindent
On the one hand, if there is no women loved by the bachelor
\nlbmath{\wforallvari y 0}, both (M2a) and (M2b) are valid.
On the other hand, if there is at least one woman he loves, 
(M2a) \nolinebreak is again valid (due to (C2))
and (M2b) \nolinebreak expresses the intended reading of (M0).

Notice that we indeed
have the possibility to 
let ``woman'' be universal (strong, \math{\delta^-}) and ``her'' existential
(weak, \math{\delta^+}), 
picking one of the women loved by the bachelor---if there are any. 
Our elegant treatment is more flexible than a similar one of \nolinebreak(D)
along supposition theory in \cite{medieval-supposition-theory}.
Moreover, both these treatments are more lucid than
the treatment of 
a sentence in %\cite{heusingeresel} and 
\cite{heusingerepsilon},
which is analogous to \nolinebreak (M0): 
As (12) on \p 183 of
\cite{heusingerepsilon}
we find the example 
\noitem\begin{quote}
{\fraknomath``\germantextthirty''}
\notop\end{quote}\begin{quote}``If a man has a dime, 
he puts it into the meter.''
\getittotheright{(our translation)}
\notop\end{quote}
{\em Mutandis mutatis}\/ and the readability improved, 
the modeling of (M0)
according to \nolinebreak (19a) on \p 185 of \cite{heusingerepsilon} would be
\par\noindent\normalsize\LINEmath{
  \exists f\stopq
  \forall i\stopq\inparenthesesoplist{
  \apptotuple\Lovessymbol
       {\displaypair
          {\varepsilon_i y.\,\Maleppp y}
          {\varepsilon_{\app f i} x.\,\Femaleppp x}}
  \oplistimplies
  \apptotuple\Marriessymbol
       {\displaypair
            {\varepsilon_{a^\ast}y.\,\Maleppp y}
            {\varepsilon_{a^\ast} x.\,\Femaleppp x}}}
}(19a\math')\par\noindent
where the index \nlbmath{a^\ast} of 
\heusinger's indexed \math\varepsilon-operator
(\cfnlb\ \sectref{section linguistic literature})
seems to denote a choice function that chooses
men as \nlbmath i does and women as \nlbmath{\app f i} does. \ 
How \math{a^\ast} is to be formalized
stays unclear in \cite{heusingerepsilon}. \ 
% I guess that the first occurrence of \nlbmath{a^\ast} is to 
% be replaced with \nlbmath i and the second is to fix a special choice
% just as in (C2) above.
The real problem, however, is that
(19a\math') does not represent the intended meaning of (M0): \ 
To wit, take an \nlbmath f such that \app f i always chooses 
a woman not loved by the man chosen by \nlbmath i; \ 
then (\tightemph\exfalsoquodlibet)
all our bachelors may stay unmarried, contradicting (M0).~\footnote
{\label{note technical advantages over Heusinger} \
 {\bf(Technical Disadvantages of \heusinger's Indexed
  \math\varepsilon-operator)}\par\noindent
 When trying to understand the semantics of sentences in natural language,
 it might be the case that
 a representation of the indefinite article  
 with (a variant \nolinebreak of) 
 our new indefinite semantics for the \nlbmath\varepsilon\ offers
 the following advantages compared to \cite{heusingerepsilon}:
 \begin{enumerate}\notop\item
 We do not have to disambiguate a specific from a non-specific usage
 in advance, contrary to \cite{heusingerepsilon} where we have to
 choose between \bigmaths{G(\varepsilon_l x.\,F(x))}{} and
 \bigmaths{\exists i.G(\,\varepsilon_i x.\,F(x))}{} eagerly\@. \ 
 Besides this, the design decision to
 pack the information on specificity into the \math\varepsilon-term
 may be questioned.
 \end{enumerate}\begin{enumerate}\notop\notop\item[2.]
 For a computer implementation, the \math l and \nlbmath i
 in these formulas have to be implemented as something isomorphic to \sfuv s
 (or \fev s)
 anyway,
 so that our representation (\ie\ \bigmaths{G(\sforallvari x{})}{} with \cc\ 
 \bigmaths{\app C{\sforallvari x{}}:=F(\sforallvari x{})})
 saves one level of indirection.
 \end{enumerate}\begin{enumerate}\notop\notop\item[3.]
 Our possibility of a formally verified instantiation of \sfuv s (\cfnlb\ 
 \sectref{section Instantiating Strong Free Universal Variables}
 and \theoref{theorem strong reduces to}(6))
 could provide a formal means in the stepwise process of approaching the
 intended semantics of sentences in natural language.
 \end{enumerate}}\pagebreak
%%%%%%%%%%%%%%%%%%%%%%%%%%%%%%%%%%%%%%%%%%%%%%%%%%%%%%%%%%%%%%%%%%%%%%%%%%%%%%%%
\subsection{Quantifiers for Computing Semantics of Natural Language?}\label
{section quantifiers adequate}
Representation of semantics of sentences and discourses in 
natural language with the help of quantifiers is of surprising difficulty.
The examples in the previous \sectrefs{section geurts}{section donkey} 
indicate that quantified logic is 
problematic as a data structure for computing the semantics 
of sentences and discourses in natural language.
Moreover, as already shown in \sectref{section quantifiers}, 
for some sentences a precise representation with the quantifiers of 
\firstorder\ logic does not exist at all. 
Furthermore, the combinations of different scopes of quantifiers give rise to 
a combinatorial explosion of different readings:
According to \cite[\p\,3]{kollerdiss}, 
the following sentence, which is easy to understand for human beings,
has ``64764 different semantic readings, 
purely due to scope ambiguity\closequotecomma ``even if one specified
syntactic analysis'' ``is fixed\closequotecolonnospace\noitem
\begin{quote}But that would give us all day Tuesday to be 
there.\noitem\end{quote} 
I agree with \cite{hobbs} in
that humans ``do not compute the 120 possible readings'' of 
\noitem\begin{quote}
In most democratic countries most politicians can fool most of the 
people on almost every issue most of the time.\noitem\end{quote}
Even if we can sometimes restrict the number of possible scopings
below \math{n!} for \math n quantifiers, \eg\ by the algorithm of
\cite{hobbsshieber}, the number of possible readings is still too 
\linebreak 
high for computers and human beings.
Therefore, the relation of quantifiers to the semantics 
\linebreak 
of natural language must be questioned. 
Notice that there are no quantifiers in natural language,
and we can avoid them in the computation of their
semantics with the help of the free-variable semantics introduced in 
this \daspaper.
Besides our most rudimental solution, 
we find the three following approaches to overcome quantifiers and scopes
in the literature:\begin{enumerate}

\noitem\item\cite{kollerdiss}
uses standard quantified logic (plus bound variables
outside the lexical scopes of their quantifiers) as basic language 
but leaves the formulas syntactically underspecified.
A drawback seems to be that the actual formulas cannot be accessed.

\noitem\item\cite{hobbs} provides directly accessible formulas, namely
some existentially quantified conjunctions.
These formulas, however, are not likely to be close to the semantics of 
natural language as they are quite unreadable (to me at least). \ 
A modern modeling of the ``typical elements'' of \cite{hobbs} should be a
new form of \fuv s obtained by changing ``some \nlbmath\pi'' 
in \defiref{definition strong validity} into ``each \nlbmath\pi'' as in \cite
[\litdefiref{5.7} (\litdefiref{4.4} in short version)]{wirthgreen}, \ 
\cfnlb\ our \sectref{subsection On the Design of Similar Operators}. \ 
Two different ``typical elements'' of the same set 
(\cf\ \cite[\p\,6 of \WWW\ version]{hobbs}) 
can then be modeled as two variables with the same \cc.
Moreover, note that our use of reduction and instantiation in 
\sectrefs{section If a man loves a woman}
{section If a bachelor loves a woman, he marries her.}
can be easily extended to 
a framework of\emph{weighted abduction} as found in 
\cite[\litchapref{3}]{hobbs-magnum-opus}.

\noitem\item Discourse Representation Theory
\ (\tightemph{DRT}, \ \cf\ \eg\ \cite{drt}, \cite{drtII}) \ 
shares with \cite{hobbs} the preference for 
existentially quantified conjunctions, but is not restricted to them.
Nevertheless, the handling of quantifiers and scopes (or their substitutes)
is quite impractical in DRT---even with the extensions for generalized
quantifiers of \cite{drtII}. \ 
For example, 
DRT \nolinebreak provides only one kind of free variables and no 
``typical elements\closequotecomma
and universal quantification comes only with implications.
Therefore, we expect an integration of our explicit characterization of 
free variables and our general way to introduce new tailored kinds
of free variables into DRT to be beneficial.
Note that also the accessibility restrictions of DRT can be captured by 
our \vc s, admitting more flexibility.\pagebreak\end{enumerate}
\fregename\ \fregelifetime\ invented \firstorder\ logic 
(including some \secondorder\ extension)
in\,1878 (so did \peircename\ independently, \cf\ \cite{peirce-1885})
and \secondorder\ logic including 
\mbox{\math\lambda-abstraction} and a \mbox{\math\iota-operator} in\,1893, 
both under the name ``{\fraknomath\Begriffsschrift}\closequotesemicolon
\cfnlb\ \makeaciteoftwo{begriffsschrift}{frege-grundgesetze} 
and our \noteref{note frege}, respectively. 
\frege\ designed his {\fraknomath\Begriffsschrift}\emph{not} for the task of 
computing the semantics of sentences in natural language, 
but actually---just as \peanoname\ 
\peanolifetime\ his {ideography}, \cfnlb\ \cite{peanoiotabar}---\tightemph
{to \nolinebreak overcome the imprecision and ambiguity of natural language}.
He cannot be blamed for
the trouble quantifiers raise in representation and computation of
the semantics of natural language.
In \nolinebreak \cite{begriffsschrift},
he is well aware of the difference of the semantics
of natural language and his {\fraknomath\Begriffsschrift}
and compares it to that of the naked eye and the microscope. 
Indeed, \frege\ saw the 
{\fraknomath\Begriffsschrift} as fundamentally different 
from natural language and as a substitute for it:

\yestop\par\noindent{\fraknomath\germantexttwenty}\getittotheright
{\cite[\p VI\f, modernized orthography]{begriffsschrift}}
\yestop\par\noindent``If it is a task of philosophy to break
the dominance of natural language over
the human mind \begin{itemize}\notop\item by discovering
the deceptions on the relations of notions resulting from the use of language
often almost inevitably,
\noitem\item by liberating the idea of what spoils it just by the linguistic 
means of expression,\notop\end{itemize}
then my {\fraknomath\Begriffsschrift}---once 
further improved for these aims---will become a 
useful tool for the philosophers. \ 
Of course---as it seems to be unavoidable for any external means of 
representation---also the {\fraknomath\Begriffsschrift}
is not able to represent the idea undistortedly; \ 
but, on the one hand, 
\begin{itemize}\notop\item
it is possible to limit these distortions to the
unavoidable and harmless, and, on the other hand\noitem\item 
a protection against a one-sided % one-sided is okay
influence of one of these means of expression is given 
already because %the means of expression 
those of the {\fraknomath\Begriffsschrift}
are completely different from %the means of expression that are 
those characteristic of language.''
% the thoughts are protected against a one-sided % one-sided is okay
% influence of these distortions already 
% because the means of expression of the {\fraknomath\Begriffsschrift}
% are completely different from the means of expression that are 
% characteristic of language.''
\getittotheright{(our translation)}\end{itemize}
%%%%%%%%%%%%%%%%%%%%%%%%%%%%%%%%%%%%%%%%%%%%%%%%%%%%%%%%%%%%%%%%%%%%%%%%%%%%%% 
\subsection{Conclusion}\label{section conclusion quantification}
In this \sectref{section philosophy of language},  
we have demonstrated our new indefinite semantics for 
\hilbert's \nlbmath\varepsilon\ and our free-variable framework
in a series of interesting applications
provided by standard examples from linguistics.
{\em Can this \whatitisgoodfor?}\/
An investigation of this question requires 
a close collaboration of experts from both linguistics and logics.
Be the answer to this question as it may, 
the field has provided us with an excellent test bed for descriptive terms and
their logical frameworks.
%%%%%%%%%%%%%%%%%%%%%%%%%%%%%%%%%%%%%%%%%%%%%%%%%%%%%%%%%%%%%%%%%%%%%%%%%%%%%
\cleardoublepage
%%%%%%%%%%%%%%%%%%%%%%%%%%%%%%%%%%%%%%%%%%%%%%%%%%%%%%%%%%%%%%%%%%%%%%%%%%%%%
\section{Conclusion}\label
{section conclusion}
Our novel indefinite semantics for \hilbert's \math\varepsilon\
presented in this \daspaper\ was developed to solve the difficult 
soundness problems 
arising during the 
combination of mathematical induction in the liberal style
of \fermat's\emph\descenteinfinie\ with state-of-the-art 
deduction.\footnote{\ 
The \wellfoundedness\ required for the soundness of\emph\descenteinfinie\
gave rise to a notion of reduction which preserves solutions,
\cf\ \defiref{definition strong reduction}. \ 
The liberalized \mbox{\math\delta-rules} as found in \cite{fitting}
do not satisfy this notion. \ 
The addition of our \cc s finally turned out to be the only way to repair 
this defect of the liberalized \math\delta-rules. \ 
\Cfnlb\ \cite{wirthcardinal} for more details.}
Thereby, it \nolinebreak had passed 
an evaluation of its usefulness
even before it was recognized as a candidate for 
the semantics that \hilbertname\ probably had in mind for his 
\nlbmath\varepsilon. \ 
While the speculation on this question will go on,
the semantical framework for \hilbert's \nlbmath\varepsilon\
proposed in this \daspaper\ definitely has the following advantages:
\begin{description}
\noitem\item[Syntax: ]
The requirement of a commitment to a choice is expressed 
syntactically and most clearly
by the sharing of a \sfuv, 
\cf\ \nolinebreak\sectref{section replacing epsilon}.
\noitem\item[Semantics: ]\headroom
The semantics of the \math\varepsilon\ 
is simple and straightforward in the sense that the 
\mbox{\math\varepsilon-operator}
becomes similar to the referential use of the indefinite article 
in some natural languages.
As we have seen in \sectref{section philosophy of language}, it is indeed 
so natural that it provides some help in understanding ideas on philosophy of
language which were not easily accessible before.
Our semantics for the 
\math\varepsilon\ is based on an abstract formal approach 
that extends a semantics 
for closed formulas
(satisfying only very weak requirements, 
 \cfnlb\ \sectref\semanticssectiontitle)
to a semantics with several kinds of free variables: 
existential\nolinebreak\ (\math\gamma),
universal\nolinebreak\ (\deltaminus), 
and \mbox{\math\varepsilon-constrained (\deltaplus)}.
\noitem\item[Reasoning: ]\headroom
In a reductive proof step, our representation of 
an \math\varepsilon-term \bigmath{\varepsilon x.\,A} can be replaced with\emph
{any} term \nlbmath t that satisfies the formula 
\bigmaths{\exists x.\,A\nottight\implies A\{x\tight\mapsto t\}},
\cfnlb\ \sectref{section Instantiating Strong Free Universal Variables}. \
Thus, the soundness of such a replacement is likely to be 
expressible and verifiable in the original calculus.
% While this carries on the tradition of the \math\varepsilon-substitution
% methods, 
Our free-variable framework for the \nlbmath\varepsilon\
is especially convenient for developing proofs in the style of a working
mathematician, \cf\ \makeaciteoftwo{wirthcardinal}{nonpermut}. \ 
Indeed, our approach
makes proof work most simple because
we do not have to consider all proper choices \nlbmath t for \nlbmath x
(as in all other semantical approaches)
but only a single arbitrary one, 
which is fixed in a proof step, just as choices are settled in program steps,
\cf\ \sectref{section do not be afraid}.
\end{description}
Finally, we hope that new semantical framework will help to 
solve further practical and theoretical problems with the \nlbmath\varepsilon\
%(\eg\ in intuitionistic logic, \cf\ \examref{example (E2) again})
and improve the applicability of the \nlbmath\varepsilon\
as a logical tool for description and reasoning.
Although we have only touched the surface of the subject
in \nolinebreak \sectref{subsection On the Design of Similar Operators}, \
a \nolinebreak 
tailoring of operators similar to our \nlbmath\varepsilon\ to meet the
special demands of specification and computation in various areas 
(such as semantics of discourses in natural language) 
seems to be especially promising.
%%%%%%%%%%%%%%%%%%%%%%%%%%%%%%%%%%%%%%%%%%%%%%%%%%%%%%%%%%%%%%%%%%%%%%%%%%%%
%\appendix

\section*{Acknowledgments}\addcontentsline{toc}{section}{Acknowledgments}
This \daspaper\ and its short conference version \cite{wirthhilbertepsilon}
were written because in 2001 \gabbayname\ considered the subject to be
worthwhile. I am grateful for his encouragement.
Furthermore, I would like to thank 
\bellname\ and \horacekname\ for some substantial help, 
\pinkalname\ and 
\wolskaname\ for some guidance through the linguistic
literature, 
\frankename\ 
and an anonymous referee for some well 
justified suggestions for improvement,
and \vandeveirename\ for translations
from and into the Italian language.
\cleardoublepage

\section*{Notes}\addcontentsline{toc}{section}{Notes}

\yestop
\begingroup
\theendnotes
\endgroup

\vfill\cleardoublepage

\nocite{writing-mathematics}
\addcontentsline{toc}{section}{\refname}
\bibliography{herbrandbib}

\end{document}
%%%%%%%%%%%%%%%%%%%%%%%%%%%%%%%%%%%%%%%%%%%%%%%%%%%%%%%%%%%%%%%%%%%%%%%%%%%%%%

%% file: header.tex
\input headerhot
\input headerproof

%Environments
\newcommand\arr[1]{\mbox{$\begin{array}#1\end{array}$}}

{\end{enumerate}\renewcommand{\theenumi}{\arabic{enumi}}}

\newcommand{\enumwithwordinbetween}[3]{
\begin{enumerate}

\item
\notop
{#1}

\vspace{-.5\parskip}
\hspace{-1.0\leftmargin}{#2}

\item
\vspace{-.5\parskip}
\vspace{-1.0\itemsep}
{#3}

\end{enumerate}
\notop
}

\newcommand{\itemizewithwordinbetween}[3]{
\begin{itemize}

\item
\notop
{#1}

\vspace{-.5\parskip}
\hspace{-1.0\leftmargin}{#2}

\item
\vspace{-.5\parskip}
\vspace{-1.0\itemsep}
{#3}

\end{itemize}
\notop
}

\mathcommand\myfootnotemark[1]{^{#1}}
\newcommand\repeatfootnotemark[1]{\myfootnotemark{\ref{#1}}}

%%%%%%%%%%%%%%%%%%%%%%%%%%%%%%%%%%%%%%%%%%%%%%%%%%%%%%%%%%%%%%%%%%%%%%%%%%%%%%
%multi-set notation
\newcommand\repname{{\rm set}}
\mathapplycommand\rep\repname
\mathcommand\repr[1]{{\repname[{#1}]}}
\mathcommand\msa{\langle}
\mathcommand\mse{\rangle}
\mathcommand\msu{\,\sqcup\,}
\mathcommand\msin{{\rm\;in\;}}
\mathcommand\mssetminus{\setminus\!\!\setminus}
\mathcommand\tightmssubseteq{\sqsubseteq}
\mathcommand\mssubseteq{\ \tightmssubseteq\ }
\mathcommand\approxapprox{\approx\:\!\!\approx}
\mathcommand\quasilquasil{\,\lesssim\!\lesssim\,}
\mathcommand\quasibquasib{\,\gtrsim\!\gtrsim\,}
\mathcommand\fmul[1]{{\rm FMul}(#1)}
\mathcommand\smul[1]{{\rm SMul}(#1)}
\mathcommand\multisetwith [2]{\msa\ {#1}\ |\ {#2}\ \mse}
\mathcommand\multisetwithq[3]{\msa\ {#2}\ |_{#1}\ {#3}\ \mse}
%%%%%%%%%%%%%%%%%%%%%%%%%%%%%%%%%%%%%%%%%%%%%%%%%%%%%%%%%%%%%%%%%%%%%%%%%%%%%%
\newcommand\superseteq     {\supseteq}

\newcommand\Algebra{\sig/\cons-algebra}
\newcommand\Algebraprime{\mbox{$\sig'/\cons'$}-algebra}
\newcommand\Model{\sig/\cons-model}
\newcommand\Homo {\sig/\cons-homo\-morphism}
\newcommand\consHomo{\cons-homo\-morphism}
\newcommand\HomoTheorem{Homo\-morphism-Theorem}
\newcommand\HomoLemma{Homo\-morphism-Lemma}
\newcommand\Epi  {\sig/\cons-epi\-morphism}
\newcommand\Iso  {\sig/\cons-iso\-morphism}
\newcommand\isomorphic{\sig/\cons-iso\-mor\-phic}
\newcommand\Congruence{\sig/\cons-congruence}
\newcommand\hiddentermgenerated[2]{#1\mbox{\rm:}#2-term-gene\-rated}
\newcommand\hiddentermgeneratedness[2]{#1\mbox{\rm:}#2-term-gene\-rated\-ness}
\newcommand\dunnodunnotermgenerated   {\hiddentermgenerated\DUNNO\dunno   }
\newcommand\consconstermgenerated     {\hiddentermgenerated\CONS \cons    }
\newcommand\sigconstermgenerated      {\hiddentermgenerated\SIG  \cons    }
\newcommand\sigconstermgeneratedness  {\hiddentermgeneratedness\SIG  \cons    }
\newcommand\sigsigtermgenerated       {\hiddentermgenerated\SIG  \sig     }
\newcommand\sigdunnotermgenerated     {\hiddentermgenerated\SIG  \dunno   }
\newcommand\consconsprimetermgenerated
{\hiddentermgenerated\CONS{\math{\cons'}}}
\newcommand\sigconsprimetermgenerated 
{\hiddentermgenerated\SIG {\math{\cons'}}}

%______________________________________________________________________________
%Orderings

\mathcommand\quasilhd
{\mathop{\mbox{\raisebox{0.31ex}{\math\lhd}\hspace{-0.75em}\raisebox
{-0.6ex}{\math\sim}}}}
\newcommand\quasirhd{\mbox{\raisebox{0.31ex}{$\rhd$}\hspace{-0.75em}\raisebox{-0.6ex}{$\sim$}}}

\mathcommand\rhdrhd{\rhd$\hspace{-0.35em}$\rhd}
\mathcommand\lhdlhd{\lhd$\hspace{-0.21em}$\lhd}

\mathcommand\quasilhdquasilhd{\quasilhd$\hspace{-0.13em}$\quasilhd}

%Subsumption
\newcommand\hiddensubSS{_{_{\rm SS}}}
\mathcommand\antisubsum     {\rhd\hiddensubSS}
\mathcommand\notantisubsum  {\ntriangleright\hiddensubSS}
\mathcommand\subsum         {\lhd\hiddensubSS}
\mathcommand\notsubsum      {\ntriangleleft\hiddensubSS}
\mathcommand\antisubsumeq   {\trianglerighteq\hiddensubSS}
\mathcommand\subsumeq       {\trianglelefteq\hiddensubSS}
\mathcommand\quasisubsum    {\,\quasilhd\raisebox{0.1ex}{$\hiddensubSS$}}
\mathcommand\antiquasisubsum{\,\quasirhd\raisebox{0.1ex}{$\hiddensubSS$}}
\mathcommand\quasiquasisubsum{\quasisubsum\!\!\quasisubsum}
\mathcommand\antiquasiquasisubsum{\antiquasisubsum\!\!\!\antiquasisubsum}

%Orderings on Alegbras
\newcommand\hiddensubH{_{_{\rm H}}}
\newcommand\hiddensubCONS{_{_\CONS}}
\mathcommand\hql   {\,\lesssim\hiddensubH}
\mathcommand\consql{\,\lesssim\hiddensubCONS}
\mathcommand\hl    {\,<       \hiddensubH}
\mathcommand\hleq  {\,\leq    \hiddensubH}
\mathcommand\consl {\,<       \hiddensubCONS}
\mathcommand\conseq{\,\approx \hiddensubCONS}

%_____________________________________________________________________________

\newcommand\Inthesequel{In what follows}
\newcommand\inthesequel{in what follows}
%_____________________________________________________________________________
%Signatures
\newcommand\cons {{\rm cons}}
\mathcommand\sigconsV{\sig/\cons/\V}
\mathcommand\sigconsR{\sig/\cons/\R}
\mathcommand\primesigconsV{\sig'\!/\cons'\!/\V'}
\mathcommand\primesigconsR{\sig'\!/\cons'\!/\R'}
\newcommand\dunno{{\rm sc}}
\newcommand\DUNNO{{\rm SC}}
\mathcommand\SIGCONS   {\{\SIG,\CONS\}}
\mathcommand\sigsortstimes{\SIGCONS\tight\times\sigsorts}

\mathcommand\TERMSSYM
{\mathcal{T\hspace{-.20em}E\hspace{-.08em}R\hspace{-.16em}M\hspace{-.10em}S}}
\mathapplycommand\condterms{\TERMSSYM}
\mathcommand\kurzregel{((l,r),C)}
\mathcommand\kurzregelprime{((l',r'),C')}
\mathcommand\kurzregelindex[1]{((l_{#1},r_{#1}),C_{#1})}
% \mathcommand\regel{((l,r),\condition)}
% \mathcommand\condition{L_{0},\ldots,L_{n-1}}
\mathapplycommand\lhs{\rm lhs}

\mathcommand\red{\redsimple} %conflicts with ps tricks

%%%%%%%%%%%%%%%%%%%%%%%%%%%%%%%%%%%%%%%%%%%%%%%%%%%%%%%%%%%%%%%%%%%%%%%%%%%%%%
%triples

\mathcommand\lemms{L}
\mathcommand\hypos{H}
\mathcommand\goals{G}
\mathcommand\lemmsprime{\lemms'}
\mathcommand\hyposprime{\hypos'}
\mathcommand\goalsprime{\goals'}
\mathcommand\lemmsprimeprime{\lemms''}
\mathcommand\hyposprimeprime{\hypos''}
\mathcommand\goalsprimeprime{\goals''}
%\triple-->\oldtriple
\mathcommand\oldtriple            {(\lemms   ,\hypos   ,\goals  )}
\mathcommand\inittriple        {(\emptyset,\emptyset,\goals  )}
\mathcommand\triplehelp[1]     {(\lemms#1,\hypos#1 ,\goals#1)}
\mathcommand\tripleprime       {\triplehelp'}
\mathcommand\triplenogoalsprime{(\lemmsprime,\hyposprime,\emptyset  )}
\mathcommand\tripleprimeprime  {\triplehelp{''}}
\mathcommand\tripleindex[1]    {\triplehelp{_{#1}}}

%%%%%%%%%%%%%%%%%%%%%%%%%%%%%%%%%%%%%%%%%%%%%%%%%%%%%%%%%%%%%%%%%%%%%%%%%%%%%%
%Contextual Rewriting
\mathcommand\constcong[1]{\,\,\sim_{\!_{#1}}\,}

%Induction
\mathapplycommand\avail{\rm\Av ail}

%%%%%%%%%%%%%%%%%%%%%%%%%%%%%%%%%%%%%%%%%%%%%%%%%%%%%%%%%%%%%%%%%%%%%%%%%%%%%%
\def\emph#1{\/ {\itshape#1}\/}
\newcommand\tightemph[1]{\/{\itshape#1}\/}
\newcommand\openquoteemph[1]{\ ``\hskip-0.15em{\itshape#1}\/}
%\hskip-0.18em is too much
\newcommand\emphtight[1]{\ {\itshape#1}}

%% file: headerproof.tex
\def\proof         {\relax}
\def\proofqed      {\relax}
\def\proofparsep   {\relax}
\def\proofparsepqed{\relax}
\def\proofshort    {\relax}
\newcommand\Proofof{Proof of}
\renewenvironment{proof}[1]{\yestop\begin{sloppypar}\noindent
{\bf\Proofof\ {#1}}\\}{\end{sloppypar}}
\renewenvironment{proofqed}[1]
{\begin{sloppypar}\def\fooqed{#1}\noindent{\bf\Proofof\ \fooqed}}
{\QEDbf\fooqed\end{sloppypar}}
\renewenvironment{proofparsep}[1]{\parindent=0pt\begin
{sloppypar}\noindent{\bf\Proofof\ {#1}}\nopagebreak\par}
{\end{sloppypar}}
\renewenvironment{proofparsepqed}[1]{\parindent=0pt\begin
{sloppypar}\def\fooqed{#1}\noindent{\bf\Proofof\ \fooqed}\nopagebreak\par}
{\nopagebreak\QEDbf\fooqed\end{sloppypar}}
\renewenvironment{proofshort}[1]{\begin{sloppypar}\noindent
{\bf\Proofof\ {#1}:}}{\end{sloppypar}}

%% file: headerforformulas.tex
\mathcommand\ident[1]{\mathsf{#1}}
\newcommand\plussymbol  {\ident{+}}
\newcommand\minussymbol {\ident{-}}
\newcommand\dividesymbol{\ident{/}}
\newcommand\timessymbol {\ident{*}}

%sorts
\newcommand\eins    {\ident{1}}
\newcommand\nat     {\ident{nat}}
\newcommand\posnat  {\ident{posnat}}
\newcommand\zeronat {\ident{zero}}
\newcommand\ord     {\ident{ord}}
\newcommand\ORD     {\ident{ORD}}
\newcommand\lists   {\ident{list}}
\newcommand\set     {\ident{set}}
\newcommand\stream  {\ident{stream}}
\newcommand\term    {\ident{term}}
\newcommand\nonemptylists{\ident{nonemptylist}}
\newcommand\emptylists   {\ident{emptylist}}
\newcommand\bool    {\ident{bool}}
\newcommand\intsort {\ident{int}}
\newcommand\funsort [2]{{#1\math\rightarrow#2}}
\newcommand\pairsort[2]{{#1\math{\!\times\!}#2}}
\newcommand\sumsort [2]{{#1\math{+}#2}}
\newcommand\ortreesort{\ident{or}}
\newcommand\andtreesort{\ident{and}}

%functionsymbols
\newcommand\naturalssymbol{\ident{naturals}}
\newcommand\gensymsymbol{\ident{gensym}}
\mathcommand\mbpsymbol{\ident{m\hspace{-0.055em}b\hspace{-0.045em}p}}
\newcommand\appsymbol   {\ident a}
\newcommand\csymbol     {\ident c}
\newcommand\esymbol     {\ident e}
\newcommand\fsymbol     {\ident f}
\newcommand\gsymbol     {\ident g}
\newcommand\hsymbol     {\ident h}
\newcommand\ksymbol     {\ident k}
\newcommand\psymbol     {\ident p}
\newcommand\ssymbol     {\ident s}
\newcommand\Everysymbol {\ident{Every}}
\newcommand\Permsymbol {\ident{Perm}}
\newcommand\RExistssymbol{\ident{Rexists}}
\newcommand\invertsymbol{\ident{invert}}
\newcommand\invsymbol{\ident{inv}}
\newcommand\abssymbol   {\ident{abs}}
\newcommand\cnssymbol   {\ident{cons}}
\mathcommand\cnsindexsymbol[1]{\ident{cons}_{#1}}
\newcommand\carsymbol   {\ident{car}}
\newcommand\cardsymbol  {\ident{card}}
\newcommand\cdrsymbol   {\ident{cdr}}
\newcommand\lengthsymbol{\ident{length}}
\newcommand\dlsymbol    {\ident{dl}}
\newcommand\dloncesymbol{\ident{delonce}}
\newcommand\rcsymbol    {\ident{rc}}
\newcommand\brsymbol    {\ident{br}}
\newcommand\revtailsymbol{\ident{revtail}}
\newcommand\revsymbol{\ident{rev}}
\newcommand\appendsymbol {\ident{append}}
\newcommand\zeropredicatesymbol{\ident{zerop}}
\newcommand\eqsymbol        {\ident{eq}}
\newcommand\ifthensymbol    {\mbox{\ident{If{}Then}}}
\newcommand\ifthenelsesymbol{\mbox{\ident{If{}ThenElse}}}
\mathcommand\eqindexsymbol        [1]{\eqsymbol        _{#1}}
\mathcommand\ifthenindexsymbol    [1]{\ifthensymbol    _{#1}}
\mathcommand\ifthenelseindexsymbol[1]{\ifthenelsesymbol_{#1}}
\newcommand\orsymbol    {\ident{or}}
\newcommand\andsymbol   {\ident{and}}
\newcommand\leqsymbol   {\ident{leq}}
\newcommand\lessymbol   {\ident{less}}
\newcommand\lexsymbol   {\ident{lex}}
\newcommand\acksymbol   {\ident{ack}}
\newcommand\switchsymbol{\ident{switch}}
\newcommand\swatchsymbol{\ident{swatch}}
\newcommand\diveinssymbol{\ident{div1}}
\newcommand\divzweisymbol{\ident{div2}}
\newcommand\divrestsymbol{\ident{divrest}}
\newcommand\diveinstailsymbol{\ident{div1tail}}
\newcommand\divzweitailsymbol{\ident{div2tail}}

\newcommand\turingmachinesymbol{\ident T}
\newcommand\terminatespsymbol  {\ident{terminatesp}}
\newcommand\statesymbol        {\ident{state}}
\newcommand\cmdsymbol          {\ident{cmd}}
\newcommand\nthsymbol          {\ident{nth}}
\newcommand\doublesymbol       {\ident{double}}

\newcommand\ppsymbol           {\ident{p}}
\newcommand\qpsymbol           {\ident{q}}
\newcommand\Epsymbol           {\ident{E}}
\newcommand\Ppsymbol           {\ident{P}}
\newcommand\Qpsymbol           {\ident{Q}}
\newcommand\Marriessymbol      {\ident{Marries}}
\newcommand\Lovessymbol        {\ident{Loves}}
\newcommand\StolenBysymbol     {\ident{StolenBy}}
\newcommand\Humansymbol        {\ident{Human}}
\newcommand\Evensymbol         {\ident{Even}}
\newcommand\Oddsymbol          {\ident{Odd}}
\newcommand\Primesymbol        {\ident{Prime}}
\newcommand\EveryPairsymbol   {\ident{EveryPair}}
\newcommand\Givesymbol         {\ident{Give}}
\newcommand\Fathersymbol       {\ident{Father}}
\newcommand\Elephantpsymbol    {\ident{Elephant}}
\newcommand\Flowerpsymbol    {\ident{Flower}}
\newcommand\Germanpsymbol      {\ident{German}}
\newcommand\Bicyclepsymbol     {\ident{Bicycle}}
\newcommand\Hugepsymbol        {\ident{Huge}}
\newcommand\Animalpsymbol      {\ident{Animal}}
\newcommand\Malepsymbol        {\ident{Male}}
\newcommand\Boypsymbol         {\ident{Boy}}
\newcommand\Girlpsymbol        {\ident{Girl}}
\newcommand\Femalepsymbol      {\ident{Female}}
\newcommand\Roundpsymbol       {\ident{Round}}
\newcommand\Quadrangularpsymbol{\ident{Quadrangular}}
\newcommand\Metpsymbol         {\ident{Met}}
\newcommand\Kissedpsymbol      {\ident{Kissed}}
\newcommand\Bishopsymbol       {\ident{Bishop}}
\newcommand\mindexsymbol[1]{\existsvari w{#1}}

\newcommand\nonnegpsymbol      {\ident{nonnegp}}
\newcommand\wellsymbol         {\ident{well}}
\newcommand\welltailsymbol     {\ident{welltail}}
\newcommand\varsymbol          {\ident{var}}
\newcommand\aritysymbol        {\ident{arity}}

\newcommand\whilesymbol        {\ident{while}}

\newcommand\nullsymbol         {\ident{null}}
\newcommand\hdsymbol           {\ident{hd}}
\newcommand\tlsymbol           {\ident{tl}}
\newcommand\insymbol           {\ident{in}}
\newcommand\applysymbol        {\ident{app}}
\newcommand\termsymbol         {\ident{term}}
%%%%%%%%%%%%%%%%%%%%%%%%%%%%%%%%%%%%%%%%%%%%%%%%%%%%%%%%%%%%%%%%%%%%%%%%%%%%%%
%formulas
\mathcommand\tightim{\longrightarrow}
\mathcommand\im{\ \tightim\ }
\mathcommand\rs{\:\rulesugar\:\:}
\mathcommand\rulesugar{\longleftarrow}

%%%%%%%%%%%%%%%%%%%%%%%%%%%%%%%%%%%%%%%%%%%%%%%%%%%%%%%%%%%%%%%%%%%%%%%%%%%%%%
\mathcommand\doublepp[1]      {\doublesymbol   \beginargs{#1}\allargs}
\mathcommand\aritypp[1]      {\aritysymbol   \beginargs{#1}\allargs}
\mathcommand\lengthpp[1]      {\lengthsymbol   \beginargs{#1}\allargs}
\mathcommand\wellpp[1]      {\wellsymbol   \beginargs{#1}\allargs}
\mathcommand\welltailpp[1]      {\welltailsymbol   \beginargs{#1}\allargs}
\mathcommand\varpp[1]      {\varsymbol   \beginargs{#1}\allargs}
\mathcommand\divrestpp[2]    {\divrestsymbol\beginargs{#1}\separgs{#2}\allargs}
\mathcommand\diveinspp[2]    {\diveinssymbol\beginargs{#1}\separgs{#2}\allargs}
\mathcommand\divzweipp[3]    {\divzweisymbol\beginargs{#1}\separgs{#2}
\separgs{#3}\allargs}
\mathcommand\diveinstailpp[4]    {\diveinstailsymbol\beginargs{#1}\separgs{#2}
\separgs{#3}\separgs{#4}\allargs}
\mathcommand\divzweitailpp[6]    {\divzweitailsymbol\beginargs{#1}\separgs{#2}
\separgs{#3}\separgs{#4}\separgs{#5}\separgs{#6}\allargs}
\mathcommand\mbppp[2]         {\mbpsymbol   \beginargs{#1}\separgs{#2}\allargs}
\mathcommand\revpp[1]     
{\revsymbol\beginargs{#1}\allargs}
\mathcommand\revppi[2]     
{\revsymbol^{#1}\beginargs{#2}\allargs}
\mathcommand\revtailpp[2]     
{\revtailsymbol\beginargs{#1}\separgs{#2}\allargs}
\mathcommand\revtailppi[3]
{\revtailsymbol^{#1}\beginargs{#2}\separgs{#3}\allargs}
\mathcommand\Permpp[2]     
{\Permsymbol\beginargs{#1}\separgs{#2}\allargs}
\mathcommand\Permppi[3]
{\Permsymbol^{#1}\beginargs{#2}\separgs{#3}\allargs}
\mathcommand\appendpp[2]      
{\appendsymbol \beginargs{#1}\separgs{#2}\allargs}
\mathcommand\appendppi[3]      
{\appendsymbol^{#1}\beginargs{#2}\separgs{#3}\allargs}
\mathcommand\Everypp[2]      
{\Everysymbol \beginargs{#1}\separgs{#2}\allargs}
\mathcommand\RExistspp[1]      
{\RExistssymbol \beginargs{#1}\allargs}
\mathcommand\appendlongpp[2]      
{\appendsymbol\left(\begin{array}{@{}l@{}}{#1}\separgs\\{#2}\end{array}\right)}
\mathcommand\cnspp[2]         {\cnssymbol   \beginargs{#1}\separgs{#2}\allargs}
\mathcommand\cnsppi[3]       {\cnssymbol^{#1}\beginargs{#2}\separgs{#3}\allargs}
\mathcommand\cnsindexpp[3]
{\cnsindexsymbol{#1}\beginargs{#2}\separgs{#3}\allargs}
\mathcommand\dlpp[2]          {\dlsymbol    \beginargs{#1}\separgs{#2}\allargs}
\mathcommand\dloncepp[2]      {\dloncesymbol\beginargs{#1}\separgs{#2}\allargs}
\mathcommand\dlonceppi[3]{\dloncesymbol^{#1}\beginargs{#2}\separgs{#3}\allargs}
\mathcommand\rcpp[2]          {\rcsymbol    \beginargs{#1}\separgs{#2}\allargs}
\mathcommand\brpp[2]          {\brsymbol    \beginargs{#1}\separgs{#2}\allargs}
\mathcommand\orpp[2]          {\orsymbol    \beginargs{#1}\separgs{#2}\allargs}
\mathcommand\andpp[2]         {\andsymbol   \beginargs{#1}\separgs{#2}\allargs}
\mathcommand\shortcnspp[2]    {\csymbol     \beginargs{#1}\separgs{#2}\allargs}
\mathcommand\tightshortcnspp[2]
{\csymbol\beginargs{#1}\tightsepargs{#2}\allargs}
\mathcommand\spp[1]           {\ssymbol     \beginargs{#1}\allargs}
\mathcommand\sppiterated[2]   {\ssymbol^{#1}\beginargs{#2}\allargs}
\mathcommand\ppp[1]           {\psymbol     \beginargs{#1}\allargs}
\mathcommand\pppiterated[2]   {\psymbol^{#1}\beginargs{#2}\allargs}
\mathcommand\zeropp           {\ident 0}
\mathcommand\Julietpp         {\ident{Juliet}}
\mathcommand\Romeopp          {\ident{Romeo}}
\mathcommand\Ipp              {\ident I}
\mathcommand\onepp            {\ident1}
\mathcommand\twopp            {\ident2}
\mathcommand\threepp          {\ident3}
\mathcommand\invertpp[1]      {\invertsymbol\beginargs{#1}\allargs}
\mathcommand\invpp[1]         {\invsymbol\beginargs{#1}\allargs}
\mathcommand\abspp[1]         {\abssymbol\beginargs{#1}\allargs}
\mathcommand\naturalspp[1]    {\naturalssymbol\beginargs{#1}\allargs}
\mathcommand\gensympp[1]      {\gensymsymbol\beginargs{#1}\allargs}
\mathcommand\nilpp            {\ident{nil}}
%\mathcommand\nilindexpp[1]    {\ident{nil}_{#1}}
\mathcommand\falsepp          {\ident{false}}
\mathcommand\truepp           {\ident{true}}
\mathcommand\FALSEpp          {\ident{FALSE}}
\mathcommand\TRUEpp           {\ident{TRUE}}
\mathcommand\weirdppp         {\ident{weirdp}}
\mathcommand\ambigppp         {\ident{ambigp}}
\mathcommand\zeropredicatepp[1]{\zeropredicatesymbol\beginargs{#1}\allargs}
\mathcommand\cppeins       [1]{\csymbol     \beginargs{#1}\allargs}
\mathcommand\cppzwei       [2]{\csymbol\beginargs{#1}\separgs{#2}\allargs}
\mathcommand\eppeins       [1]{\esymbol     \beginargs{#1}\allargs}
\mathcommand\fppeins       [1]{\fsymbol     \beginargs{#1}\allargs}
\mathcommand\fppeinsindex  [2]{\fsymbol_{#1}\beginargs{#2}\allargs}
\mathcommand\fppeinsiterated[2]{\fsymbol^{#1}\beginargs{#2}\allargs}
\mathcommand\gppeins       [1]{\gsymbol     \beginargs{#1}\allargs}
\mathcommand\gppzwei       [2]{\gsymbol     \beginargs{#1}\separgs{#2}\allargs}
\mathcommand\hppeins       [1]{\hsymbol     \beginargs{#1}\allargs}
\mathcommand\kppeins       [1]{\ksymbol     \beginargs{#1}\allargs}
\mathcommand\appzero          {\ident a}
\mathcommand\bppzero          {\ident b}
\mathcommand\cppzero          {\ident c}
\mathcommand\dppzero          {\ident d}
\mathcommand\eppzero          {\ident e}
\mathcommand\eqindexpp[3]{\eqindexsymbol{#1}\beginargs{#2}\separgs{#3}\allargs}
\mathcommand\ifthenindexpp
[3]{\ifthenindexsymbol{#1}\beginargs{#2}\separgs{#3}\allargs}
\mathcommand\ifthenelseindexpp
[4]{\ifthenelseindexsymbol{#1}\beginargs{#2}\separgs{#3}\separgs{#4}\allargs}
\mathcommand\eqpp[2]{\eqsymbol\beginargs{#1}\separgs{#2}\allargs}
\mathcommand\leqpp[2]{\leqsymbol\beginargs{#1}\separgs{#2}\allargs}
\mathcommand\lespp[2]{\lessymbol\beginargs{#1}\separgs{#2}\allargs}
\mathcommand\lexpp[3]{\lexsymbol\beginargs{#1}\separgs{#2}\separgs{#3}\allargs}
\mathcommand\ackpp[2]{\acksymbol\beginargs{#1}\separgs{#2}\allargs}
\mathcommand\switchpp[1]{\switchsymbol\beginargs{#1}\allargs}
\mathcommand\swatchpp[1]{\swatchsymbol\beginargs{#1}\allargs}
\mathcommand\whilepp[2]{\whilesymbol\beginargs{#1}\separgs{#2}\allargs}
\mathcommand\nullpp[1]{\nullsymbol\beginargs{#1}\allargs}
\mathcommand\nullppiterated[2]{\nullsymbol^{#1}\beginargs{#2}\allargs}
\mathcommand\hdpp[1]{\hdsymbol\beginargs{#1}\allargs}
\mathcommand\hdppiterated[2]{\hdsymbol^{#1}\beginargs{#2}\allargs}
\mathcommand\tlpp[1]{\tlsymbol\beginargs{#1}\allargs}
\mathcommand\tlppiterated[2]{\tlsymbol^{#1}\beginargs{#2}\allargs}
\mathcommand\inpp[2]{\insymbol\beginargs{#1}\separgs{#2}\allargs}
\mathcommand\inppiterated[3]{\insymbol^{#1}\beginargs{#2}\separgs{#3}\allargs}
\mathcommand\applypp[2]{\applysymbol\beginargs{#1}\separgs{#2}\allargs}
\mathcommand\applyppiterated
[3]{\applysymbol^{#1}\beginargs{#2}\separgs{#3}\allargs}
\mathcommand\termpp[2]{\termsymbol\beginargs{#1}\separgs{#2}\allargs}
\mathcommand\setpp[1]{\set\beginargs{#1}\allargs}

\mathcommand\Tpp[6]{\turingmachinesymbol\beginargs{#1}\separgs{#2}\separgs
{#3}\separgs{#4}\separgs{#5}\separgs{#6}\allargs}
\mathcommand\Tppseven[7]{\turingmachinesymbol\beginargs{#1}\separgs{#2}\separgs
{#3}\separgs{#4}\separgs{#5}\separgs{#6}\separgs{#7}\allargs}
\mathcommand\foreverppp[6]{\ident{foreverp}\beginargs{#1}\separgs{#2}\separgs
{#3}\separgs{#4}\separgs{#5}\separgs{#6}\allargs}
\mathcommand\terminatesppp[6]{\terminatespsymbol\beginargs{#1}\separgs
{#2}\separgs{#3}\separgs{#4}\separgs{#5}\separgs{#6}\allargs}
\mathcommand\terminatespppone[1]{\terminatespsymbol \beginargs{#1}\allargs}
\mathcommand\statepp
[3]{\statesymbol\beginargs{#1}\separgs{#2}\separgs{#3}\allargs}
\mathcommand\tightstatepp
[3]{\statesymbol\beginargs{#1}\tightsepargs{#2}\tightsepargs{#3}\allargs}
\mathcommand\cmdpp
[3]{\cmdsymbol  \beginargs{#1}\separgs{#2}\separgs{#3}\allargs}
\mathcommand\tightcmdpp
[3]{\cmdsymbol  \beginargs{#1}\tightsepargs{#2}\tightsepargs{#3}\allargs}
\mathcommand\stoppp           {\ident{stop}}
\mathcommand\leftpp           {\ident{left}}
\mathcommand\rightpp          {\ident{right}}
\mathcommand\nthpp         [2]{\nthsymbol  \beginargs{#1}\separgs{#2}\allargs}
\mathcommand\pppp          [1]{\ppsymbol\beginargs{#1}            \allargs}
\mathcommand\qppp          [2]{\qpsymbol\beginargs{#1}\separgs{#2}\allargs}
\mathcommand\Eppp          [1]{\Epsymbol\beginargs{#1}            \allargs}
\mathcommand\Epppzwei      [2]{\Epsymbol\beginargs{#1}\separgs{#2}\allargs}
\mathcommand\Pppp          [1]{\Ppsymbol\beginargs{#1}            \allargs}
\mathcommand\Ppppdrei      
[3]{\Ppsymbol\beginargs{#1}\separgs{#2}\separgs{#3}\allargs}
\mathcommand\Ppppvier
[4]{\Ppsymbol\beginargs{#1}\separgs{#2}\separgs{#3}\separgs{#4}\allargs}
\mathcommand\Qppp          [2]{\Qpsymbol\beginargs{#1}\separgs{#2}\allargs}
\mathcommand\Qpppeins      [1]{\Qpsymbol\beginargs{#1}\allargs}
\mathcommand\Qpppdrei      
[3]{\Qpsymbol\beginargs{#1}\separgs{#2}\separgs{#3}\allargs}
\mathcommand\Fatherpp      [2]{\Fathersymbol\beginargs{#1}\separgs{#2}\allargs}
\mathcommand\Marriespp     [2]{\Marriessymbol\beginargs{#1}\separgs{#2}\allargs}
\mathcommand\Lovespp       [2]{\Lovessymbol\beginargs{#1}\separgs{#2}\allargs}
\mathcommand\StolenBypp    [2]
{\StolenBysymbol\beginargs{#1}\separgs{#2}\allargs}
\mathcommand\Humanpp       [1]{\Humansymbol\beginargs{#1}\allargs}
\mathcommand\Evenpp        [1]{\Evensymbol\beginargs{#1}\allargs}
\mathcommand\Evenppi       [2]{\Evensymbol^{#1}\beginargs{#2}\allargs}
\mathcommand\Oddpp         [1]{\Oddsymbol\beginargs{#1}\allargs}
\mathcommand\Primepp       [1]{\Primesymbol\beginargs{#1}\allargs}
\mathcommand\EveryPairpp  [2]{\EveryPairsymbol\beginargs{#1}\separgs
{#2}\allargs}
\mathcommand\mindexppeins  [2]{\mindexsymbol{#1}\beginargs{#2}\allargs}
\mathcommand\Givepp        [3]{\Givesymbol
\beginargs{#1}\separgs{#2}\separgs{#3}\allargs}
\mathcommand\mindexppzwei  [3]{\mindexsymbol
{#1}\beginargs{#2}\separgs{#3}\allargs}
\mathcommand\mindexppdrei  [4]{\mindexsymbol
{#1}\beginargs{#2}\separgs{#3}\separgs{#4}\allargs}

\mathcommand\nonnegppp     [1]{\nonnegpsymbol\beginargs{#1}\allargs}

\mathcommand\anonymouscsymbol{c}
\mathcommand\anonymouscindexsymbol[1]{\anonymouscsymbol_{#1}}
\mathcommand\anonymousfsymbol{f}
\mathcommand\anonymouscpp
[2]{\anonymouscsymbol\beginargs{#1}\separgs\ldots\separgs{#2}\allargs}
\mathcommand\anonymouscindexpp
[3]{\anonymouscindexsymbol{#1}\beginargs{#2}\separgs\ldots\separgs{#3}\allargs}
\mathcommand\anonymousfpp
[2]{\anonymousfsymbol\beginargs{#1}\separgs\ldots\separgs{#2}\allargs}
\mathcommand\coerceindexpp[3]{[#3]_{#1}^{#2}}

\mathcommand\Elephantppp    [1]{\Elephantpsymbol\beginargs{#1}\allargs}
\mathcommand\Flowerppp      [1]{\Flowerpsymbol  \beginargs{#1}\allargs}
\mathcommand\Bicycleppp     [1]{\Bicyclepsymbol \beginargs{#1}\allargs}
\mathcommand\Germanppp      [1]{\Germanpsymbol  \beginargs{#1}\allargs}
\mathcommand\Hugeppp        [1]{\Hugepsymbol    \beginargs{#1}\allargs}
\mathcommand\Animalppp      [1]{\Animalpsymbol  \beginargs{#1}\allargs}
\mathcommand\Maleppp        [1]{\Malepsymbol    \beginargs{#1}\allargs}
\mathcommand\Boyppp         [1]{\Boypsymbol     \beginargs{#1}\allargs}
\mathcommand\Girlppp        [1]{\Girlpsymbol    \beginargs{#1}\allargs}
\mathcommand\Femaleppp      [1]{\Femalepsymbol  \beginargs{#1}\allargs}
\mathcommand\Roundppp       [1]{\Roundpsymbol   \beginargs{#1}\allargs}
\mathcommand\Bishoppp       [1]{\Bishopsymbol   \beginargs{#1}\allargs}
\mathcommand\Quadrangularppp[1]{\Quadrangularpsymbol  \beginargs{#1}\allargs}
\mathcommand\Kissedppp[2]{\Kissedpsymbol\beginargs{#1}\separgs{#2}\allargs}
\mathcommand\Metppp[2]   {\Metpsymbol   \beginargs{#1}\separgs{#2}\allargs}

%%%%%%%%%%%%%%%%%%%%%%%%%%%%%%%%%%%%%%%%%%%%%%%%%%%%%%%%%%%%%%%%%%%%%%%%%%%%%%%
%%GENERAL STUFF
\newcommand\bound     {{\rm bound}}
\newcommand\free      {{\rm free}}

\mathcommand\Vtripleindex[3]{\V\!_{{#1},\,{#2},\,{#3}}}
\mathcommand\Vdoubleindex[2]{\V\!_{{#1},\,{#2}}}
\mathcommand\Vsingleindex[1]{\V\!_{{#1}}}

\newcommand\exrelation{\mbox{\math\Gammaoffont-}relation}
\newcommand\unrelation{\mbox{\math\Deltaoffont-}relation}
\mathcommand\Erel[1]{\Gammaoffont\!_{#1}}
\mathcommand\Urel[1]{\Deltaoffont_{#1}}

\newcommand\exRsub  {\math R-sub\-sti\-tu\-tion  on \nolinebreak\Vsome}
\newcommand\aexRsub {an \math R-sub\-sti\-tu\-tion  on \nolinebreak\Vsome}
\newcommand\exRsubs {\math R-sub\-sti\-tu\-tions on \nolinebreak\Vsome}
\newcommand\ExRSubs {\math R-Sub\-sti\-tu\-tions on \nolinebreak\Vsome}
\newcommand\Rsub    {\math R-sub\-sti\-tu\-tion}
\newcommand\aRsub   {an \math R-sub\-sti\-tu\-tion}
\newcommand\RSub    {\math R-Sub\-sti\-tu\-tion}
\newcommand\Rsubs   {\math R-sub\-sti\-tu\-tions}
\newcommand\RSubs   {\math R-Sub\-sti\-tu\-tions}
\newcommand\aRprimesub{an \math{R'}-sub\-sti\-tu\-tion}
\newcommand\aexRprimesub{\aRprimesub\ on \nolinebreak\Vsome}
\newcommand\exemptysetsub
{\math\emptyset-sub\-sti\-tu\-tion  on \nolinebreak\Vsome}

\newcommand\quasiexRsubs
{\Rsubs\ on \nolinebreak\math{\Vsome\cup\Vsall}}
\newcommand\aquasiexRsub
{\aRsub\ on \nolinebreak\math{\Vsome\cup\Vsall}}
\newcommand\isquasiexistential{is a substitution on \nlbmath{\Vsome\cup\Vsall}}

%\newcommand\sentencewithquasi
%{For an \Rsub\ \nlbmath{\sigma}\ on \math{\Vsome\,\opt{\cup\,\Vsall}}}
\newcommand\aexRsuboptquasi
{\aRsub\ on \nolinebreak\math{\Vsome\,\opt{\cup\,\Vsall}}}

\newcommand\strongsigmaupdate{\math\sigma-update}

\newcommand\strongexvalof[2]{\pair{#1}{#2}-valu\-a\-tion}
\newcommand\astrongexvalof[2]{an \strongexvalof{#1}{#2}}
\newcommand\strongexRval {\strongexvalof\salgebra R}
\newcommand\StrongexRVal {\pair\salgebra R-Valu\-a\-tion}
\newcommand\strongexRvals{\pair\salgebra R-valu\-a\-tions}
\newcommand\StrongexRVals{\pair\salgebra R-Valu\-a\-tions}
\newcommand\astrongexRval{an \strongexRval}
\newcommand\strongexRprimeval{\pair\salgebra{R'}-valu\-a\-tion}
\newcommand\astrongexRprimeval{an \strongexRprimeval}

\newcommand\cc{choice-condition}
\newcommand\CC{Choice-Condition}
\newcommand\Cc{Choice-condition}
\newcommand\vc{vari\-able-con\-di\-tion}
\newcommand\VC{Vari\-able-Con\-di\-tion}
\newcommand\Vc{Vari\-able-con\-di\-tion}
\newcommand\semanticobject{\mbox{\math\Sigmaoffont-}struc\-ture}

\newcommand\validity[2]{\pair{#1}{#2}-validity}
\newcommand\Validity[2]{\pair{#1}{#2}-Validity}
\newcommand\valid[2]{\pair{#1}{#2}-valid}
\newcommand\validstrongtrip[3]{\trip{#1}{#2}{#3}-valid}
\newcommand\validitystrongtrip[3]{\trip{#1}{#2}{#3}-validity}

\newcommand\stronglyreduces[3]{\pair{#1}{#2}-reduces}
\mathcommand\theRprimefromstrongtoweak{
  \inparenthesesinlinetight{
     \domres\id{\Vwall\cup\Vsome\setminus\RAN\varsigma}
     \nottight{\nottight\uplus}
     \reverserelation\varsigma
  }
  \nottight{\circ}
  \ranres
    {\transclosureinline R}
    {\Vwall\cup\Vsome\setminus\RAN\varsigma}
  \nottight{\nottight{\nottight{\uplus}}}
  \Vsome\tighttimes\Vsall
}

\mathcommand\deltaminus{\delta^-}
\mathcommand\deltaplus{\delta^+}
\mathcommand\deltaplusplus{\delta^{+^+}}
\mathcommand\deltastar{\delta^*}
\mathcommand\deltastarstar{\delta^{*^*}}
\newcommand\varihelper[1]{free \discretionary
{\mbox{\math{#1}-vari-}}{\mbox{able}}{\mbox{\math{#1}-variable}}}
\newcommand\Varihelper[1]{Free \discretionary
{\mbox{\math{#1}-vari-}}{\mbox{able}}{\mbox{\math{#1}-variable}}}
\newcommand\fev {\varihelper\gamma}
\newcommand\fuv {\varihelper\delta}
\newcommand\wfuv{\varihelper{\delta^-}}
\newcommand\sfuv{\varihelper{\delta^+}}
\newcommand\Fev {\Varihelper\gamma}
\newcommand\Fuv {\Varihelper\delta}
\newcommand\Wfuv{\Varihelper{\delta^-}}
\newcommand\Sfuv{\Varihelper{\delta^+}}
\newcommand\FUV {Free \math{\delta  }-Variable}
\newcommand\SFUV {Free \math{\delta^+}-Variable}

\mathcommand\Vall     {\Vsingleindex\indexdelta         }
\mathcommand\Vwall    {\Vsingleindex\indexdeltaminu     }
\mathcommand\Vsall    {\Vsingleindex\indexdeltaplus     }
\mathcommand\Vgsome   {\Vsingleindex\indexgammaplus     }
\mathcommand\Vsome    {\Vsingleindex\indexgamma         }
\mathcommand\Vfree    {\Vsingleindex\indexfree          }
\mathcommand\Vbound   {\Vsingleindex\indexbound         }
\mathcommand\Vsomesall{\Vsingleindex\indexgammadeltaplus}

\mathapplycommand\VARall      {\VARsingleindex\indexdelta         }
\mathapplycommand\VARwall     {\VARsingleindex\indexdeltaminu     }
\mathapplycommand\VARsall     {\VARsingleindex\indexdeltaplus     }
\mathapplycommand\VARgsome    {\VARsingleindex\indexgammaplus     }
\mathapplycommand\VARsome     {\VARsingleindex\indexgamma         }
\mathapplycommand\VARfree     {\VARsingleindex\indexfree          }
\mathapplycommand\VARbound    {\VARsingleindex\indexbound         }
\mathapplycommand\VARsomesall {\VARsingleindex\indexgammadeltaplus}
\mathcommand\displayVARsall[1]{\VARsingleindex\indexdeltaplus
\!\!\!\:\left(\begin{array}{@{}c@{}}#1\end{array}\right)}

\mathcommand\rigidvari     [2]{#1_{#2}^\indexgammadeltaplus}
\mathcommand\existsvari    [2]{#1_{#2}^\indexgamma    }
\mathcommand\forallvari    [2]{#1_{#2}^\indexdelta    }
\mathcommand\freevari      [2]{#1_{#2}^\indexfree     }
\mathcommand\wforallvari   [2]{#1_{#2}^\indexdeltaminu}
\mathcommand\sforallvari   [2]{#1_{#2}^\indexdeltaplus}
\mathcommand\gexistsvari   [2]{#1_{#2}^\indexgammaplus}
\mathcommand\boundvari     [2]{#1_{#2}}
\mathcommand\vari          [2]{#1_{#2}}
\mathcommand\wforallvarilow[2]{#1_{#2}^
{\raisebox{-.82ex}{\math\indexdeltaminu}}}

\newcommand\indexhelper[1]{{\scriptscriptstyle#1\:\!\!}}
\newcommand\indexdeltaplus
{\indexhelper{\delta^{\raisebox{-.17ex}{\fvesf\hskip-0.14em +}}}}
\newcommand\indexdeltaminu
{\indexhelper{\delta^{\mbox{\fvesf\hskip-0.14em\rule[.2ex]{.7em}{.15ex}}}}}
\newcommand\indexgammaplus
{\indexhelper{\gamma^{\mbox{\fvesf\hskip-0.14em +}}}}
\newcommand\indexgammadeltaplus
{\indexhelper{\gamma\delta^{\raisebox{-.17ex}{\fvesf\hskip-0.14em +}}}}

\newcommand\indexdelta{\indexhelper\delta}
\newcommand\indexgamma{\indexhelper\gamma}
\newcommand\indexfree
{{\scriptscriptstyle\free}}
\newcommand\indexbound
{{\scriptscriptstyle\bound}}

\newcommand\Wellfsymb{\ident{Wellf}}
\mathapplycommand\Wellfpp{\Wellfsymb}
%%%%%%%%%%%%%%%%%%%%%%%%%%%%%%%%%%%%%%%%%%%%%%%%%%%%%%%%%%%%%%%%%%%%%%%%%%%%%%%
%%OBSOLETE STUFF
%%variables
%\mathcommand\x {x\:}\mathcommand\y {y\:}\mathcommand\z {z\:}
%\mathcommand\el{l\:}\mathcommand\k {k\:}\mathcommand\be{b\:}
%\mathcommand\h {h\:}
%%constants
%\mathcommand\zero{\ident{0}\:}\mathcommand\nil{\ident{nil}\:}
%\mathcommand\strangelist{\mbox{\ident{strange--list}}\:}
%\mathcommand\true {\ident{true}\:}\mathcommand\false{\ident{false}\:}
%%functions
%\mathcommand\f  {\ident{f}\:}\mathcommand\s  {\ident{s}\:}
%\mathcommand\cns{\ident{cons}\:}\mathcommand\car{\ident{car}\:}
%\mathcommand\cdr{\ident{cdr}\:}\mathcommand\identminus{\minussymbol\:}
%\mathcommand\mbp{\mbpsymbol\:}\mathcommand\plus{\plussymbol\:}
%\mathcommand\dl {\ident{dl} \:}\mathcommand\ssp{\ident{ssp}\:}
%%literals
%\mathcommand\eqcolon{\boldequal\:}\mathcommand\necolon{\boldunequal\:}

%% file: headersugarterms.tex
\mathcommand\beginargs{(}
\mathcommand\allargs  {)}
\mathcommand\separgs  {,\,}
\mathcommand\tightsepargs{,}

\mathcommand\minusppnoparentheses  [2]{{#1}\,\minussymbol\,{#2}}
\mathcommand\tightminusppnoparentheses  [2]{{#1}\minussymbol{#2}}
\mathcommand\divideppnoparentheses [2]{{#1}\,\dividesymbol\,{#2}}
\mathcommand\plusppnoparentheses   [2]{{#1}\,\plussymbol \,{#2}}
\mathcommand\plusppnoparenthesesi  [3]{{#2}\,\plussymbol^{#1}\,{#3}}
\mathcommand\tightplusppnoparentheses   [2]{{#1}\plussymbol{#2}}
\mathcommand\timesppnoparentheses  [2]{{#1}\,\timessymbol\,{#2}}
\mathcommand\undppnoparentheses    [2]{{#1}\und            {#2}}
\mathcommand\oderppnoparentheses   [2]{{#1}\oder           {#2}}
\mathcommand\impliesppnoparentheses[2]{{#1}\implies        {#2}}
\mathcommand\leqinfixppnoparentheses[2]{{#1}\,\tight\leq\,{#2}}
\mathcommand\geqinfixppnoparentheses[2]{{#1}\,\tight\geq\,{#2}}
\mathcommand\dividepp [2]{(\divideppnoparentheses {#1}{#2})}
\mathcommand\minuspp  [2]{(\minusppnoparentheses  {#1}{#2})}
\mathcommand\pluspp   [2]{(\plusppnoparentheses   {#1}{#2})}
\mathcommand\timespp  [2]{(\timesppnoparentheses  {#1}{#2})}
\mathcommand\undpp    [2]{(\undppnoparentheses    {#1}{#2})}
\mathcommand\oderpp   [2]{(\oderppnoparentheses   {#1}{#2})}
\mathcommand\impliespp[2]{(\impliesppnoparentheses{#1}{#2})}

%% file: headerframe.tex
\mathcommand\synconseins{S}
\mathcommand\synconszwei{S'}
%\mathcommand\infoeins{\pi}
%\mathcommand\infozwei{\pi'}
%\mathcommand\synconsinfoeins{(\synconseins,\infoeins)}
%\mathcommand\synconsinfozwei{(\synconszwei,\infozwei)}
\mathcommand\inductionorderinglesssim{\lesssim_{\rm ind}}
\mathcommand\inductionorderinggtrsim {\gtrsim}
\mathcommand\inductionorderingless   {<}
\mathcommand\inductionorderinggtr    {>}
\newcommand\FAIL{{\rm FAIL}}
% foundedness
\mathcommand\strictlyfounded{\searrow}
\mathcommand\antistrictlyfounded{\swarrow}
\mathcommand\notstrictlyfounded{\not\searrow}
\mathcommand\notantistrictlyfounded{\hskip.3em\setminus\hskip-1.1em\swarrow}
\mathcommand\founded{\curvearrowright}
\mathcommand\antifounded{\curvearrowleft}
\mathcommand\strictquasifoundedindex[1]
{\,\strictlyfounded\mbox{\math/}\!\founded_{#1}\,}
\newcommand\strictquasifounded{\strictquasifoundedindex{}}

% GENERAL

\newcommand\stepsound{sound}
\newcommand\stepsoundness{soundness}
\newcommand\Stepsoundness{Soundness}
\newcommand\Inductiveness{Inductiveness}
\newcommand\inductiveness{inductiveness}
\newcommand\Inductive{Inductive}
\newcommand\inductive{inductive}
\newcommand\inductivelysound{inductively sound}
\newcommand\InductiveSoundness{Inductive Soundness}
\newcommand\Inductivesoundness{Inductive soundness}
\newcommand\inductivesoundness{inductive soundness}

\newcommand\K{{\rm K}} 

\newcommand\syncons{{\rm SynCons}}

% FORMULAS

\newcommand\abstractformulas{{\rm Form}}
\newcommand\abstractatoms{\ATSYM}

\mathcommand\Feins{\Gamma}
\mathcommand\Fzwei{\Delta}
\mathcommand\Fdrei{\Pi}  
\mathcommand\Fvier{\Lambda}
\mathcommand\Ffuen{\Theta}
\mathcommand\Fsech{\Xi}
\mathcommand\Feinsprime{\Feins'}
\mathcommand\Fzweiprime{\Fzwei'}
\mathcommand\Fdreiprime{\Fdrei'}
\mathcommand\Feinsprimeprime{\Feins''}
\mathcommand\Feinsprimeprimeprime{\Feins'''}

\mathcommand\Feinszwei{\Feins\!\Fzwei}

\newcommand\getformulaname{{\rm logic}}
\mathcommand\getformula[1]{\getformulasofset{\{#1\}}}
\mathapplycommand\getformulasofset{\getformulaname}

% INDUCTION ORDERING

\newcommand\hiddenIOsubs[1]{^{#1}}

\mathcommand\iql    [1]{\;\lesssim \hiddenIOsubs{#1}\,}
\mathcommand\iqb    [1]{\;\gtrsim  \hiddenIOsubs{#1}\,}
\mathcommand\il     [1]{\,<        \hiddenIOsubs{#1}\,}
\mathcommand\ileq   [1]{\,\leq     \hiddenIOsubs{#1}\,}
\mathcommand\ib     [1]{\,>\!\!    \hiddenIOsubs{#1}\,}
\mathcommand\notib  [1]{\,\ngtr\!\!\hiddenIOsubs{#1}\,}
\mathcommand\notil  [1]{\,\nless   \hiddenIOsubs{#1}\,}
\mathcommand\iapprox[1]{\,\approx  \hiddenIOsubs{#1}\,}

%% file: headerepsi.tex
\newcommand\sectionproandcontrareference{%
\myepsisection
{%\meinong\ \vs\ \russell: 
Introduction:
Pro and Contra Reference}\label
{section russell meinong}
\noindent
In this \sectref{section russell meinong}, 
we introduce to the 
description of the semantics of determiners in natural languages,
and show that the \math\varepsilon\ is useful for it.
We take the historical path by shedding some
light on the following two seminal philosophic papers,
which capture most opposite views pro and contra reference in natural languages:
\begin{description}\item[Pro:]\mediumheadroom
\cite{meinong}:
``{\fraknomath\Ue ber \Gegenstandstheorie}'' by \meinongname\ 
\mbox\meinonglifetime
\item[Contra:] \cite{denoting}:
``On Denoting'' by \russellname\ \mbox\russelllifetime
\mediumfootroom\end{description}
As our \math\varepsilon\ does not surrender to the present king of France,
it could help \russell\ to return to his original 
appreciation of \meinong's ideas and position, which he still expressed in
\cite{onmeinong}.

\myepsisubsection{I met a man.\hfill(R1)}\label{section i met a man}
\noindent
In \cite{denoting}, the affirmation (R1) is taken to be 
\par\halftop\noindent\LINEnomath
{`I met \nlbmath x and \math x is human' is not always false.}(R2)
\par\halftop\noindent 
(\russell's identification of being a man and being human is not relevant 
 to us here.)
From \cite[\litchapref{XV}]{mathematicalphilosophy}, 
it \nolinebreak becomes clear that (R2) means 
what today would be \mbox{stated as}\par\halftop\noindent\LINEmath{
  \exists x\stopq\inparentheses{\Metppp\Ipp x\und\Humanpp x}
}(W1)\par\halftop\noindent
In our free-variable framework, we can omit the \mbox
{\math\gamma-quantifier of (W1)}
and replace its bound \mbox{\math\gamma-variable \boundvari x{}} either with
a \fev\ \nlbmath{\existsvari  x 1} or with a 
\sfuv\ \nlbmath{\sforallvari x 0} constrained with a tautological \cc. \ 
This results in one of the logically equivalent forms of either
\par\halftop\noindent\LINEmath{
  \Metppp\Ipp{\existsvari  x 1}\und\Humanpp{\existsvari  x 1}
}(W2)\\\noindent or\\\noindent\LINEmaths{
  \Metppp\Ipp{\sforallvari x 0}\und\Humanpp{\sforallvari x 0}}{}(W3)
  \\with \cc\\\noindent
  \LINEmaths{\app C{\sforallvari x 0}
  \nottight{\nottight{:=}}
  \truepp}{}(W4)\par\halftop\noindent
Note that (W3)+(W4) is logically equivalent to each of (W1) and (W2)
because we have chosen an implicit existential 
quantification for our \sfuv s. \ 
This could be changed to universal or generalized quantification for the design 
of operators for descriptive terms similar to our novel 
\math\varepsilon-operator. \ 
Our preferred reading of (R1), however, is the following:
\par\halftop\noindent\LINEmath{
  \Metppp\Ipp{\sforallvari x 0}
}(W5)\\\noindent with \cc\\\noindent\LINEmaths{
  \app C{\sforallvari x 0}
  \nottight{\nottight{:=}}
  \Metppp\Ipp{\sforallvari x 0}\und\Humanpp{\sforallvari x 0}
}{}(W6)\par\halftop\noindent 
Only if there is an emphasis on the conviction that it was a human indeed
whom I met,
\ie\ that the \cc\ (W6) denotes, we would model (R1) as (W3)+(W6).

The logical equivalence of (W3)+(W6) with (W1) is nothing but a version of
\hilbert's original axiom \nlbmath{\inpit{\varepsilon_1}} 
for elimination the existential quantifier
in our free-variable framework, where \sforallvari x 0 replaces the
\nlbmath\varepsilon-term \bigmaths
{\varepsilon x\stopq\inparentheses{\Metppp\Ipp x\und\Humanpp x}}.

(W5)+(W6), however, is a proper logical consequence of (W1) in general: \ 
If (W1) is false, (W5)+(W6) \nolinebreak still would be true 
if I were Redcap and met the wolf disguised as a human. \ 
A listener knowing for certain that there are no men in the forest
may make sense out of what \redcap\ tells him
by assuming that the man she met denotes a specific object, 
and then find out that it must be the wolf. \ 
Then (W5)+(W6) is true; to wit, pick the wolf for \nlbmath{\sforallvari x 0}. \ 
This is so because we may choose any object if the \cc\ does not denote,
\ie\ if it is unsatisfiable in the context under consideration.
%%%%%%%%%%%%%%%%%%%%%%%%%%%%%%%%%%%%%%%%%%%%%%%%%%%%%%%%%%%%%%%%%%%%%%%%%%%%%%%%
\myepsisubsection{Scott is the author of Waverley.\hfill(SW1)}\label
{section Scott}
\noindent
(SW1) is another famous example from 
\cite[\litchapref{XVI}]{mathematicalphilosophy}\@. \  
We \nolinebreak model it as\par\halftop\noindent\LINEnomath
{\mbox{Scott} = \sforallvari z{}}(SW2)\\with \cc\\\LINEmath 
{\app C{\sforallvari z{}}\nottight{\nottight{:=}}
\mbox{``\sforallvari z{} is author of Waverley''}}(SW3)\par\halftop\noindent
For the \cc\ \nlbmath C of \nolinebreak(SW3), \
our version \nlbmath{\app{Q_C}{\sforallvari z{}}}
of \hilbert's original axiom \nlbmath{\inpit{\varepsilon_0}} reads \\\LINEmaths{
\exists z\stopq \mbox{``\boundvari z{} is author of Waverley''}
\nottight{\implies}
\mbox{``\sforallvari z{} is author of Waverley''}}{}\par\halftop\noindent 
For global application of the substitution 
\maths{\sigma:=\{\sforallvari z{}\mapsto\mbox{Scott}\}},
we have to show \nlbmaths{\inpit{\app{Q_C}{\sforallvari z{}}}\sigma}. \ 
This is valid, provided that Scott\emph{is} author of Waverley.
Thus, by \reductiontheorem{6a}, we can infer the validity of \nolinebreak 
(SW2)+(SW3)
% \ \mbox{``Scott = \sforallvari z{}''} \ 
from the validity of the \math\sigma-instance of \nolinebreak(SW2), 
which is the tautology \ \mbox{``Scott = Scott}\closequotefullstopextraspace
This is in blank opposition to the following statement 
(\cite{denoting}):\begin{quote}
``The proposition `Scott was the author of Waverley'\,'' \ldots\
``does not contain any constituent 
`the \nolinebreak author of Waverley' for which we
could substitute `Scott'.''
\noitem\end{quote}
%%%%%%%%%%%%%%%%%%%%%%%%%%%%%%%%%%%%%%%%%%%%%%%%%%%%%%%%%%%%%%%%%%%%%%%%%%%%%%%%
One could try to defend \russell's statement by arguing that he may only
have seen his reduced syntactical form \par\halftop\noindent\LINEmaths
{\exists z\stopq\inparentheses{\forall y\stopq\inparentheses{\inpit{y\boldequal 
z}\equivalent\mbox{``\math y is the author of Waverley''}}\und\mbox{Scott}=z}
},(SW4)\par\halftop\noindent but---after a deeper contemplation on 
\nolinebreak (SW4) and the
concept of reference in general, and after a closer look at 
\cite{denoting} and 
\cite{mathematicalphilosophy}---\russell's statement seems to be
just an outcome of \russell's strange philosophy of sometimes 
ignoring \kant's distinction on\emph{a posteriori}
and\emph{a priori}, \frege's notion of\emph{sense}, 
and, in general, the\emph{functions of syntax and 
logical calculi, mixing them up with semantics}. \ 
That \russell\ did so 
 becomes more obvious from the following simpler statement
 \cite[\p 175]{mathematicalphilosophy}:\par\halftop\noindent\LINEnomath
{`` `Scott is Sir Walter' is the same 
 trivial proposition as `Scott is Scott'.''}\par\halftop\noindent
% Note that---even among the best---this ignorance  was not uncommon 
% at that time. (Search \cite{tractatus} for further examples!)
Beyond that, \cite{denoting} argues\emph{contra} reference in general,
but in favor of encoding reference into mere predicate logic as
in \nolinebreak(SW4). \ 
Advantages of \nolinebreak(SW2)+(SW3) over \nolinebreak(SW4), however,
are its elegance and simplicity as well as the introduction of the formal 
reference object \nlbmath{\sforallvari z{}}, 
which is non-trivially constrained by \nolinebreak(SW3) and
may be reused for further reference\@.

\myepsisubsection
{Ignoring the Definite, Generic, Qualitative, Salient, Specific, \etc}\label
{section proviso}
\noindent
As a most interesting standard example we will consider 
``a/the round quadrangle is quadrangular'' 
in \nolinebreak\sectref{section quadrangle is quadrangular new short}. \ 
We do not want to emphasize
``\tightemph{the} round quadrangle'' 
%(or ``\tightemph{the} \nolinebreak present king of France'')
as \nolinebreak
we want to circumvent the extra complication introduced by the definite
syntactical form, which may or may not indicate properties 
such as uniqueness, specifity, or salience. \ 
We \nolinebreak also do not want to emphasize 
``\tightemph{a} round quadrangle is quadrangular''
as we are not interested in the generic reading
\par\noindent\LINEmaths
{\forall x\stopq\inparentheses{
 {\Roundppp x\und\Quadrangularppp x}
 {\nottight{\nottight\implies}}
 \Quadrangularppp x}}.\par\noindent
To be precise, we have to specify that we are interested in\emph{reference}
(\ie\ not in\emph{predication}) and that the
usage of indefinite determiners (articles and pronouns)
we intend to mirror with the \math\varepsilon\ here
is\emph{particular} (\ie\ not\emph{generic}) and\emph{referential} 
(\ie\ not\emph{qualitative}).
Generic and qualitative usage of indefinite articles 
is\emph{merely quantificational} and refers to 
a property (predication) and not to an object having it (reference).
For example,
in the sentence 
{``An \nolinebreak elephant is a huge animal.''}
the ``An'' is generic and the ``a'' is
qualitative. It simply says \bigmaths{
   \forall x\stopq\inparentheses{
       \Elephantppp x
     {\nottight{\nottight\implies}}
     {\Hugeppp x\und\Animalppp x}}
}.
Moreover, our modeling with the \math\varepsilon\ does not 
presuppose that the 
description\emph{designates} or that it is\emph{salient} or\emph{specific}.
Note that the distinctions
% of indefinite descriptions 
on salience and specifity depend on a discourse and 
the referential status for the speaker, \resp, 
which we do not take into account here.

\myepsisubsection{The round quadrangle is quadrangular.}\label
{section quadrangle is quadrangular new short}
\noindent
With the\emph{proviso} of \sectref{section proviso},
we now model ``a/the round quadrangle is quadrangular'' as 
\par\halftop\noindent\LINEmaths{\Quadrangularppp{\sforallvari y 1}
}{}(S)\\\noindent with \cc\\\noindent\LINEmaths{\mbox{}~~~~~~~~~~~~~~~
\app C{\sforallvari y 1}\nottight{\nottight{\nottight{:=}}}\mbox
{\Roundppp{\sforallvari y 1}\und\Quadrangularppp{\sforallvari y 1}}}{}(C1)
\par\halftop\noindent
The \cc\ \app C{\sforallvari y 1} of \nolinebreak(C1)
is equivalent to \nlbmaths{\falsepp}{}
if we have the sequent
\par\halftop\noindent\LINEmaths{\neg\Roundppp{\wforallvari u{}}\comma
\neg\Quadrangularppp{\wforallvari u{}}}{}(A)\par\halftop\noindent
available as a lemma. \ 
Then, we may choose any object for \sforallvari y 1. \ 
If we chose a quadrangular one, (S) \nolinebreak becomes true. \ 
Thus, the statement \nolinebreak (S)+(C1) is valid,
due to our choice of an implicit existential quantification for 
the \sfuv s. \ 
We cannot follow the critique of \cite{denoting} against 
\cite{meinong} here, namely that \sforallvari y \nolinebreak\ would be
``apt to infringe the Law of Contradiction\closequotefullstopextraspace
Firstly, \sforallvari y{} \nolinebreak is well-specified by \nolinebreak(C1)
and denoting a well-defined object. \ 
Secondly, 
we do not even follow \cite{denoting} 
insofar as undefined objects in a domain would
be in conflict with the Law of Contradiction. \ 
Indeed, for very good practical reasons, we find\begin{itemize}\noitem\item[(H)]
linguistically motivated models with 
arbitrary\emph{not really existing objects} in 
\cite{hobbs} and \cite[\litsectref{2.2}]{hobbs-magnum-opus}, \
(as part of the\emph{ontological promiscuity} 
 suggested for the\emph{logical notation},
 \cfnlb\ \cite[\litsectref 1, \litnoteref{14}]{hobbs-magnum-opus}), \ and 
\noitem\item[(W)]
logically motivated models with 
arbitrary\emph{undefined objects} in 
% \cite{wgcade}, 
\cite{kwspec},  
%\cite{wirthdiss}, 
and \cite{wirth-jsc}.\noitem\end{itemize}
The ``really existing'' or ``defined'' objects are simply those
for which a predicate ``\RExistssymbol'' or ``\Def\,'' holds. \ 
This treatment obeys both the Law of Contradiction and
the Law of the Excluded Middle. \ 
While in the approach of (H) we have to replace \nolinebreak(A) with 
\par\noindent\LINEmaths{\neg\Roundppp{\wforallvari u{}}\comma
\neg\Quadrangularppp{\wforallvari u{}}\comma
\neg\RExistspp{\wforallvari u{}}}{}\par\noindent
there is an additional possibility 
in the approach of \nolinebreak(W) where the standard variables
range over defined objects only: \ 
When the variable \nlbmath{\sforallvari y 1} in 
\nolinebreak(C1) is a\emph{general} variable, ranging over the defined as 
well as the undefined objects, 
and the variable \nlbmath{\wforallvari u{}} in \nolinebreak(A)
is a standard variable, ranging over the defined objects only,
then there are models of
\nolinebreak(A) with undefined objects that are both round and quadrangular. \ 
Note that this syntactical trick of \nolinebreak(W)
is not just syntactical sugar improving the readability of formulas
critically, but also cuts down logical inference by restricting notions
such as matching, unification, and rewriting.

\myepsisubsection{\englishtexteleventwo{The}.}\label{section certainly}
\noindent
\russell's critique on \cite{meinong} is justified, however, 
insofar as a single statement of \cite{meinong} 
seems to be in conflict with the Law of Contradiction, indeed:
Assuming \nolinebreak(A) from above,
there seems to be no way to model the \nth 2 line of the following 
sentence as true in two-valued logics 
\cite[\p\,8, modernized orthography]{meinong}:
\begin{quote}{\fraknomath``\germantexteleven''}\hfill (V)\hspace*{-\rightmargin}
\notop\end{quote}\begin{quote}
``\englishtextelevenone''
\getittotheright{(our translation)}
\end{quote}
Of course, one could consider a trivial generic reading:
\par\noindent\LINEmaths
{\forall x\stopq\inparentheses{\smallheadroom\smallfootroom
   {\Roundppp x\und\Quadrangularppp x}
   \nottight{\nottight{\nottight\implies}}
   {\Roundppp x\und\Quadrangularppp x}}}.
\par\noindent 
But this is most unlikely to be intended due to the definite forms
of the articles, especially due to the one in the \nth 1 line of 
\nolinebreak(V)\@. \ 
Our preferred reading of the \nth 2 line of \nolinebreak(V) is
\par\halftop\noindent\LINEmaths{
\Roundppp{\sforallvari y 1}\und\Quadrangularppp{\sforallvari y 1}
},(P)\par\halftop\noindent
referring to the \cc\ of (C1) above. \ 
Now (P) \nolinebreak simplifies to \nlbmath\falsepp\
under the assumption of \nlbmath(A), unless either 
\begin{itemize}\noitem\item we restrict 
\nlbmath{\wforallvari u{}} in \nolinebreak(A) to the defined, as sketched in 
\sectref{section quadrangle is quadrangular new short}
(for which there is no indication in \cite{meinong}, however), \ or else
\item we choose some para-consistent logics, where a contradiction does 
not imply triviality, \cf\ \cite{sep-logic-paraconsistent} \hskip.3em
(for which, however, there is no indication in \cite{meinong}, either). \ 
Contrary to its promising title ``\meinong's Theory of Objects and \hilbert's \math\varepsilon-symbol\closequotecommaextraspace
\cite{meinong-hilbert} 
does not contain relevant information 
(neither on \meinong's 
{\fraknomath\Gegenstandstheorie} nor on \hilbertsepsilon) \hskip.2em
besides sketching a para-consistent logic.
\end{itemize}
Alternatively, \cite{meinong} may be consistently understood as follows:

Note that \meinong's ``{\fraknomath\Gegenstand}'' is best translated
as ``object'' and his ``{\fraknomath\Gegenstandstheorie}'' as 
``Theory of Objects\closequotefullstopextraspace
Possible other translations would be 
``subject\closequotecomma
``concept reference\closequotecomma 
or 
``referent with \cc\closequotecomma
but neither ``thing'' nor ``objective\closequotefullstopextraspace
According to \cite[\p\,40\f]{meinong}, {\fraknomath\Gegenstandstheorie}
is the most general aprioristic science, 
whereas metaphysics is the most general aposterioristic science.
Thus, {\fraknomath\Gegenstandstheorie} is not less general than
metaphysics, and we should carefully exclude from
\meinong's notion of a {\fraknomath\Gegenstand}
any connotation of being physical or realizable.
There is
no explicit definition of {\fraknomath\Gegenstand} in \cite{meinong},
but only a parenthetical indication:

%\samepage{

\begin{quote}{\fraknomath\ldots\ ``\germantextthirteen'' \ldots}\getittotheright
{\cite[\p\,2]{meinong}}\nopagebreak\par\halftop\noindent\ldots\ 
``\englishtextthirteen'' \ldots\getittotheright
{(our translation)}\\\getittotheright{(``pointing to'' may
 be replaced with ``aiming \nolinebreak at'')}
\end{quote}

%}%samepage

\noindent
Having clarified the notion of {\fraknomath\Gegenstand} a little, 
let us come back to the \nth 2 line of \nolinebreak (V)\@. \ 
On the one hand, as stated already above, 
our preferred reading \nolinebreak (P) 
contradicts the above sequent \nolinebreak (A)\@. \ 
On the other hand, the reading
\ \mbox{``\math{\Roundppp{\sforallvari y 1}\equivalent
\Quadrangularppp{\sforallvari y 1}}''} \ 
implies 
\ \mbox{``\math{\neg\Roundppp{\sforallvari y 1}\und
\neg\Quadrangularppp{\sforallvari y 1}}''} \ 
by \nolinebreak (A)\@. \ This reading, however, is
very unlikely to be intended by somebody whose German is as excellent as 
\meinong's. \ 
\mbox{I assume}
that \meinong\ wanted to say that a specification for reference---such
as \app C{\sforallvari y 1} in (C1) above---is 
meaningful and should denote, no matter
whether there is an object that satisfies it. \ Indeed, we read:
\begin{quote}{\fraknomath\ldots\ ``\germantexttwelve''}\getittotheright
{\cite[\p 13]{meinong}}\notop\end{quote}\begin{quote}
\ldots\ ``what is not contingent to a {\fraknomath\Gegenstand} 
but establishes its proper
character constitutes its suchness, which sticks to the 
{\fraknomath\Gegenstand}, 
may it be or not be.'' \getittotheright{(our translation)}
\notop\end{quote}
In this light, might the \nth 2 line of \nolinebreak(V) 
even be boldly read as the\emph{valid} statement
\par\halftop\noindent\LINEmaths{
\Roundppp{\sforallvari y 1}\und\Quadrangularppp{\sforallvari y 2}
},(P\math')
\\with (C1) and\\
\LINEmaths{\app C{\sforallvari y 2}
\nottight{\nottight{\nottight{:=}}}\mbox
{\Roundppp{\sforallvari y 2}\und\Quadrangularppp{\sforallvari y 2}}}{}(C2)
%%%%%%%%%%%%%%%%%%%%%%%%%%%%%%%%%%%%%%%%%%%%%%%%%%%%%%%%%%%%%%%%%%%%%%%%%%%%%%%
\myepsisubsection{Conclusion}
\noindent
All in all, we may conclude that the
\math\varepsilon\,---and especially our novel treatment of it---is 
useful for describing the semantics of determiners in natural languages:
We can formalize some of \meinong's ideas on philosophy of language 
and contribute to the
defense of his points of view against \russell's critique,
even on empty descriptions.}
%%%%%%%%%%%%%%%%%%%%%%%%%%%%%%%%%%%%%%%%%%%%%%%%%%%%%%%%%%%%%%%%%%%%%%%%%%%%%%%%

%% file: quotation.tex
\newcommand\germantexthilberteinsneunnullnull
{Exi\esi tenz de\es\ Inbegriff\es\ aller reellen Zahlen und unendlicher Mengen \ue berhaupt}
\newcommand\germantextone
{ist ein Ding de\es\ In\-di\-vi\-du\-\ewithtrema n\-be\-reich\es, 
 und zwar ist diese\es\ Ding gem\ae\sz\ der inhaltlichen \Ue ber\-setzung der 
 Formel (\nlbmath{\varepsilon_0})
  \ \ \ \mbox{ei n}
  \ \ \ \mbox{s o l ch e \es},
  \ \ \ \mbox{au f}
  \ \ \ \mbox{d a \es} 
  \ \ \ \mbox{j e n e \es}
  \ \ \ \mbox{P\ r\ \ae\ d i k a t}~~~\math A 
  \ \ \ \mbox{z u t r i ff t}, 
  \ \ \ \mbox{v o r au \es\ g e s e tz t},
  \ \ \ \mbox{d a \ses}
  \ \ \ \mbox{e \es}
  \ \ \ \mbox{\ue\ b e r h au p t}
  \ \ \ \mbox{au f} 
  \ \ \ \mbox{ei n}
  \ \ \ \mbox{D i n g}
  \ \ \ \mbox{d e \es}
  \ \ \ \mbox{I n d i v i d u \ewithtrema\ n b e r ei ch \es}
  \ \ \ \mbox{z u t r i ff t}.}
\newcommand\germantexttwo
{Ich suche
  \ \ \ \mbox{ei n}
  \ \ \ \mbox{B u ch},
  \ \ \ \mbox{d a \es}
  \ \ \ \mbox{i ch}
  \ \ \ \mbox{g e s t e r n}
  \ \ \ \mbox{b e k o m m e n}
  \ \ \ \mbox{h a b e};
  \ \ \ e\es\ ist ein sch\oe\-ne\es.}
\newcommand\germantextthree
{Dar\ue ber hinau\es\ hat da\es\ \math\varepsilon\ 
 die Rolle der Au\esi wahl\-funk\-tion,
 \dasheisst\ im Falle, wo \bigmaths{A\,a}{} 
 auf mehrere Dinge zutreffen kann,
 ist \bigmaths{\varepsilon\,A}{}
  \ \ \ \mbox{i r g e n d ei n e \es}
  \ \ \ von den Dingen \nlbmath a, auf welche \bigmath{A\,a} zutrifft.}
\newcommand\germantextfour
{Da\es\ \math\varepsilon-Symbol bildet somit eine Art der Verallgemeinerung
 de\es\ \math\mu-Symbol\es\ f\ue r einen beliebigen 
 In\-di\-vi\-du\-\ewithtrema n\-be\-reich.
 Der Form nach stellt e\es\ 
 eine Funktion eine\es\ variablen Pr\ae\-di\-ka\-te\es\ dar,
 welche\es\ au\sz er demjenigen Argument, auf welche\es\ sich die zu dem 
 \math\varepsilon-Symbol ge\-h\oe\-ri\-ge gebundene Variable bezieht, noch freie
 Variable al\es\ Argumente (``Parameter'') enthalten kann.
 Der Wert dieser Funktion f\ue r ein be\-stimm\-te\es\ Pr\ae\-di\-kat \nlbmath A
 (bei Festlegung der Parameter) \germantextone}
\newcommand\germantextfive
{untergeordnete \mbox{\math\varepsilon-Au\esi dr\ue cke}}
\newcommand\germantextsix
{Die Amme hatte die Schuld. -- }
\newcommand\germantextseven
{Der kleine Herr Friedemann}
\newcommand\germantexteight
{Aller\-ding\es\ ist dieser Begriff von Au\esi wahl\-funk\-tion so kompliziert,
 da\ses\ sich seine Verwendung in der inhaltlichen Mathematik kaum empfiehlt.}
\newcommand\germantextnine
{Angesicht\es\ der Kompliziertheit de\es\ Be\-griff\es\ 
 der Au\esi wahl\-funk\-tion dritter Art
 ergibt sich die Frage, ob bei \hilbert-\bernays\ ('' \ldots\ ``) 
 wirklich beabsichtigt war, diesen Begriff von 
 Au\esi wahl\-funk\-tion axiomatisch zu beschreiben. Au\es\ der Darstellung bei 
 \hilbert-\bernays\ glaube ich entnehmen zu k\oe nnen, da\ses\ da\es\ nicht
 der Fall ist,}
\newcommand\germantextten
{Ich suche (irgend)
  \ \ \ \mbox{ei n}
  \ \ \ \mbox{B u ch};
  \ \ \ e\es\ soll ein sch\oe\-ne\es\ sein.}
\newcommand\germantexteleven
{Nicht nur der vielberufene goldene Berg ist von Gold, sondern auch\\
 da\es\ runde Viereck ist so gewi\ses\ rund al\es\ e\es\ viereckig ist.}
\newcommand\germantexttwelve
{wa\es\ dem \Gegenstand e in keiner Weise \ae u\sz er\-lich ist, vielmehr
 sein eigent\-liche\es\ Wesen au\esi macht, 
 in seinem Sosein besteht, da\es\ dem \Gegenstand\
 anhaftet, mag er sein oder nicht sein.}
\newcommand\germantextthirteen
{die Bezugnahme, ja da\es\ au\esi dr\ue ck\-liche Gerichtetsein auf jene\es\ 
 \glq etwa\es\grqcomma oder wie man ja ganz ungezwungen sagt, 
 auf einen \Gegenstand}
\newcommand\germantexttwenty
{``Wenn e\es\ eine Aufgabe der Philosophie ist,
 die Herrschaft de\es\ Worte\es\ \ue ber den menschlichen Geist zu brechen,
 indem sie die T\ae u\-schungen aufdeckt, die durch den Sprachgebrauch
 \ue ber die Beziehungen der Begriffe oft fast unvermeidlich entstehen,
 indem sie den Gedanken von demjenigen befreit, womit ihn allein
 die Beschaffenheit de\es\ sprachlichen Au\esi druck\esi mittel\es\ behaftet,
 so wird meine \Begriffsschrift, f\ue r diese Zwecke weiter au\esi ge\-bil\-det,
 den Philosophen ein brauch\-bare\es\ Werkzeug werden k\oe nnen.
 Freilich 
 %giebt 
 gibt auch \nolinebreak sie, 
 wie e\es\ bei einem \ae u\sz ern Dar\-stellung\esi mittel
 wohl nicht ander\es\ m\oe g\-lich ist, 
 den Gedanken nicht rein \nolinebreak wieder;
 aber \nolinebreak einer\-seit\es\ kann man diese Abweichungen auf da\es\ 
 Unvermeidliche und Un\-sch\ae d\-liche be\-schr\ae n\-ken,
 anderer\-seit\es\ ist schon dadurch, da\ses\ sie ganz
 anderer Art sind al\es\ die der Sprache 
 % eigen\-th\ue m\-li\-chen, 
 eigen\-t\ue m\-li\-chen, 
 ein Schutz gegen
 eine einseitige Beeinflussung durch eine\es\ dieser 
 Au\esi druck\esi mittel gegeben.''}
\newcommand\germantextthirty
{Wenn ein Mann einen Groschen hat, wirft er ihn in die Parkuhr.}
\newcommand\germantextninehundredcore
{Wir haben daher in der Mathematik keinen Anla\ses, 
 Exi\esi tenz in einem grund\-s\ae tz\-lich anderen Sinne anzunehmen,
 al\es\ wir da\es\ Bestehen von gesetzlichen Beziehungen annehmen.}
\newcommand\germantextninehundred
{{\germanfont\germantextninehundredcore}}
\newcommand\germantextninehundredfootnote
{{\germanfontfootnote\germantextninehundredcore}}
\newcommand\germantextninehundredcitation
{\cite[\p\,23]{bernays-existenz}. \ With old-fashioned orthography also in \cite[\p\,104]{bernays-philosophie}}
\newcommand\germantextninehundredandonecore
{Der Gedanke einer solchen mathematischen Tats\ae chlichkeit
 bedeutet andererseit\es\ nicht ein Zur\ue ckkommen auf die Ansicht von
 einer selb\-st\ae ndigen Exi\esi tenz der mathematischen Gegenst\ae nde.
 E\es\ handelt sich dabei nicht um ein Dasein,
 sondern um beziehung\esi m\ae\sz ige, strukturelle Bindungen und um 
 da\es\ Hervorgehen (Induziert-werden) von ideellen Gegenst\ae nden au\es\ 
 anderen solchen Gegenst\ae nden.}
\newcommand\germantextninehundredandone
{{\germanfont\germantextninehundredandonecore}}
\newcommand\germantextninehundredandonefootnote
{{\germanfontfootnote\germantextninehundredandonecore}}
\newcommand\germantextninehundredandonecitation
{\cite[\p\,23]{bernays-existenz}. \ 
 Also in \citep[\p\,105]{bernays-philosophie}}
\newcommand\germantextninehundredandtwo
{Die \kant ische Meinung, 
 da\ses\ die Struktur unseres eigenen Erkennen\es\ f\ue r un\es\ a priori
 bestimmbar sein m\ue sse,
 beruht offensichtlich auf einer T\ae uschung.
 Die Struktur unserer geistigen Organisation ist ja f\ue r unser Bewusstsein
 gleicherma\sz en transzendent wie die Beschaffenheit der \ae u\sz eren Natur.}
\newcommand\germantextninehundredandtwocitation
{\citep[\p\,471, modernized orthography]{bernays-vertrautes}}
\newcommand\englishtextelevenone
{\englishtexteleventwo{Not only the notorious golden mountain is of gold, 
 but also\\the}.}
\newcommand\englishtexteleventwo[1]
{#1 round quadrangle is just as certainly round as it is quadrangular}
\newcommand\englishtextone
{is an object of the universe 
for which---according
to the semantical translation of the formula \nolinebreak 
(\math{\varepsilon_0})---{\em the predicate
 \math A holds, provided that \math A holds for any object of the universe
 at \nolinebreak all.}}
\newcommand\englishtexttwo[1]
{I am looking for {\em #1{a book} which I got yesterday}; 
 it is a beautiful one.}
\newcommand\englishtextfour
{Thus, the \math\varepsilon-symbol forms a kind of generalization of the
 \math\mu-symbol for arbitrary universes.
 Syntactically, it provides a function of a variable predicate,
 which---besides the argument 
 to which the variable bound by the \math\varepsilon-symbol refers---may 
 contain free variables as arguments (``parameters'').
 The value of this function for a given predicate \nlbmath A
 (for fixed values of the parameters)}
\newcommand\englishtextsix[1]
{#1{The nurse} bore the blame. --- }
\newcommand\englishtextten[1]
{I am looking for #1{{\em a}\/(n arbitrary) {\em book}};
 it is to be a beautiful one.}
\newcommand\englishtextthirteen
{the reference, indeed the explicit pointing to that 
 `something\closesinglequotecomma or---as one would very informally say---to
 a {\fraknomath\Gegenstand}}
\newcommand\germantextfifty
{Diese\es\ gelingt nach einer Methode, welche unabh\ae ngig voneinander
J. \herbrand\ und M. \presburger\ au\esi gebildet haben.
Diese Methode besteht darin, da\ses\ man Formeln, welche gebundene Variablen
enthalten, \glq Reduzierte\grq\ zuordnet, in denen keine gebundenen Variablen
mehr au\efi tre\-ten und welche im Sinne der inhaltlichen Deutung jenen Formeln
gleichwertig sind.}
\newcommand\germantextfiftyquotation{\cite
[\Vol\,I, \p\,233, note-mark omitted, orthography modern\-ized]{grundlagen}}
\newcommand\germantexttwohundred
{Jede Fluchtz\ae hlgleichung ist bereit\es\ in einem abz\ae hlbaren Denkbereich
 nicht mehr f\ue r beliebige Werte der Relativko\ewithtrema 
 ffizienten erf\ue llt.}
\newcommand\germantextskolemeins{{%\germanfont
 Er mu\ses\ also sozusagen einen Umweg \ue ber da\es\ 
 Nichtabz\ae hlbare machen.}}
\newcommand\germantexthilbertsicherheit{{\germanfont
Und wo sonst soll Sicherheit und Wahrheit zu finden sein,
wenn sogar da\es\ mathematische Denken versagt?}}
\newcommand\germantexthilbertignorabimus{{\germanfont
E\es\ bildet ja gerade einen Hauptreiz bei der 
Besch\ae ftigung mit einem mathematischen Problem, da\ses\
wir in un\es\ den steten Zuruf h\oe ren:
da ist ein Problem, suche die L\oe sung;
du kannst sie durch reine\es\ Denken finden;
denn in der Mathematik gibt e\es\ kein {\rm Ignorabimus}.}}
\newcommand\germantexthilbertcantorsparadise{{\germanfont 
Au\es\ dem Paradie\es, da\es\ un\es\ \cantor\ geschaffen,
soll un\es\ niemand ver\-treiben.}}
\newcommand\germantexthilbertcantorsparadisecitation{\cite[\p 170]{unendliche}}
\newcommand\germantexthilbertonkant{{\germanfont 
Schon \kant\ hat gelehrt -- und zwar bildet die\es\ einen
integrierenden Bestandteil seiner Lehre \nolinebreak--,
da\ses\ die Mathematik \ue ber einen unabh\ae ngig von aller Logik 
gesicherten Inhalt verf\ue gt und daher nie und nimmer allein durch Logik
begr\ue ndet werden kann,
we\esi halb auch die Bestrebungen von \frege\ und \dedekind\ scheitern
mu\sesi ten. \ 
Vielmehr ist al\es\ Vorbedingung f\ue r die Anwendung logischer Schl\ue sse
und f\ue r die Bet\ae tigung logischer Operationen schon etwa\es\ in der 
Vorstellung gegeben:
gewisse, au\sz er-logische konkrete 
Objekte, die anschaulich al\es\ unmittelbare\es\
Erlebni\es\ vor allem Denken da sind.}}
\newcommand\germantextkantsnotionofproblematisch
{Ich nenne einen Begriff problematisch, der
   keinen Widerspruch enth\ae lt, der auch al\es\ eine Begrenzung
   gegebener Begriffe mit anderen Erkenntnissen zusammenh\ae ngt,
   dessen objektive Realit\ae t aber auf keine Weise erkannt werden kann.}
\newcommand\germantextartineins
{Eine rationale Funktion \nlbmath{F(x_1,x_2,\cdots,x_n)} von 
 \math n \nolinebreak Ver\ae nderlichen hei\sz e definit,
 wenn sie f\ue r kein reelle\es\ Wertsy\esi tem der \nlbmath{x_i}
 negative Werte annimmt.}
\newcommand\germantextartinzwei
{Satz\,4: E\es\ sei \math R ein reeller Zahlk\oe rper, der sich nur auf eine
 Weise ordnen l\ae sst, wie zum Beispiel der K\oe rper der rationalen Zahlen,
 oder der aller reellen algebraischen Zahlen oder der aller reellen Zahlen.\\
 Dann ist jede rationale definite Funktion von \nlbmath{x_1,x_2,\cdots,x_n}
 mit Ko\ewithtrema 
 ffizienten au\es\ \nlbmath R\\Summe von Quadraten von rationalen
 Funktionen der \nlbmath{x_i} mit Ko\ewithtrema ffizienten au\es\ \nlbmath R.}
\newcommand\frenchtextone{%
Je fus longtemps sans pouvoir appliquer 
ma m\'ethode aux questions affirmatives, 
parce que le tour et le biais pour y venir est beaucoup plus 
malais\'e que celui dont je me sers aux n\'egatives. 
De sorte que, lorsqu'il me fallut d\'emontrer que\emph
{tout nombre premier, qui surpasse de l'unit\'e un multiple de 4, 
 est compos\'e de deux quarr\'es}, je me trouvai en belle peine.
Mais enfin une m\'editation diverses fois r\'eit\'er\'ee me donna
les lumi\`eres qui me manquoient, et les questions affirmatives pass\`erent 
par ma m\'ethode, \`a l'aide de quelques nouveaux principes qu'il
y fallut joindre par n\'ecessit\'e.}
\newcommand\frenchtextonecitation{\cite[\Vol\,II, \p\,432]{fermat-oeuvres}}
\newcommand\frenchtextninty{{\frenchfont 
Nous entendons par raisonnement intuitionniste, un raisonnement qui
 satisfait aux conditions suivantes:
 on n'y consid\`ere jamais qu'un nombre fini
 d\'etermin\'e d'objets et de fonctions;
 celles-ci sont bien d\'efinies, leur d\'efinition
 permettant de calculer leur valeur d'une mani\`ere univoque;
 on n'affirme
 jamais l'existence d'un objet sans donner le moyen de le construire;
 on n'y consid\`ere jamais l'en\-semble de tous les objets \nlbmath x d'une
 collection infinie;
 et quand on dit qu'un raisonnement (ou un th\'eor\`eme) est vrai pour tous ces
 \nlbmath x,
 cela signifie que pour chaque \nlbmath x pris en particulier, 
 il est possible de r\'ep\'eter le raisonnement g\'en\'eral en question
 qui ne doit \^etre consid\'er\'e que comme le prototype de ces raisonnements 
 particuliers.}}
\newcommand\frenchtextnintylocationoriginal
{\cite[\p\,3, \litnoteref 3]{herbrand-consistency-of-arithmetic}}
\newcommand\frenchtextnintylocationoriginalmodifierforreprint
{Without the comma after
  ``{\frenchfont intuitionniste}\closequotecomma
  but with an additional comma after
  ``{\frenchfont question}''}
\newcommand\frenchtextnintylocationreprint
{\cite[\p\,225, \litnoteref 3]{herbrand-ecrits-logiques}}
\newcommand\frenchtextonehundred{{\frenchfont
Nous pouvons dire que la d\'emonstration de {\em\loewenheim}\/
  \'etait suf\-fi\-sante
  en Math\'e\-matiques; 
  mais il nous a fallu, dans ce travail, 
  la rendre `m\'eta\-math\'e\-ma\-ti\-que'
  (voir l'introduction),
  pour qu'elle nous soit de quelque utilit\'e.}}
\newcommand\frenchtextonehundredsourcelocationoriginal
{\cite[\p 118]{herbrand-PhD}} 
\newcommand\frenchtextonehundredsourcelocationmodifierforreprint
{Without the emphasis on ``{\frenchfont\em\loewenheim}\,'' and with
 ``{\frenchfont math\'e\-matiques}'' instead of 
 ``{\frenchfont Math\'e\-matiques}'' and
 ``{\frenchfont l'Introduction)}''
 instead of ``{\frenchfont l'introduction),}''}
\newcommand\frenchtextonehundredsourcelocationreprint
{\cite[\p 144]{herbrand-ecrits-logiques}}
\newcommand\frenchtextonehundredandone{{\frenchfont
(On remarquera que cette d\'efinition
diff\`ere de la d\'efinition qu'on pourrait croire la plus naturelle seulement
par le fait que, quand le nombre \nlbmath p augmente, 
le nouveau champ \nlbmath{C'} et les nouvelles valeurs ne peuvent
pas forc\'ement \^etre consid\'er\'es comme un `prolongement' des anciens;
mais cependant, la connaissance de \nlbmath{C'} et des valeurs pour un nombre
\nlbmath p d\'etermin\'e, % dieses Komma steht nur im Pariser Original,
% nicht jedoch in Herbrand (1968)!
entra\^\i ne celle d'un champ et de valeurs 
convenant pour tout nombre inf\'erieur;
seul donc un `principe de choix' pourrait conduire \`a prendre un syst\`eme
de valeurs fixe dans un champ infini).}}
\newcommand\frenchtextonehundredandonesourcelocationoriginal
{\cite[\p 109]{herbrand-PhD}} 
\newcommand\frenchtextonehundredandonesourcelocationmodifierforreprint
{Without the comma after ``{\frenchfont d\'etermin\'e}'' also in:}
\newcommand\frenchtextonehundredandonesourcelocationreprint
{\cite[\p 135\f]{herbrand-ecrits-logiques}}
\newcommand\frenchtextonehundredandtwo{{\frenchfont
Il est absolument n\'ecessaire de
prendre de telles d\'efinitions, pour donner un sens pr\'ecis aux mots:
`vrai dans un champ infini\closesinglequotecomma
qui ont souvent \'et\'e employ\'es sans explication suffisante,
et pour justifier une proposition \`a laquelle on fait souvent allusion,
d\'emontr\'ee par {\em\loewenheim}\commanospace\footnotemark[1]
sans bien remarquer que cette proposition
n'a aucun sens pr\'ecis sans d\'efinition pr\'ealable, et que la d\'emonstration
de {\em\loewenheim} est au surplus totalement insuffisante pour notre but 
(voir 6.4).}}
\newcommand\frenchtextonehundredandtwosourcelocationoriginal
{\cite[\p 110]{herbrand-PhD}} 
\newcommand\frenchtextonehundredandtwosourcelocationmodifierforreprint
{Without the signs after ``{\frenchfont d\'efinitions}\closequotecomma
``{\frenchfont mots}\closequotecomma and
``{\frenchfont pr\'ealable}'' and the emphasis on 
``{\frenchfont\em\loewenheim}\,\closequotecomma
but with the correct ``{\frenchfont 6.2}''
instead of the misprint ``{\frenchfont 6.4}''}
\newcommand\frenchtextonehundredandtwosourcelocationreprint
{\cite[\p 136\f]{herbrand-ecrits-logiques}}
\newcommand\frenchtextonehundredeleven{{\frenchfont
Cette expression signifie: traduites en language ordinaire, 
consid\'er\'ees comme une propri\'et\'e des entiers, 
et non comme un pur symbole.}}

\newcommand\frenchtextonehundredtwelve{{\frenchfont
On pourra aussi introduire un nombre quelconque de 
\\fonctions \bigmathnlb{\fsymbol_i x_1 x_2\cdots x_{n_i}}{}
avec des hypoth\`eses telles que:
\begin{enumerate}\notop\item[a)]{\em
Elles ne contiennent pas de variables apparentes.}
\item[b)]\sloppy{\em
Consid\'er\'ees intuitionnistiquement\commanospace\footnotemark[5] 
elles permettent de faire effectivement le calcul de\ 
\math{\fsymbol_i x_1 x_2\cdots x_{n_i}},
pour tout syst\`eme particulier de nombres;
et l'on puisse d\'emontrer intuitionnistiquement que
l'on obtient un r\'esultat bien d\'etermin\'e.}
 \ {\bf (Groupe\,C.)}
\noitem\end{enumerate}}}

\newcommand\frenchtextonehundredandthirteenwithquotes{``{\frenchfont
\frenchtextonehundredtwelve
\noindent{\footnotesize\frenchfont
\footnotemark[5]\frenchtextonehundredeleven{\normalsize''}\noitem\par}}
%Das sollte aus Secundaerzitatgruenden hierstehen bleiben! CP
\getittotheright
{\frenchtextonehundredandthirteentwosourcelocationoriginal}}

\newcommand\frenchtextonehundredandthirteentwosourcelocationoriginal
{\citep[\p\,5]{herbrand-consistency-of-arithmetic}}
\newcommand
\frenchtextonehundredandthirteensourcelocationmodifierforreprint
{Without the colon after ``{\frenchfont que}'' and the comma after
 ``\math{\fsymbol_i x_1 x_2\cdots x_{n_i}}\closequotecomma
 with a semi-colon instead of a full-stop after 
``{\frenchfont apparentes}\closequotecomma
 and with ``{\frenchfont langage}'' instead of ``{\frenchfont language}''
 also in: \ \citep[\p\,226\f]{herbrand-ecrits-logiques}}
\newcommand\frenchtextthreehundred{{\frenchfont 
Un mauvais sort semble s'\^etre 
acharn\'e sur la logique en France.'\notop
\end{quote}\begin{quote}`A terrible curse
seemed to have fallen upon logic in France.}}
\newcommand\frenchtextfourhundred{{\frenchfont
\citet{loewenheim-1915} a publi\'e du r\'esultat \'enonc\'e dans ce paragraphe
  une d\'e\-mon\-stra\-tion dont nous avons montr\'e 
  les graves lacunes dans notre
  travail d\'ej\`a cit\'e (\litchapwithsectref 5{6.2}).}}
\newcommand\frenchtextfivehundredwithfrontandend[2]{
\begin{enumerate}\item[#1{\frenchfont\math{1^0}.}]{\frenchfont
Si une des \'egalit\'es \`a satisfaire \'egale une variable
restreinte \nlbmath x \`a un autre individu;
ou bien cet individu contient \nlbmath x,
et on ne peut y satisfaire;
ou bien il ne contient pas \nlbmath x;
cette \'egalit\'e sera alors une des \'egalit\'es normales cherch\'ees;
et on remplacera \nlbmath x par cette fonction dans les autres \'egalit\'es
\`a satisfaire.}
\item[\frenchfont\math{2^0}.]{\frenchfont
Si une des \'egalit\'es \`a satisfaire \'egale une variable g\'en\'erale
\`a un autre individu, qui ne soit pas une variable restreinte,
il est impossible d'y satisfaire.}
\item[\frenchfont\math{3^0}.]{\frenchfont
Si une des \'egalit\'es \`a satisfaire \'egale
\\\LINEnomath{
\math{f_1(\varphi_1,\varphi_2,\ldots,\varphi_n)} \`a
\math{f_2(\psi_1,\psi_2,\ldots,\psi_m)},}\\
ou bien les fonctions \'el\'ementaires \math{f_1} et \math{f_2} 
sont diff\'erentes, auquel cas il es impossible d'y satisfaire;
ou bien les fonctions \math{f_1} et \math{f_2} sont les m\^emes; 
auquel cas on remplace l'\'egalit\'e
par celles obtenues en \'egalant 
\math{\varphi_i} \`a \nlbmath{\psi_i}.}#2\end{enumerate}}
\newcommand\frenchtextfivehundredsource
{\cite[\p\,96\f]{herbrand-PhD}}
\newcommand\frenchtextfivehundredsourcemodifier
{Note that the reprint in \citep[\p124]{herbrand-ecrits-logiques}
has \mbox{``{\frenchfont associ\'ees}''} instead of 
``{\frenchfont normales}\closequotecomma
which is not an improvement,
and the English translation of \englishtextfivehundredsource\
is based on this contorted version.}
\newcommand\englishtextskolemeins
{Thus he must make a detour, so to speak, through the non-denumerable.}
\newcommand\englishtextfivehundredwithoutfrontandend
{If one of the equations to be satisfied equates a 
\math\gamma-variable \nlbmath{\existsvari x{}} to another term;
either this term contains \nlbmath{\existsvari x{}},
and then the equation cannot be satisfied;
or else the term does not contain \nlbmath{\existsvari x{}},
and then the equation will be one of the normal-form
equations we are looking for,
and we replace \nlbmath{\existsvari x{}} 
by the term in the other equations to be satisfied.
\item[2.] 
If one of the equations to be satisfied 
equates a \math\delta-variable to another term that is not a 
\math\gamma-variable, the equation cannot be satisfied.
\item[3.]
If one of the equations to be satisfied equates
\math{f_1(\varphi_1,\varphi_2,\ldots,\varphi_n)} to 
\math{f_2(\psi_1,\psi_2,\ldots,\psi_m)};
either the function symbols \math{f_1} and \math{f_2}
are different, and then the equation cannot be satisfied;
or they are the same, and then we replace this equation
with those that 
equate \math{\varphi_i} to \nlbmath{\psi_i}.}
\newcommand\englishtextfivehundredwithfrontandend[2]
{\begin{enumerate}
\item[#1\relax 1.] 
\englishtextfivehundredwithoutfrontandend
#2\end{enumerate}}
\newcommand\englishtextfivehundredsource
{\cite[\p148]{herbrand-logical-writings}}
\newcommand\englishtextfifty
{This is achieved by a method developed independently by 
 J. \herbrand\ and M. \presburger. \ 
 The method consists in associating `reduced forms' to formulas with
 bound variables, in which no bound variables occur anymore and 
 which are semantically equivalent to the original formulas.}
\newcommand\englishtextonehundred
{We could say that \loewenheim's proof was sufficient in mathematics;
    but, in the present work, 
    we had to make it `metamathematical' (see Introduction) 
    so that it would be of some use to us.}
\newcommand\englishtextonehundredsourcelocationwithoutpagenumber
{herbrand-logical-writings}
\newcommand\englishtextonehundredsourcelocationpagenumber
{\p 176}
\newcommand\englishtextonehundredandone{We observe that this definition
differs from the definition that would seem the most natural only in that, 
as the number \nlbmath p increases, 
the new domain \nlbmath{C'} and the new values need not be regarded as forming
an `extension' of the previous ones.
Clearly, if we know \nlbmath {C'} and the values for a given number \nlbmath p,
then for each smaller number we know a domain and values that answer to the
number;
but only a `principle of choice' could 
lead us to take a fixed system of values in an infinite domain.}
\newcommand\englishtextonehundredeleven
{This expression means: when they are translated into ordinary language,
considered as a property of integers and not as a mere symbol.}
\newcommand\englishtextonehundredandtwo{It is absolutely necessary to adopt
such definitions if we want to give a precise sense to the words
`true in an infinite domain\closesinglequotecomma
words that have frequently been used without sufficient explanation,
and also if we want to justify a proposition proved by \loewenheim,
a proposition to which many refer without clearly seeing that \loewenheim's
proof is totally inadequate for our purposes (see \nolinebreak 6.4) and that, 
indeed,
the proposition has no precise sense until such a definition has been given.}
\newcommand\englishtextonehundredandthirteen
{We can also introduce any number of functions 
 \bigmathnlb{\fppeinsindex i{x_1,x_2,\ldots,x_{n_i}}}{}
 together with some hypotheses such that 
 \begin{enumerate}\item[(a)]{\em
 The hypotheses contain no bound variables.}\item[(b)]\sloppy{\em
 Considered intuitionistically\commanospace\footnotemark[5] they make the 
% actual 
 effective computation of the 
 \math{\fppeinsindex i{x_1,x_2,\ldots,x_{n_i}}}
 possible for every given set of numbers, and it is 
 possible to prove intuitionistically that we obtain a 
% well-determined 
 well-defined 
 result.}
 \ {\bf (Group\,C.)}\end{enumerate}
 {\footnotesize\footnotemark[5]This expression means: 
  translated into ordinary language,
  considered as a property of integers and not as a mere symbol.}}
\newcommand\englishtextonehundredandthirteenquotation
% {translation by \heijenoort\ in: \citep[\p\,290\f]{herbrand-logical-writings}}
{our translation}
\newcommand\englishtexttwohundred
{If \math\phi\ denotes an unknown function and \math{\psi_1,\ldots,\psi_k}
are known functions, and if the \math\psi's and the \math\phi\ are 
substituted in one another in the most general fashions and certain
pairs of the resulting expressions are equated, then, if the resulting set of
functional equations has one and only one solution for \nlbmath\phi, \ \ 
\math\phi\ \nolinebreak is a recursive function.}
\newcommand\englishtextfourhundred
{\citet{loewenheim-1915} published a proof of the result stated in this section.
In \citep[\litchapwithsectref 5{6.2}]{herbrand-PhD}, we pointed out that there 
are serious gaps in his proof.}
\newcommand\englishtextgoedelonherbrand
{I have never met \herbrand.
His suggestion was made in a letter in 1931, 
and it was formulated {\em exactly}\/ as on \p\,26 of my lecture notes, that is,
without any reference to computability. \ 
However, since \herbrand\ was an intuitionist, 
this definition for him evidently meant that there is a
{\em constructive}\/ proof for the existence and uniqueness of \nlbmath\phi.}
\newcommand\propertyA{Property\,A} %%% Two macros for freeness of change
\newcommand\PropertyA{Property\,A} %%% in capitalization!
\newcommand\propertyB{Property\,B} 
\newcommand\propertyC{Property\,C} 
\newcommand\PropertyC{Property\,C} 
\newcommand\termsofdepth[2]{\app{{\mathcal T}_{#1}}{#2}}
\mathcommand\termsofdepthnovars[1]{{\mathcal T}_{#1}}
\newcommand\englishherbrandribettheorem
{``An odd prime $p$ is called irregular if the class number of the field ${\bf
    Q}(\mu_p)$ is divisible by $p$ ($\mu_p$ being, as usual, the group of $p$-th
  roots of unity). According to Kummer's criterion, $p$ is irregular if and only
  if there exists an even integer $k$ with $2\leq k \leq p-3$ such that $p$
  divides (the numerator of) $k$-th Bernoulli number $B_k$, given by the
  expansion
  $$
   \frac{t}{e^t-1}+\frac{t}{2}-1=\sum_{n\geq 2} \frac{B_n}{n!} t^n.
  $$
  The purpose of this paper is to strengthen Kummer's criterion.
  
  Let $A$ be the ideal class group of $\Q(\mu_p)$, and let $C$ be the
  $F_p$-vector space $A/A^p$. The Galois group $\mbox{Gal}(\overline\Q/\Q)$
  acts on $C$ through its quotient $\Delta = \mbox{Gal}(\Q(\mu_p)/\Q)$. Since
  all characters of $\Delta$ with values in $\overline{{\bf F }}^*_p$ are powers of the
  standard character
  $$
  \chi: \mbox{Gal}(\overline\Q/\Q) \rightarrow \Delta
  \tilde{\longrightarrow} {\bf F}^*_p
  $$
  giving the action of $\mbox{Gal}(\overline\Q/\Q)$ on $\mu_p$, the
  vector space $C$ has a canonical decomposition
  $$
  C = \bigoplus_{i\bmod(p-1)} C(\chi^i),
  $$
  where 
  $$
  C(\chi^i) = \{c\in C|\sigma c = \chi^i(\sigma)\, c \mbox{ for all }
  \sigma\in\Delta\}. 
  $$\yestop\par
  (1.1) {\bf Main Theorem.}
  \\Let $k$ be even, $2\leq k\leq p-3$.\\Then $p|B_k$ if
  and only if $C(\chi^{1-k}) \not= 0$

  In fact, the statement that $C(\chi^{1-k}) \not= 0$ implies $p|B_k$ is well known \citep[Th.\,3]{herbrand-1932}%
%\footnote{Reference modified according to our reference list.}
''\getittotheright{\citep[\littheoref{1.1}, \p\,151]{ribet76:_q}}}
%% \footnote:{Let $\Q(\zeta_p)$ be a cyclotomic extension of $
%%   \Q$, with $ p$ an odd prime, let $ A$ be the Sylow $ p$-subgroup of
%%   the ideal class group of $\Q(\zeta_p)$, and let $ G$ be the Galois
%%   group of this extension.  Note that the character group of $ G$, denoted $
%%   \hat{G}$, is given by $\displaystyle \hat{G}=\{\chi^i\mid0\leq i\leq p-2\}$.
%% For each $ \chi\in\hat{G}$, let $ \varepsilon_\chi$ denote the corresponding orthogonal idempotent of the group ring, and note that the $ p$-Sylow subgroup of the ideal class group is a $ \mathbb{Z}[G]$-module under the typical multiplication. Thus, using the orthogonal idempotents, we can decompose the module $ A$ via $ A=\sum_{i=0}^{p-2}A_{\omega^i}\equiv\sum_{i=0}^{p-2}A_i$.
%% Last, let $ B_k$ denote the $ k$th Bernoulli number.
%% {\bf Theorem  (\herbrandribet)}   Let $ i$ be odd with $ 3\leq i\leq p-2$. Then $ A_i\neq 0 \Leftrightarrow p\mid B_{p-i}$.
%% The left-to-right direction of this theorem was proved by \herbrand\ himself. 
%% The converse is much more intricate, and was proved by \ribetname.}\footnote

\newcommand\englishtexthilbertonkant{\kant\ already taught ---~and indeed it 
is part and parcel of his doctrine~--- that mathematics has 
at its disposal a content secured independently
of all logic and hence can never be provided with a foundation by means of
logic alone;
that is why the efforts of \frege\ and \dedekind\ were bound to fail. \ 
Rather, as a condition for the use of logical inferences
and the performance of logical operations, 
something must already be given to our faculty of representation,
certain extra-logical concrete objects that are intuitively present as 
immediate conceptions prior to all thought.}

\newcommand\englishtextninehundred
% {Thus, in mathematics we have no reason to assume existence to have a 
%  fundamentally different meaning other than the validity
%  of relations.}
{Thus, in mathematics, we have no reasons to assume any meaning of
 `existence' that would be fundamentally different from that of 
 `the validity of axiomatic relations\closesinglequotefullstopnospace}
\newcommand\englishtextninehundredandonepartone
{The thought of such a mathematical \englishtatsaechlichkeit, however,
 does not mean a return to the opinion that mathematical objects
 have an independent existence}
\newcommand\englishtextninehundredandoneintermission
{for instance, as ideal things in themselves in the sense of \cite{KdrV},
 independent of an observer}
\newcommand\englishtextninehundredandoneparttwo
{This existence is not a there-being,
 but consists of structural relational connections and the 
 % generation (induction) 
 (inductive) generation 
 of intellectual objects from other intellectual objects.}

\newcommand\englishquoteforstertext
{Early twentieth century mathematicians used the expression
 `The Crisis in Foundations\closesinglequotefullstop
 This crisis had many causes and 
 ---~despite the disappearance of the expression from contemporary speech~---
 has never really been resolved.
 One of its many causes was the increasing formalisation of mathematics,
 which brought with it the realisation that the paradox of the liar could
 infect even mathematics itself.
 This appears most simply in the form of \russell's paradox,
 appropriately in the heart of set theory.}
\newcommand\englishquoteforstertextplace{\cite[\p\,838]{forster-years-on}}

\newcommand\englishquotefefermanonkreiselnciwithquotes
{``While the nature of the n.c.i.\ and results for it are clear enough,
 the proofs are quite another matter.
 \kreisel's own proof of the n.c.i.\ for arithmetic requires
 familiarity with the ins and outs of \hilbert's \math\varepsilon-calculus
 and with its application in \cite{ackermann-consistency-of-arithmetic}.
 For those few brought up on \cite{grundlagen-first-edition-volume-two}
 (as, \eg, \kreisel\ himself)
 this would not be an obstacle.
 But for the rest of us,
 the proof in \makeaciteoftwo{kreisel-1951}{kreisel-1952} 
 does not invite a detailed study;
 it just looks like a thicket.
 I myself have never tried to wade through it, and don't know
 anyone who has.\footnotemark[8]
 \par\noindent{\footnotesize\footnotemark[8]%
 I asked my colleague, \mintsname, an expert in \hilbert's
 \math\varepsilon-substitution method, whether he had ever studied
 \kreisel's proof.
 He said that he had not because by the time he learned of the 
 significance of the results,
 there were more understandable proofs available \opt{see below}.
 After writing the above, 
 I learned from \parsonsname\ 
 that back in the 50's he had managed to work his way through
 the proof in  \makeaciteoftwo{kreisel-1951}{kreisel-1952},
 though only by relying on \cite{ackermann-consistency-of-arithmetic}
 to help to fill in the details.{\normalsize''}\par}}
\newcommand\englishquotefefermanonkreiselncitextplace{\cite[\p\,254]{unwinding}}

%% file: namedhelper.tex
\def\citep{\cite}
\def\citet#1{\citeauthor{#1} \shortcite{#1}}
\newcommand\startcite{{\raise.2ex\hbox{[}}}
\newcommand\stopcite {\raise.2ex\hbox{]}}
\newcommand\citehelper[1]{\startcite #1\stopcite}
\newcommand\makeaciteoftwo[2]
{\citehelper{\citeauthor{#1}, \citeyear{#1}; \citeyear{#2}}}